\documentclass{article}

\usepackage{microtype}
\usepackage{graphicx}
\usepackage{subfigure}
\usepackage{amssymb}
\usepackage{booktabs} % for professional tables
\usepackage{multirow}
\usepackage{afterpage}

\usepackage{bm}
\usepackage{enumitem}

\usepackage{hyperref}

\setlist[itemize]{leftmargin=*}
\setlist[enumerate]{leftmargin=*}

\newcommand{\guo}[1]{\textcolor{red}{GUO: #1}}
\newcommand{\yuzhe}[1]{\textcolor{red}{Yuzhe: #1}}

\newcommand{\name} {ME-Net}

\newenvironment{Itemize}%
{\begin{itemize}%
\setlength{\itemsep}{5pt}%
\setlength{\topsep}{0pt}%
\setlength{\partopsep}{0pt}%
\setlength{\parskip}{0pt}}%
{\end{itemize}}
\usepackage[accepted]{icml2019}

\icmltitlerunning{\name: Towards Effective Adversarial Robustness with Matrix Estimation}

\begin{document}

\twocolumn[
\icmltitle{\name: Towards Effective Adversarial Robustness with Matrix Estimation}



\icmlsetsymbol{equal}{*}

\begin{icmlauthorlist}
\icmlauthor{Yuzhe Yang}{mit}
\icmlauthor{Guo Zhang}{mit}
\icmlauthor{Dina Katabi}{mit}
\icmlauthor{Zhi Xu}{mit}
\end{icmlauthorlist}

\icmlaffiliation{mit}{MIT CSAIL, Cambridge, MA, USA}
% \icmlaffiliation{goo}{Googol ShallowMind, New London, Michigan, USA}
% \icmlaffiliation{ed}{School of Computation, University of Edenborrow, Edenborrow, United Kingdom}

\icmlcorrespondingauthor{Yuzhe Yang}{yuzhe@mit.edu}
\icmlcorrespondingauthor{Zhi Xu}{zhixu@mit.edu}

\icmlkeywords{Machine Learning, ICML}

\vskip 0.3in
]



\printAffiliationsAndNotice{}  % leave blank if no need to mention equal contribution

%%%%%%%%%%%%%%%%%%%%%%%%%%%%%%%%%%%%%%%%%%%%%%%%%%%%%%%%%%%%
%%%%%%%%%%%%%%%%%%%%%%%%  Abstract  %%%%%%%%%%%%%%%%%%%%%%%%
%%%%%%%%%%%%%%%%%%%%%%%%%%%%%%%%%%%%%%%%%%%%%%%%%%%%%%%%%%%%
\begin{abstract}
%Deep neural networks are vulnerable to adversarial attacks. The literature is rich with algorithms that can easily craft successful adversarial examples. In contrast, the performance of defense techniques still lacks behind. This paper proposes ME-Net, a defense method that leverages matrix estimation~(ME). In ME-Net, images are pre-processed using two steps: first pixels are randomly dropped from the image; then, the image is reconstructed using matrix completion. We show that this preprocessing method increases the classification margin (i.e., distance between classes), which naturally increases neural network robustness. Interestingly, and in contrast to past adversarial countermeasures which tend to hamper the performance of benign examples, ME-Net improves both robustness to adversarial attacks and performance with benign examples.  Our preprocessing alone is sufficient to foil traditional attacks, but can be susceptible to powerful white box attacks that propagate the gradient through the preprocessing step. We address this issue by combining our preprocessing step with training using adversarial examples. We conduct comprehensive experiments on prevailing benchmarks such as MNIST, CIFAR-10, SVHN and Tiny-ImageNet. Comparing ME-Net with state-of-the-art defense mechanisms shows that ME-Net consistently outperforms prior techniques, improving robustness against both black-box and white-box attacks. 

Deep neural networks are vulnerable to adversarial attacks. The literature is rich with algorithms that can easily craft successful adversarial examples. In contrast, the performance of defense techniques still lags behind. This paper proposes \name, a defense method that leverages matrix estimation~(ME). In \name, images are preprocessed using two steps: first pixels are randomly dropped from the image; then, the image is reconstructed using ME.
We show that this process destroys the adversarial structure of the noise, while re-enforcing the global structure in the original image.  Since humans typically rely on such global structures in classifying images, the process makes the network mode compatible with human perception.
We conduct comprehensive experiments on prevailing benchmarks such as MNIST, CIFAR-10, SVHN, and Tiny-ImageNet. Comparing \name\ with state-of-the-art defense mechanisms shows that \name\ consistently outperforms prior techniques, improving robustness against both black-box and white-box attacks.

%We conduct comprehensive experiments on prevailing benchmarks such as MNIST, CIFAR-10, SVHN and Tiny-ImageNet. The results verify that ME-Net consistently improves the state-of-the-art black-box and white-box robustness.
\end{abstract}

%%%%%%%%%%%%%%%%%%%%%%%%%%%%%%%%%%%%%%%%%%%%%%%%%%%%%%%%%%%%
%%%%%%%%%%%%%%%%%%%%%%  Introduction  %%%%%%%%%%%%%%%%%%%%%%
%%%%%%%%%%%%%%%%%%%%%%%%%%%%%%%%%%%%%%%%%%%%%%%%%%%%%%%%%%%%
\section{Introduction}
\label{intro}
State-of-the-art deep neural networks~(NNs) are vulnerable to adversarial examples~\cite{szegedy2013intriguing}. By adding small human-indistinguishable perturbation to the inputs, an adversary can fool neural networks to produce incorrect outputs with high probabilities. This phenomena raises increasing concerns for safety-critical scenarios such as the self-driving cars where NNs are widely deployed.

An increasing body of research has been aiming to either generate effective perturbations, or construct NNs that are robust enough to defend against such attacks. Currently, many effective algorithms exist to craft these adversarial examples, but defense techniques seem to be lagging behind. 
For instance, the state-of-the-art defense can only achieve less than 50\% adversarial accuracy for $\ell_\infty$ perturbations on datasets such as CIFAR-10~\cite{madry2017towards}. Under recent strong attacks, most defense methods have shown to break down to nearly 0\% accuracy~\cite{obfuscated-gradients}. 

As adversarial perturbations are carefully generated structured noise, a natural conjecture for defending against them is to destroy their structure.  A naive approach for doing so would randomly mask (i.e., zero out) pixels in the image. While such method can eliminate the adversarial structure within the noise through random information drop, it is almost certain to fail since it equally destroys the information of the original image, making NN inference even worse.

However, this naive starting point raises an interesting suggestion: instead of simply applying a random mask to the images, a preferable method should also reconstruct the images from their masked versions. In this case, the random masking destroys the crafted structures, but the reconstruction recovers the global structures that characterize the objects in the images.  Images contain some global structures. An image classified as cat should have at least a cat as its main body.  Humans use such global structure to classify images. In contrast the structure in adversarial perturbation is more local and defies the human eye.  If both training and testing are performed under the same underlying global structures (i.e., there is no distributional shift in training and testing), the network should be generalizable and robust. If the reconstruction can successfully maintain the underlying global structure, the masking-and-reconstruction pipeline can redistribute the carefully constructed adversarial noises to non-adversarial structures.

In this paper, we leverage matrix estimation~(ME) as our reconstruction scheme. ME is concerned with recovering a data matrix from noisy and incomplete observations of its entries, where exact or approximate recovery of a matrix is theoretically guaranteed if the true data matrix has some \emph{global structures}~(e.g., low rank). We view a masked adversarial image as a noisy and incomplete realization of the underlying clean image, and propose \name, a preprocessing-based defense that reverts a noisy incomplete image into a denoised version that maintains the underlying global structures in the clean image. \name\ realizes adversarial robustness by using such denoised global-structure preserving representations.

We note that the \name\ pipeline can be combined with different training procedures. In particular, we show that \name\ can be combined with standard stochastic gradient descent (SGD) or adversarial training, and in both cases improves adversarial robustness. This is in contrast with many preprocessing techniques which cannot leverage the benefits of adversarial training~\citep{buckman2018thermometer,song2018pixeldefend,guo2017countering}, and end up failing under the recent strong white-box attack~\cite{obfuscated-gradients}.

We provide extensive experimental validation of \name\ under the strongest black-box and white-box attacks on established benchmarks such as MNIST, CIFAR-10, SVHN, and Tiny-ImageNet, where \name\ outperforms state-of-the-art defense techniques.
Our implementation is available at: {\footnotesize \url{https://github.com/YyzHarry/ME-Net}}.
%\footnote{Code available at: {\scriptsize \url{https://github.com/YyzHarry/ME-Net}}}

We summarize our contributions as follows:
% \vspace{-5pt}
\begin{Itemize}
    \item We are the first to leverage matrix estimation as a general pipeline for image classification and defending against adversarial attacks.
    \item We show empirically that \name\ improves the robustness of neural networks under various $\ell_\infty$ attacks: 
    \begin{enumerate}
        \item \name\ alone significantly improves the state-of-the-art results on black-box attacks;
        \item Adversarially trained \name\ consistently outperforms the state-of-the-art defense techniques on white-box attacks, including the strong attacks that counter gradient obfuscation~\citep{obfuscated-gradients}.
    \end{enumerate}
    Such superior performance is maintained across various datasets: CIFAR-10, MNIST, SVHN and Tiny-ImageNet.
    \item We show additional benefits of \name\ such as improving generalization (performance on clean images).
\end{Itemize}

%%%%%%%%%%%%%%%%%%%%%%%%%%%%%%%%%%%%%%%%%%%%%%%%%%%%%%%%%%%%
%%%%%%%%%%%%%%%%%%%%%%%%  Methods  %%%%%%%%%%%%%%%%%%%%%%%%%
%%%%%%%%%%%%%%%%%%%%%%%%%%%%%%%%%%%%%%%%%%%%%%%%%%%%%%%%%%%%
%\section{Motivation and Method}
\section{\name}
\label{method}
We first describe the motivation and high level idea underlying our design. We then provide the formal algorithm.

\subsection{Design Motivation}
Images contain noise: even  ``clean'' images taken from a camera contain white noise from the environment. Such small, unstructured noise seems to be tolerable for modern deep NNs, which achieve human-level performance. However, the story is different for carefully constructed noise. Structured, adversarial noise (i.e., adversarial examples) can easily corrupt the NN results, leading to incorrect prediction from human's perspective. This means that to achieve robustness to adversarial noise, we need to eliminate/reduce the crafted adversarial structure. Of course, while doing so, we need to maintain the intrinsic structures in the image that allow a human to make correct classifications.  

We can model the problem as follows: An image is a superposition of: 1) intrinsic true structures of the data in the scene,  2) adversarial carefully-structured noise, and 3) non-adversarial noise. Our approach is first to destroy much of the crafted structure of the adversarial noise by randomly masking (zeroing out) pixels in the image. Of course, this process also increases the overall noise in the image (i.e., the non-adversarial noise) and also negatively affects the underlying intrinsic structures of the scene. Luckily however there is a well-established theory for recovering the underlying intrinsic structure of data from noisy and incomplete (i.e., masked) observations.  Specifically, if we think of an image as a matrix, then we can leverage a well-founded literature on matrix estimation (ME) which allows us to recover the true data in a matrix from noisy and incomplete observations~\citep{candes2009exact,keshavan2010matrix,chatterjee2015matrix}.  Further, ME provides provable guarantees of exact or approximate recovery of the true matrix  if the true data has some global structures (e.g., low rank)~\citep{davenport2016overview,chen2018harnessing}. Since images naturally have global structures (e.g., an image of a cat, has a cat as a main structure), ME is guaranteed to restore the intrinsic structures of the clean image. 

Another motivation for our method comes from adversarial training, where an NN is trained with adversarial examples.  Adversarial training is widely adopted to increase the robustness of neural networks. However, recent theoretical work formally argues that adversarial training requires substantially more data to achieve robustness \citep{schmidt2018adversarially}. The natural question is then how to automatically obtain more data, with the purpose of creating samples that can help robustness. Our masking-then-reconstruction pipeline provides exactly one such automatic solutions. By using different random masks, we can create variations on each image, where all such variations maintain the image's underlying true global structures.  We will see later in our results that this indeed provides significant gain in robustness.

\subsection{Matrix Estimation Pipeline}
Having described the intuition underlying \name, we next provide a formal description of matrix estimation (ME), which 
constitutes the reconstruction step in our pipeline.  
%ME is a fairly mature topic with strong theoretical guarantees and appealing practical performance.

\begin{figure}[!t]
\centering
\subfigure[MNIST]{
    \label{cdf_mnist}
    \includegraphics[height=0.163\textwidth]{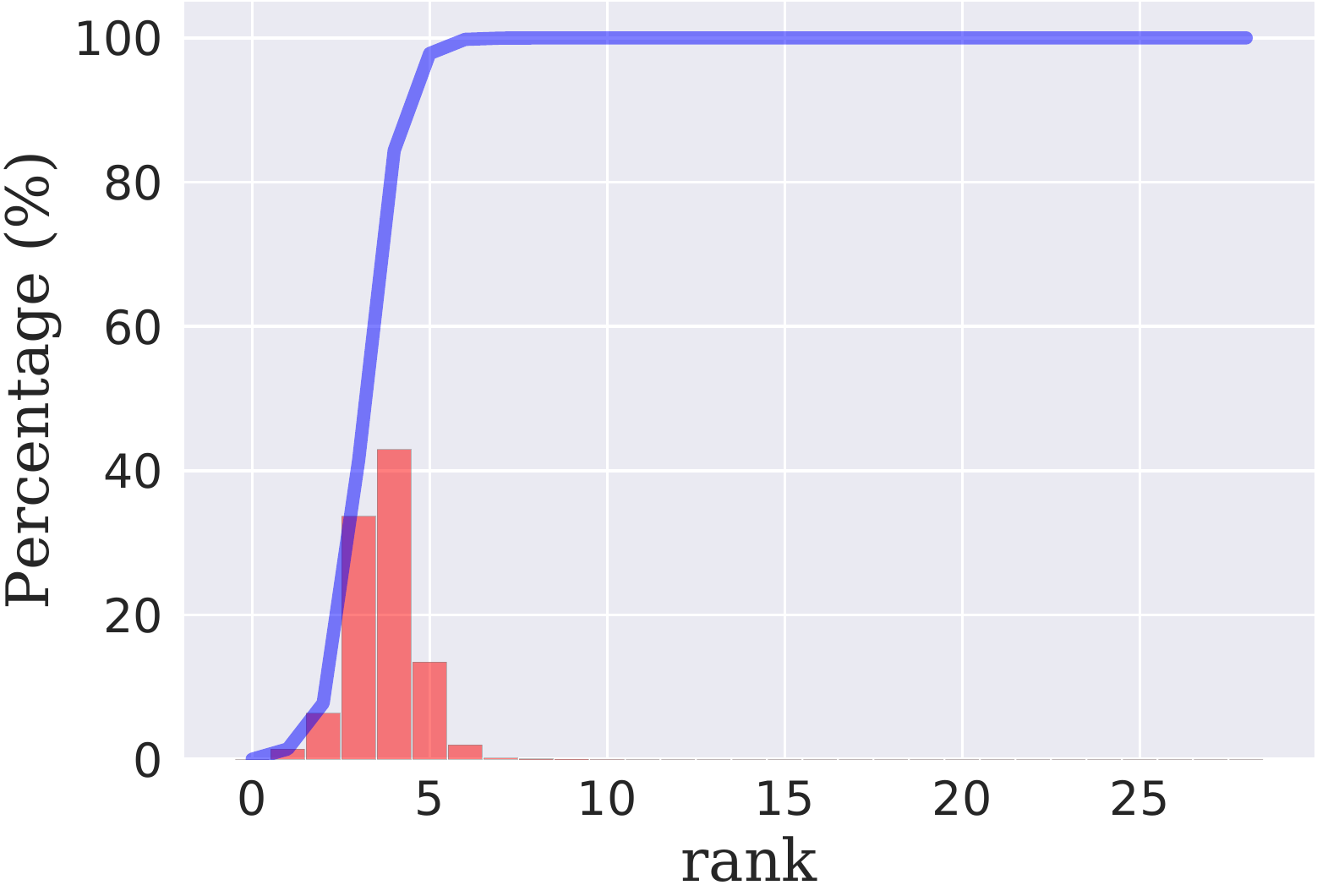}
}
\hspace{-1.75ex}
\subfigure[CIFAR-10]{
    \label{cdf_cifar}
    \includegraphics[height=0.163\textwidth]{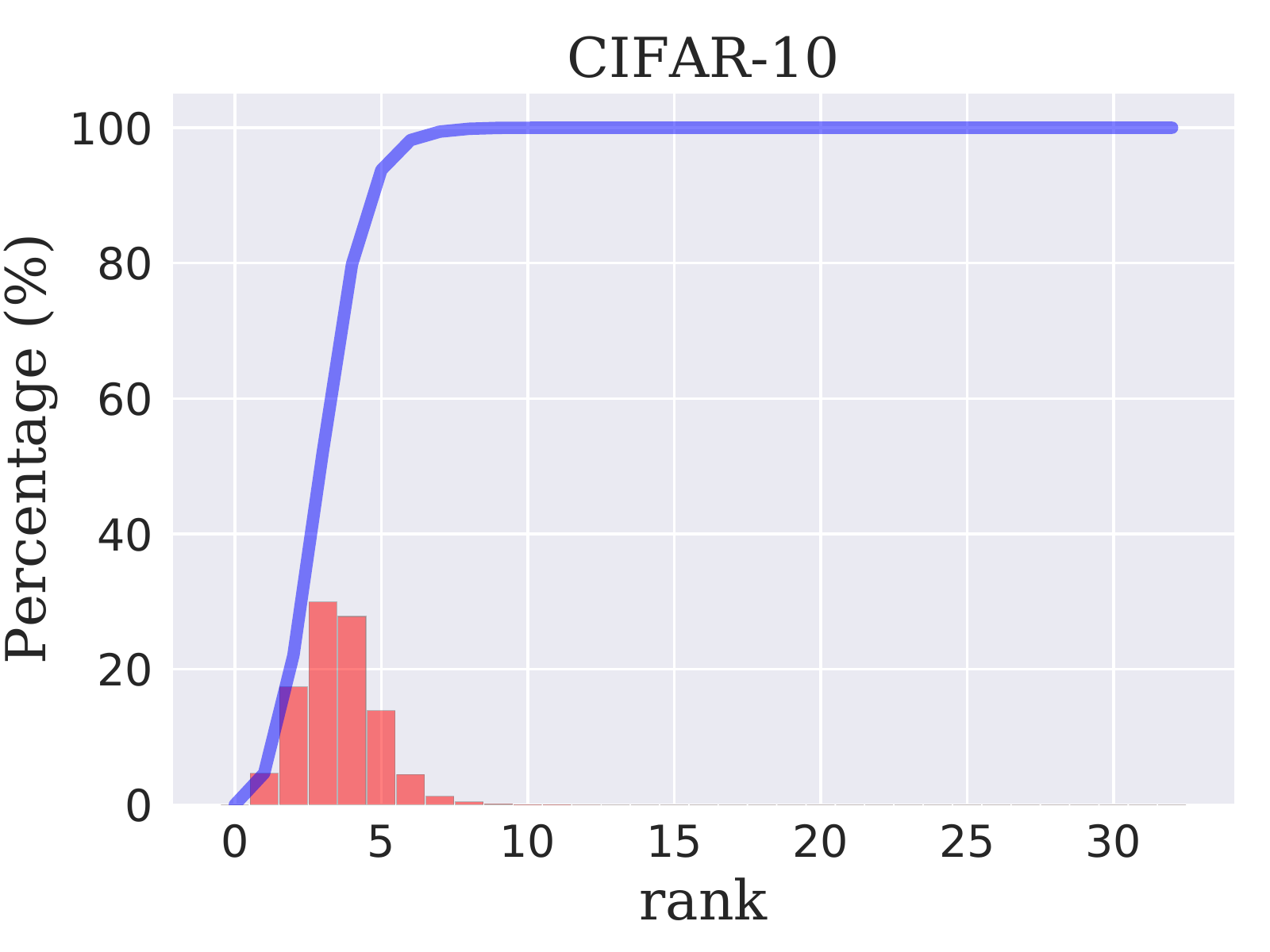}
}
\\[-1ex]
\subfigure[SVHN]{
    \label{cdf_svhn}
    \includegraphics[height=0.163\textwidth]{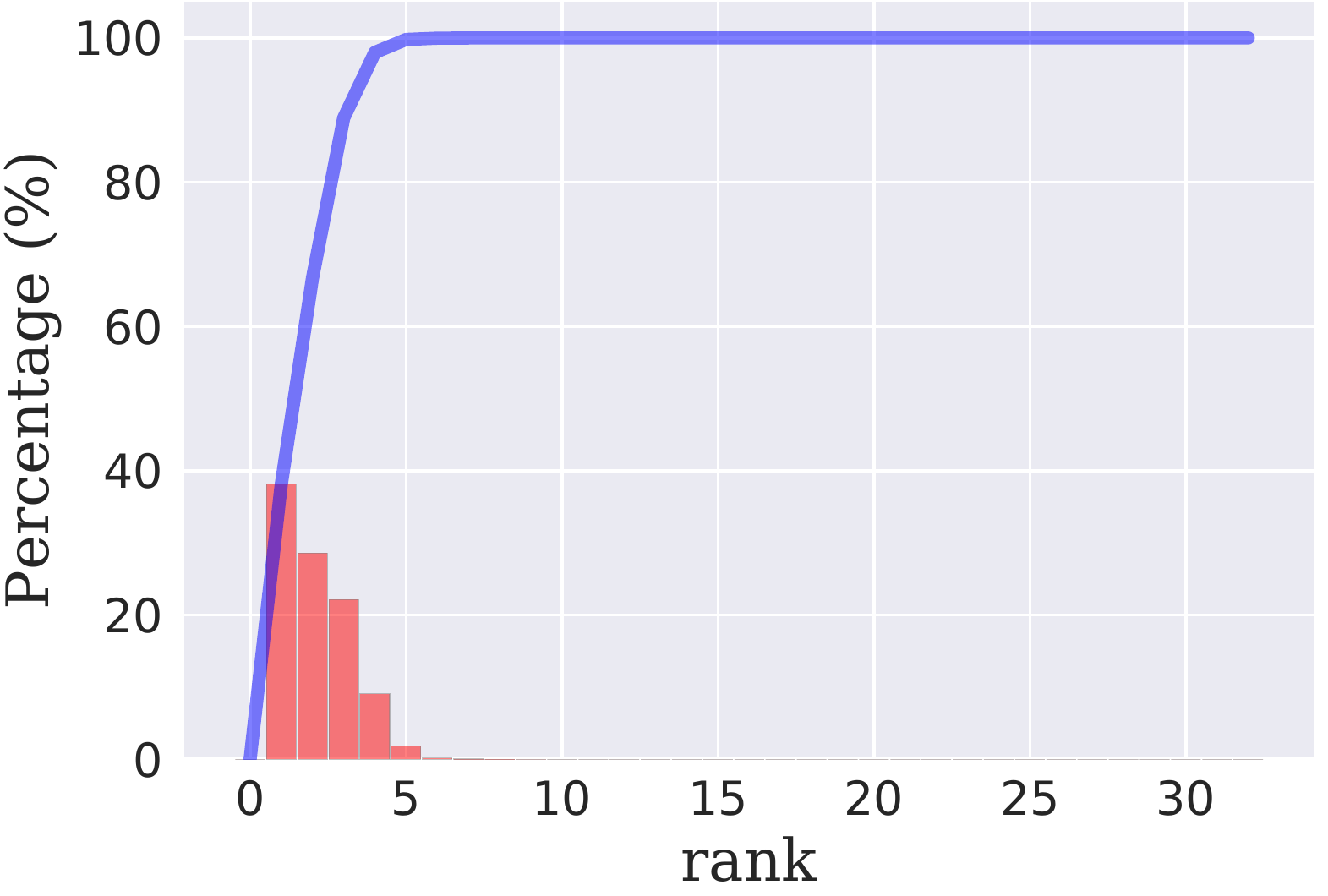}
}
\hspace{-1.8ex}
\subfigure[Tiny-ImageNet]{
    \label{cdf_timgnet}
    \includegraphics[height=0.163\textwidth]{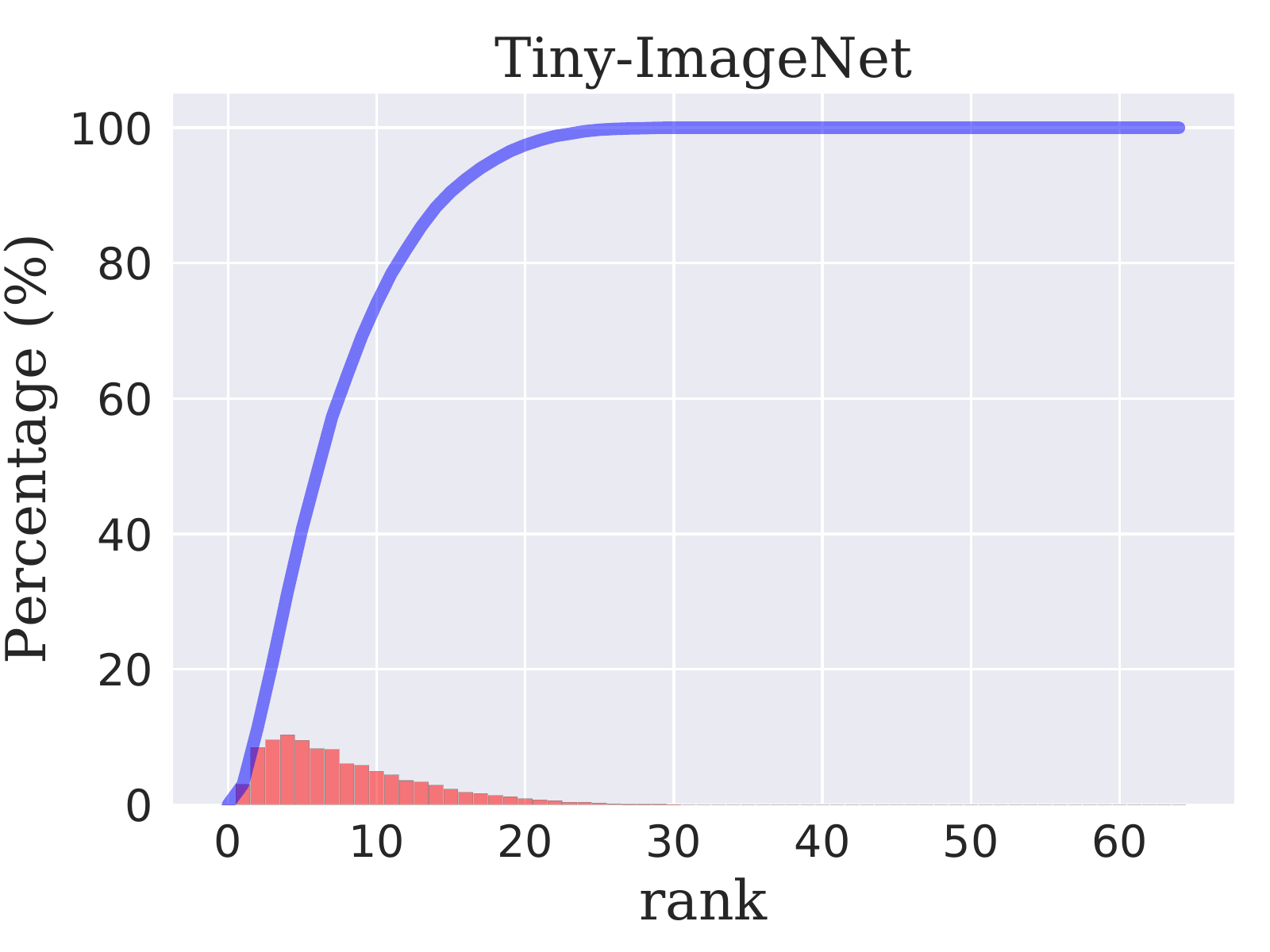}
}
\vspace{-0.5cm}
\caption{\textbf{The approximate rank of different datasets.} We plot the histogram~(in red) and the empirical CDF~(in blue) of the approximate rank for images in each dataset.}
\label{fig:dataset cdf}
\vspace{-0.2cm}
\end{figure}

{\bf Matrix Estimation.}~~
Matrix estimation is concerned with recovering a data matrix from noisy and incomplete observations of its entries. Consider a true, unknown data matrix $M\in\mathbb{R}^{n\times m}$. Often, we have access to a subset $\Omega$ of entries from a noisy matrix $X\in\mathbb{R}^{n\times m}$ such that $\mathbb{E}[X]=M$. For example, in recommendation system, there are true, unknown ratings for each product from each user. One often observes a subset of noisy ratings if the user actually rates the product online. Technically, it is often assumed that each entry of $X$, $X_{ij}$, is a random variable independent of the others, which is observed with probability $p\in(0,1]$ (i.e., missing with probability $1-p$). The theoretical question is then formulated as finding an estimator $\hat{M}$, given noisy, incomplete observation matrix $X$, such that $\hat{M}$ is ``close'' to $M$. The closeness is typically measured by some matrix norm, $||\hat{M}-M||$, such as the Frobenius norm.

Over the years, extensive algorithms have been proposed. They range from simple spectral method such as universal singular value thresholding (USVT) \citep{chatterjee2015matrix}, which performs SVD on the observation matrix $X$ and discards small singular values (and corresponding singular vectors), to convex optimization based methods, which minimize the nuclear norm \citep{candes2009exact}, i.e.:
\begin{equation}
    \min_{\hat{M}\in\mathbb{R}^{n\times m}} ||\hat{M}||_* \quad \textrm{s.t.}\:\hat{M}_{ij}\approx X_{ij},\:\forall\:(i,j)\in\Omega,
\end{equation}
where $||\hat{M}||_*$ is the nuclear norm of the matrix~(i.e., sum of the singular values). To speed up the computation, the Soft-Impute algorithm  \citep{mazumder2010spectral} reformulates the optimization using a regularization parameter $\lambda \geq 0$:
\begin{equation}
     \min_{\hat{M}\in\mathbb{R}^{n\times m}} \frac{1}{2}\sum_{(i,j)\in\Omega}\left(\hat{M}_{ij}-X_{ij}\right)^2+\lambda||\hat{M}||_*. \label{eq:softimpute}
\end{equation}
In this paper, we view ME as a reconstruction oracle from masked images, rather than focusing on specific algorithms. 

The key message in the field of ME is: if the true data matrix $M$ has some \emph{global structures}, exact or approximate recovery of $M$ can be theoretically guaranteed~\citep{candes2009exact,chatterjee2015matrix,chen2018harnessing}. This strong theoretical guarantee serves as the foundation for employing ME to reconstruct structures in images. In the literature, the most studied global structure is low rank. Latent variable models, where each row $i$ and each column $j$ are associated with some features $u_i\in\mathbb{R}^{r}$ and $v_j\in\mathbb{R}^r$ and $M_{ij}=f(u_i,v_j)$ for some function $f$, have also been investigated~\citep{chatterjee2015matrix,borgs2017thy}. To some extent, both could be good models for images.

{\bf Empirical Results.}~~
Before closing, we empirically show that images have strong global structures (i.e., low rank).  We consider four datasets:  MNIST, CIFAR-10, SVHN, and Tiny-ImageNet. We perform SVD on each image and compute its approximate rank, which is defined as the minimum number of singular values necessary to capture at least 90\% of the energy in the image.
Fig.~\ref{fig:dataset cdf} plots the histogram and the empirical CDF of the approximate ranks for each dataset. As expected, images in all datasets are relatively low rank. Specifically, the vast majority of images in MNIST, CIFAR-10, and SVHN have a rank less than 5. The rank of images in Tiny-ImageNet is larger but still significantly less than the image dimension ($\sim$10 vs. 64). 
This result shows that images tend to be low-rank, which implies the validity of using ME as our reconstruction oracle to find global structures.

\begin{figure}[!t]
    \centering
    \includegraphics[width=0.5\textwidth]{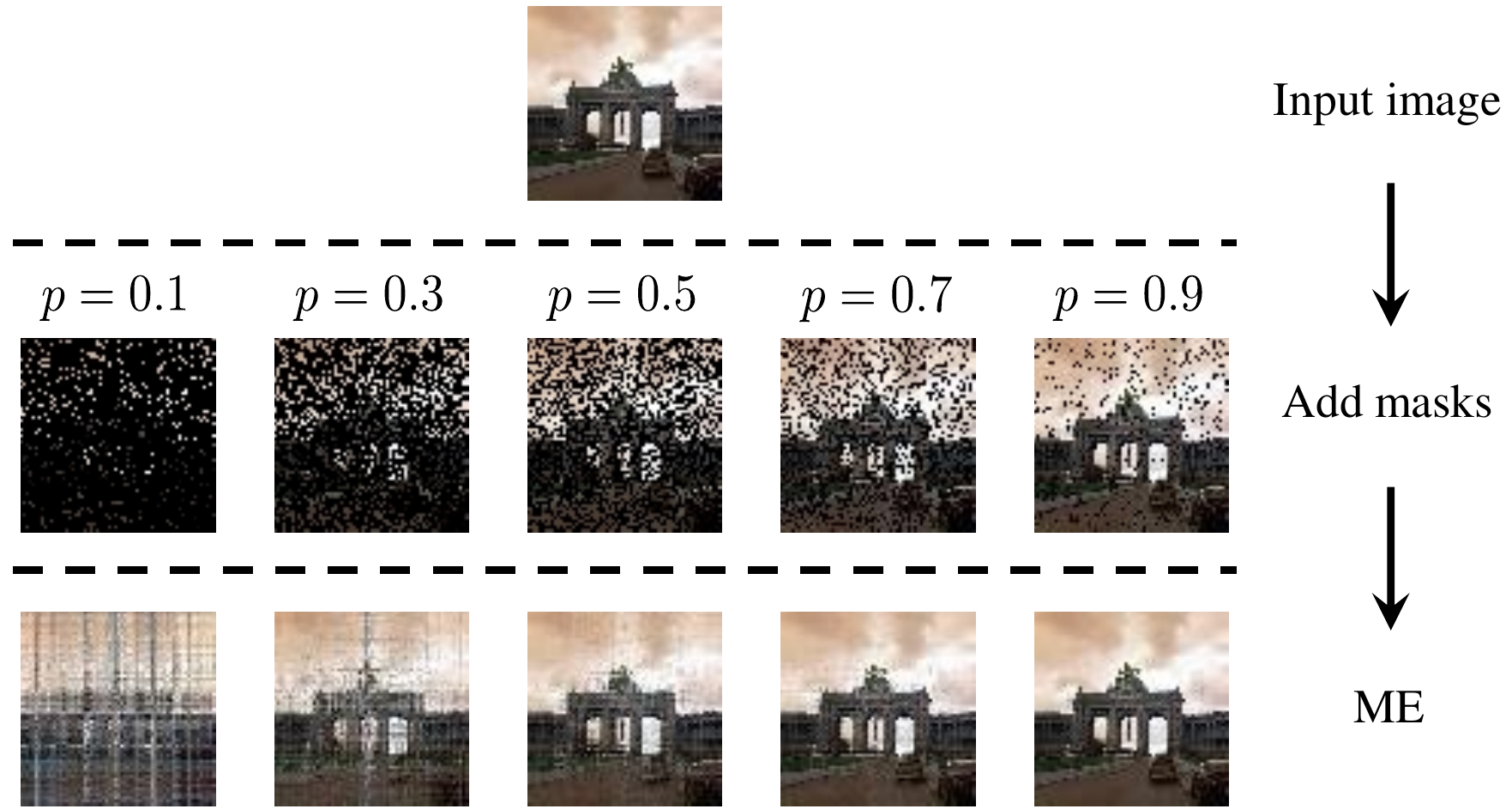}
\vspace{-0.5cm}
    \caption{\textbf{An example of how ME affects the input images.} We apply different masks and show the reconstructed images by ME.}
    \label{fig:visual me}
    \vspace{-0.2cm}
\end{figure}

Next, we show in Fig.~\ref{fig:visual me} the results of ME-based reconstruction for different masks. Evidently, the global structure~(the gate in the image) has been maintained even when $p$, the probability of observing the true pixel, is as low as $0.3$. This shows that despite random masking we should be able to reconstruct the intrinsic global image structure from the masked adversarial images. Our intuition is that humans use such underlying global structures for image classification, and if we can maintain such global structures while weakening other potentially adversarial structures, we can force both training and testing to focus on human recognizable structures and increase robustness to adversarial attacks.  

% \vspace{-0.2cm}

% \newtheorem{remark}{\bf Remark}
% \begin{remark}
% Strictly speaking, images are tensors because of the three RGB channels. In this work, we view the three channels as independent matrices: for each image, we apply the same mask to each channel and then separately run ME to recover the matrix. A more accurate approach might be to apply some tensor estimation algorithms. We apply ME because it is well studied and computationally faster, but we believe that applying tensor estimation could be a very promising future direction.
% \end{remark}

\begin{figure}[!t]
    \centering
    \includegraphics[width=0.5\textwidth]{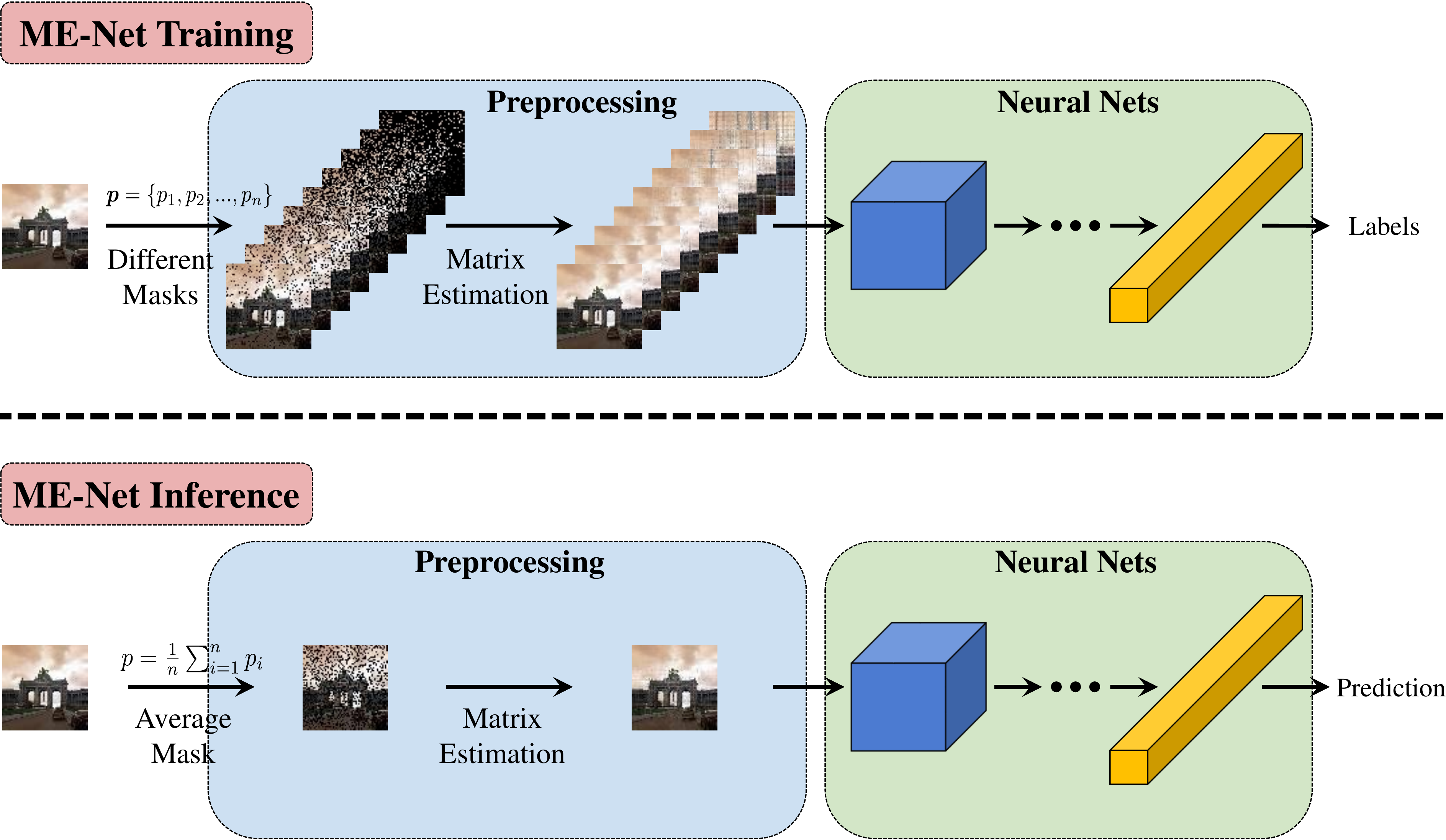}
\vspace{-0.5cm}
    \caption{An illustration of ME-Net training and inference process.}
    \label{fig:menet}
\vspace{-0.3cm}
\end{figure}

\subsection{Model}\label{sec:model}
We are now ready to formally describe our technique, which we refer as \name. The method is illustrated in Fig.~\ref{fig:menet} and summarized as follows:

\begin{Itemize}
    \item \textbf{\name\ Training:}   Define a mask as an image transform in which each pixel is preserved with probability $p$  and set to zero with probability $1-p$. For each training image $X$, we apply $n$ masks with  probabilities $\{p_1,p_2,\dots,p_n\}$, and obtain $n$ masked images $\{X^{(1)}, X^{(2)},\dots, X^{(n)}\}$. An ME algorithm is then applied to obtain reconstructed images $\{\hat{X}^{(1)}, \hat{X}^{(2)},\dots, \hat{X}^{(n)}\}$. We train the network on the reconstructed images $\{\hat{X}^{(1)}, \hat{X}^{(2)},\dots, \hat{X}^{(n)}\}$ as usual via SGD. Alternatively, adversarial training can also be readily applied in our framework.
\vspace{6pt}
    \item \textbf{\name\ Inference:} For each test image $X$, we randomly sample a mask with probability $p=\frac{1}{n}\sum_{i=1}^n p_i$, i.e., the average of the masking probabilities during training. The masked image is then processed by the same ME algorithm used in training to obtain $\hat{X}$. Finally, $\hat{X}$ is fed to the network for prediction. 
\end{Itemize}
\vspace{-6pt}

Note that we could either operate on the three RGB channels separately as independent matrices or jointly by concatenating them into one matrix. In this paper, we take the latter approach as their structures are closely related. 
%as independent matrices: for each image, we apply the same mask to each channel and then separately run ME to recover the matrix.} 
We provide additional details of ME-Net in Appendix~\ref{appendix:pseudo code} and \ref{appendix:training details}.

%%%%%%%%%%%%%%%%%%%%%%%%%%%%%%%%%%%%%%%%%%%%%%%%%%%%%%%%%%%%
%%%%%%%%%%%%%%%%%%%%%%%  Understand  %%%%%%%%%%%%%%%%%%%%%%%
%%%%%%%%%%%%%%%%%%%%%%%%%%%%%%%%%%%%%%%%%%%%%%%%%%%%%%%%%%%%
%\section{Understanding \name}
%\label{understand}
%\input{understand.tex}

%%%%%%%%%%%%%%%%%%%%%%%%%%%%%%%%%%%%%%%%%%%%%%%%%%%%%%%%%%%%
%%%%%%%%%%%%%%%%%%%%%%  Experiments  %%%%%%%%%%%%%%%%%%%%%%%
%%%%%%%%%%%%%%%%%%%%%%%%%%%%%%%%%%%%%%%%%%%%%%%%%%%%%%%%%%%%
\section{Evaluation}
\label{experiment}
We evaluate \name\ empirically under $\ell_\infty$-bounded attacks and compare it with state-of-the-art defense techniques.

{\bf Experimental Setup:}~~
We implement \name\ as described in Section~\ref{sec:model}. During training, for each image we randomly sample 10 masks with different $p$ values and apply matrix estimation for each masked image to construct the training set. During testing, we sample a single mask with $p$ set to the average of the values used during training, apply the \name\ pipeline, and test on the reconstructed image. Unless otherwise specified, 
we use the Nuclear Norm minimization method~\cite{candes2009exact} for matrix estimation.
% we use universal singular value thresholding (USVT) \citep{chatterjee2015matrix} for matrix estimation.

We experiment with two versions of \name: the first version uses standard stochastic gradient descent (SGD) to train the network, and the second version uses adversarial training, where the model is trained with adversarial examples. 

For each attack type, we compare \name\ with state-of-the-art defense techniques for the attack under consideration.  
For each technique, we report accuracy as the percentage of adversarial examples that are correctly classified.\footnote{
%Some prior work generates adversarial examples only from clean images that are correctly classified by the defense scheme~\citep{xie2017mitigating}, and reports accuracy on those examples only. In contrast, 
To be consistent with literature, we generate adversarial examples from the whole dataset and use all of them to report accuracy. }
As common in prior work~\citep{madry2017towards,buckman2018thermometer,song2018pixeldefend}, we focus on robustness against $\ell_\infty$-bounded attacks, and  generate adversarial examples using standard methods such as the CW attack~\citep{carlini2017towards}, Fast Gradient Sign Method~(FGSM)~\citep{goodfellow2015explaining}, and Projected Gradient Descent~(PGD) which is a more powerful adversary that performs a multi-step variant of FGSM~\citep{madry2017towards}.

{\bf Organization:}
We first perform an extensive study on CIFAR-10 to validate the effectiveness of \name\ against black-box and white-box attacks. We then extend the results to other datasets such as MNIST, SVHN, and Tiny-ImageNet.
We also  provide additional supporting results in Appendix~\ref{appendix:cifar}, \ref{appendix:mnist}, \ref{appendix:svhn}, \ref{appendix:timgnet}, \ref{appendix:tradeoff} and \ref{appendix:majority vote}.
Additional hyper-parameter studies, such as random restarts and different number of masks, can be found in Appendix~\ref{appendix:random restart}, \ref{appendix:diff me methods} and \ref{appendix:hyperparam}.

\subsection{Black-box Attacks}
In black-box attacks, the attacker has no access to the network model; it only observes the inputs and outputs.  We evaluate \name\ against three kinds of black-box attacks:
\begin{Itemize}
    \item \textbf{Transfer-based attack:} A copy of the victim network is trained with the same training settings. We apply CW, FGSM and PGD attacks on the copy network to generate black-box adversarial examples. We use the same attack parameters as in~\citep{madry2017towards}: total perturbation $\varepsilon$ of $8/255$ (0.031), step size of $2/255$ (0.01). For PGD attacks, we use 7, 20 and 40 steps. Note that we only consider the \emph{strongest} transfer-based attacks, i.e., we use \emph{white-box} attacks on the independently trained copy to generate black-box examples.
    \item \textbf{Decision-based attack:} We apply the newly proposed Boundary attack~\citep{brendel2017decision} which achieves better performance than transfer-based attacks. We apply 1000 attack steps to ensure convergence.
    \item \textbf{Score-based attack:} We also apply the state-of-the-art SPSA attack~\citep{uesato2018adversarial} which is strong enough to bring the accuracy of several defenses to near zero. We use a batch-size of 2048 to make the SPSA strong, and leave other hyper-parameters unchanged.
\end{Itemize}

As in past work that evaluates robustness on CIFAR-10 \citep{madry2017towards,buckman2018thermometer}, we use the standard ResNet-18 model in~\citep{He_2016_CVPR}. In training \name, we experiment with different settings for $p$. We report the results for $p \in [0.8,1]$ below, and refer the reader to the Appendix for the results with other $p$ values. 

Since most defenses experimented only with transfer-based attacks, we first compare \name\ to past defenses under transfer-based attacks. For comparison, we select a state-of-the-art adversarial training defense~\citep{madry2017towards} and a preprocessing method~\citep{buckman2018thermometer}.  We compare these schemes against \name\ with standard SGD training. The results are shown in Table~\ref{table:cifar black compare}. They reveal that even without adversarial training, \name\ is much more robust than prior work to black-box attacks, and can improve accuracy by 13\% to 25\%, depending on the attack.  
\begin{figure}[!t]
\centering
\subfigure[Vanilla under adv. attack.]{
    \label{tsne_adv}
    \includegraphics[width=0.225\textwidth]{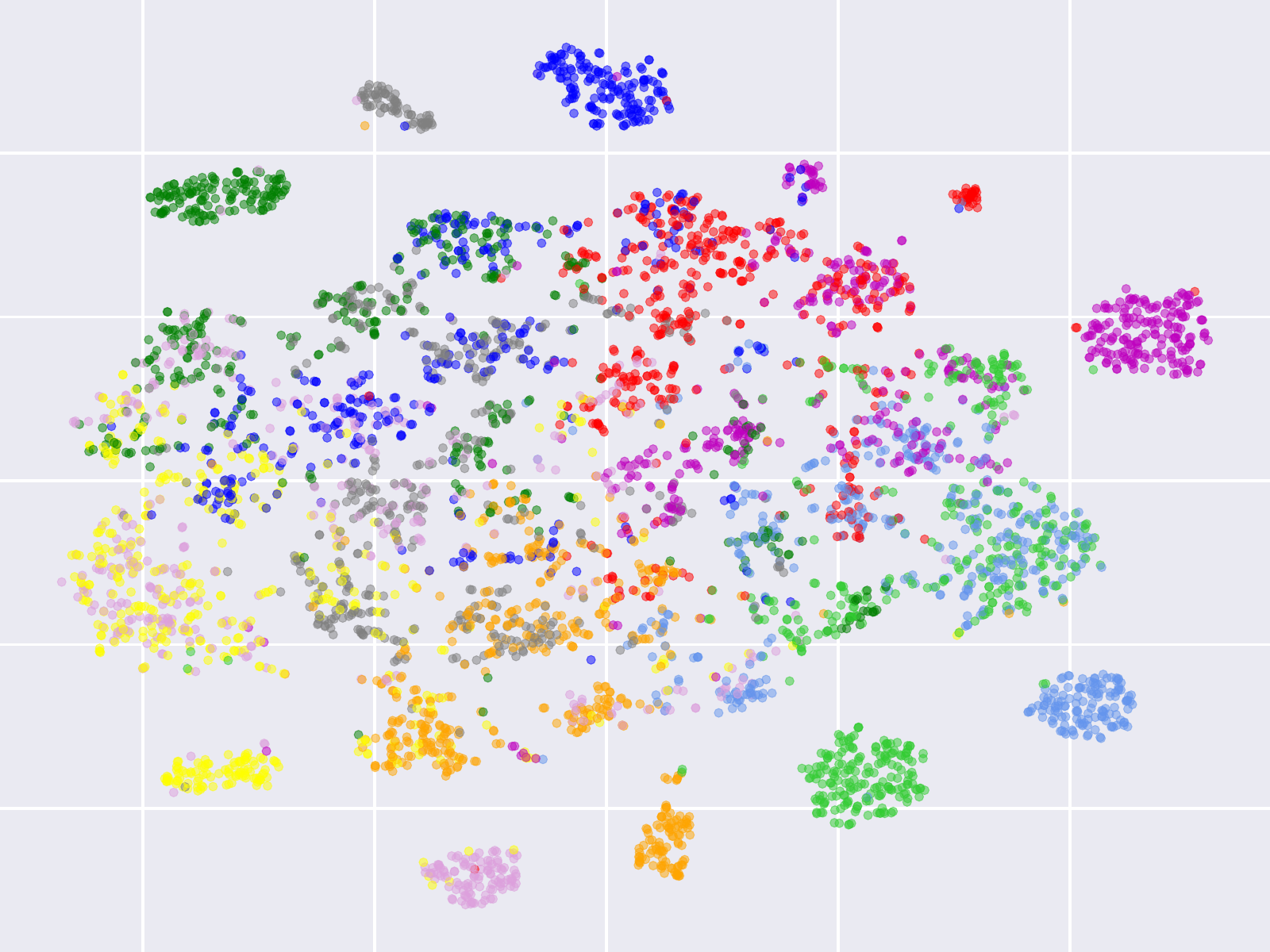}
}
\hspace{0.1ex}
\subfigure[ME-Net under adv. attack.]{
    \label{tsne_mc}
    \includegraphics[width=0.225\textwidth]{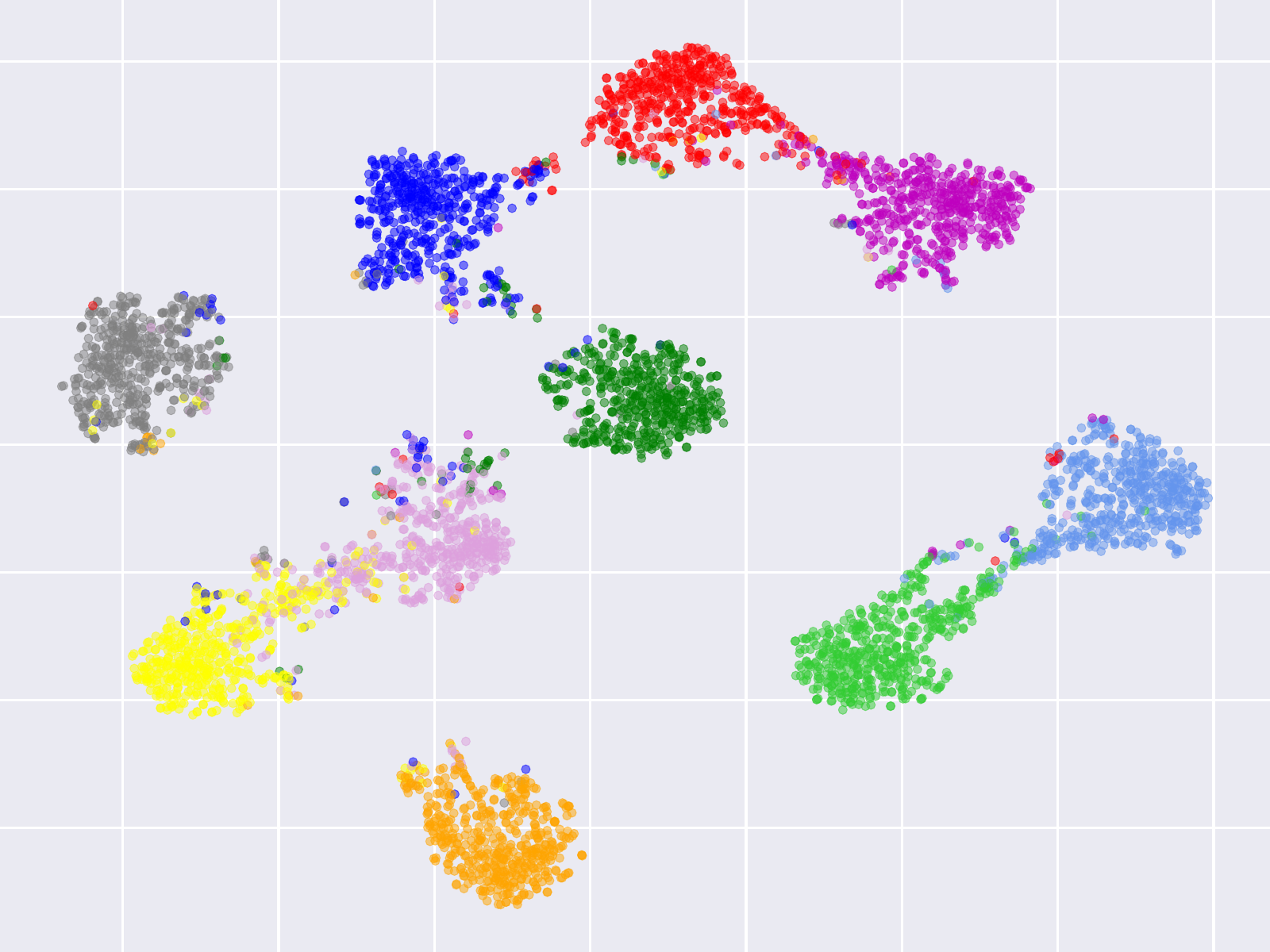}
}
\vspace{-0.5cm}
\caption{\textbf{Class separation under black-box adversarial attack.} The vectors right before the softmax layer are projected to a 2D plane using t-SNE~\cite{maaten2008visualizing}.}
\label{fig:tsne}
\vspace{-0.3cm}
\end{figure}

\begin{figure}[!t]
\centering
\subfigure[Clean data.]{
    \label{cdf_clean_mcclean}
    \includegraphics[height=0.196\textwidth]{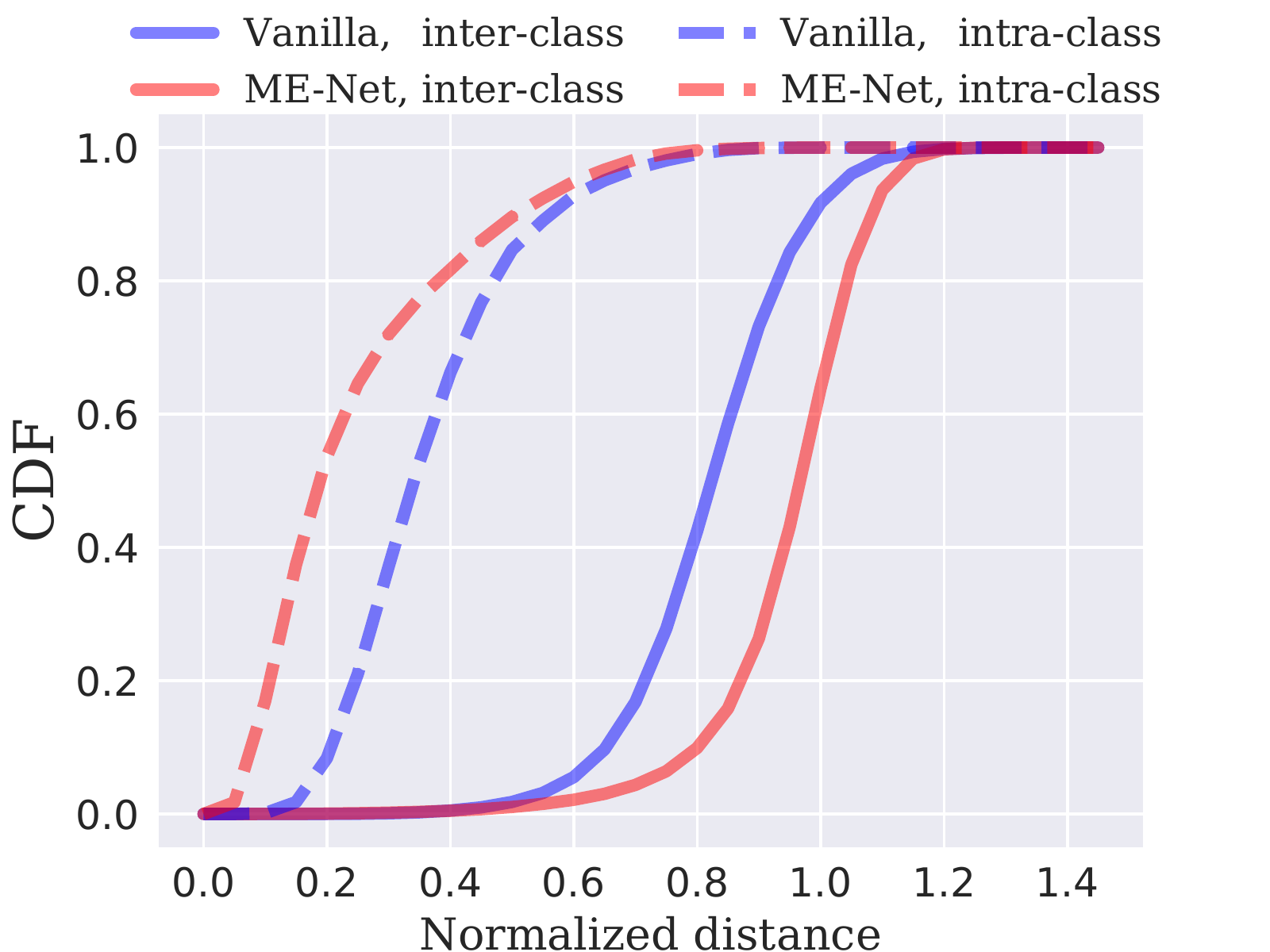}
}
\hspace{-1.73ex}
\subfigure[Black-box adv. attack.]{
    \label{cdf_adv_mc}
    \includegraphics[height=0.196\textwidth]{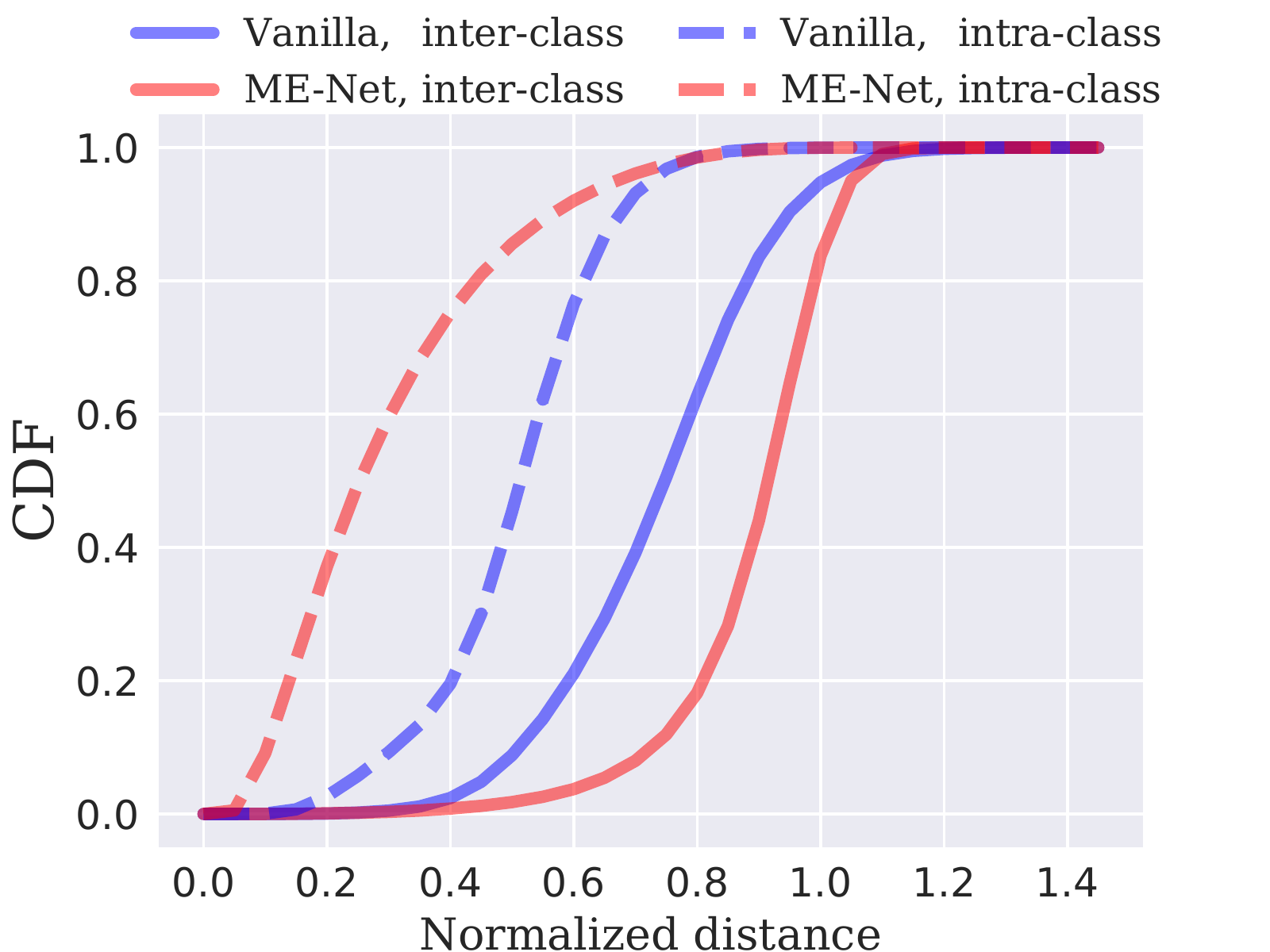}
}
\vspace{-0.5cm}
\caption{\textbf{The empirical CDF of the distance within and among classes.} We quantitatively show the intra-class and inter-class distances between vanilla model and ME-Net on clean data and under black-box adversarial attacks.}
\label{fig:intra inter distance}
\vspace{-0.3cm}
\end{figure}

To gain additional insight, we look at the separation between different classes under black-box transfer-based attack, for the vanilla network and \name. Fig.~\ref{tsne_adv} and \ref{tsne_mc} show the 2D projection of the vectors right before the output layer (i.e., softmax layer), for the test data in the vanilla model and \name. The figures show that when the vanilla model is under attack, it loses its ability to separate different classes. In contrast, \name\ can sustain clear separation between classes even in the presence of black-box attack. 

To further understand this point, we compute the Euclidean distance between classes and within each class. Fig.~\ref{fig:intra inter distance} plots the empirical CDFs of the intra-class and inter-class distance between the vectors before the output layer, for both the vanilla classifier and \name. The figure shows results for both clean data and adversarial examples.  Comparing \name\ (in red) with the vanilla classifier (in blue), we see that  \name\  both reduces the distance within each class, and improves the separation between classes; further this result applies to both clean and adversarial examples. Overall, these visualizations offer strong evidence supporting the improved robustness of \name.

% =========================== CIFAR Black Compare =========================== %
\begin{table}[!t]
\centering
\scalebox{0.88}{
\begin{tabular}{@{}ccccc@{}}
\toprule
\textbf{Method}  &  \textbf{Training}  &  \textbf{CW}  &  \textbf{FGSM}  &  \textbf{PGD (7 steps)} \\ \midrule\midrule
Vanilla  &   SGD   &  8.9\%  &  24.8\%  &  7.6\%  \\ \midrule
Madry    &  Adv. train  &  78.7\%  &  67.0\%  &  64.2\%   \\ \midrule
Thermometer  &  SGD  &  $-$  &  $-$  &  53.5\%  \\ \midrule
Thermometer  &  Adv. train  &  $-$  &  $-$  &  77.7\%  \\ \midrule
\textbf{ME-Net}  &  SGD  &  \textbf{93.6\%}  &  \textbf{92.2\%}  &  \textbf{91.8\%}  \\ \bottomrule
\end{tabular}
}
\vspace{-1ex}
\caption{\textbf{CIFAR-10 black-box results under transfer-based attacks.} We compare \name\ with state-of-the-art defense methods under both SGD and adversarial training.}
\label{table:cifar black compare}
\end{table}
% =========================================================================== %

Finally, we also evaluate \name\ under other strong black-box attacks.  Table~\ref{table:cifar black ours} summarizes these results demonstrating that  \name\ consistently achieves high robustness under different black-box attacks.

% ========================== CIFAR Black Extensive ========================== %
\begin{table}[!t]
\centering
\scalebox{0.68}{
\begin{tabular}{@{}cccccccc@{}}
\toprule
\multirow{2.5}{*}{\textbf{Attacks}} & \multirow{2.5}{*}{\textbf{CW}} & \multirow{2.5}{*}{\textbf{FGSM}} & \multicolumn{3}{c}{\textbf{PGD}} & \multirow{2.5}{*}{\textbf{Boundary}} & \multirow{2.5}{*}{\textbf{SPSA}}\\ \cmidrule(lr){4-6}
 & & &  7 steps  &  20 steps  &  40 steps  & &  \\ \midrule\midrule
Vanilla &  8.9\%  &  24.8\%  &  7.6\%  &  1.8\%  &  0.0\%  &  3.5\%  & 1.4\%  \\ \midrule
\textbf{ME-Net}  &  \textbf{93.6\%}  &  \textbf{92.2\%}  &  \textbf{91.8\%}  &  \textbf{91.8\%}  &  \textbf{91.3\%}  &  \textbf{87.4\%}  & \textbf{93.0\%}  \\ \bottomrule
\end{tabular}
}
\vspace{-1ex}
\caption{\textbf{CIFAR-10 extensive black-box results.} We show significant adversarial robustness of \name\ under different strong black-box attacks.}
\label{table:cifar black ours}
\end{table}
% =========================================================================== %

\subsection{White-box Attacks}
In white-box attacks, the attacker has full information about the neural network model~(architecture and weights) and defense methods. To evaluate robustness against such white-box attacks, we use the BPDA attack proposed in~\citep{obfuscated-gradients}, which has successfully circumvented a number of previously effective defenses, bringing them to near 0 accuracy. Specifically, most defense techniques rely on preprocessing methods which can cause \emph{gradient masking} for gradient-based attacks, either because the preprocessing is not differentiable or the gradient is useless. BPDA addresses this issue by using a ``differentiable approximation'' for the backward pass. As such, until now no preprocessing method is effective under white-box attacks. In \name, the backward pass is not differentiable, which makes BPDA the strongest white-box attack. We use PGD-based BPDA and experiment with different number of attack steps.

For white box attacks, we distinguish two cases: defenses that use only preprocessing (without adversarial training), and defenses that incorporate adversarial training. All defenses that incorporate adversarial training, including \name, are trained with PGD with 7 steps. 

Table~\ref{table:cifar white compare} shows a comparison of the performance of various preprocessing methods against the BPDA white-box attack. 
We compare \name\  with three preprocessing defenses, i.e.,
the PixelDefend method~\citep{song2018pixeldefend}, the Thermometer method~\citep{buckman2018thermometer}, and the total variation~(TV) minimization method~\citep{guo2017countering}. The results in the table for \citep{song2018pixeldefend,buckman2018thermometer} are directly taken from~\citep{obfuscated-gradients}.
Since the TV minimization method is not tested on CIFAR-10, we implement this method using the same setting used with \name.  The table shows that preprocessing alone is vulnerable to the BPDA white-box attack, as all schemes perform poorly under such attack. Interestingly however, the table also shows that \name's preprocessing is significantly more robust to BPDA than other preprocessing methods. We attribute this difference to that \name's preprocessing step focuses on protecting the global structures in images.

% ==================== CIFAR White Compare (Prep. only) ===================== %
\begin{table}[!t]
\centering
\scalebox{0.95}{
\begin{tabular}{@{}cccc@{}}
\toprule
\textbf{Method} & \textbf{Type} & \textbf{Steps} & \textbf{Accuracy} \\ \midrule\midrule
%Madry          & Adv. train & 100          & \textbf{45.1\%} \\ \midrule\midrule
Thermometer    & Prep.      & 40           & 0.0\%*          \\ \midrule
PixelDefend    & Prep.      & 100          & 9.0\%*          \\ \midrule
TV Minimization    & Prep.      & 100      & 0.4\%          \\ \midrule
\textbf{ME-Net} & Prep.      & 1000         & \textbf{40.8\%} \\ \bottomrule
\end{tabular}
}
\vspace{-1ex}
\caption{\textbf{White-box attack against pure preprocessing schemes.} We use PGD or BPDA attacks in white-box setting. Compared to other pure preprocessing methods, \name\ can increase robustness by a significant margin. *Data from~\cite{obfuscated-gradients}.}
\label{table:cifar white compare}
\vskip -0.15in
\end{table}
% =========================================================================== %

Next we report the results of white-box attacks on schemes that use adversarial training. 
One key characteristic of \name\ is its orthogonality with adversarial training.
Note that many preprocessing methods propose combining adversarial training, but the combination actually performs worse than adversarial training alone~\citep{obfuscated-gradients}. Since \name's preprocessing already has a decent accuracy under the strong white-box attacks, we envision a further improvement when combining with adversarial training.
We compare \name\ against two baselines: we compare against~\citep{madry2017towards}, which is the state-of-the-art in defenses against white-box attacks.  We also compare with the Thermometer technique in~\citep{buckman2018thermometer}, which like \name, combines a preprocessing step with adversarial training. For all compared defenses, adversarial training is done using PGD with 7 steps. We also use BPDA to approximate the gradients during the backward pass. For our comparison we use ResNet-18 and its wide version since they were used in past work on robustness with adversarial training. As for the attacker, we allow it to use the \emph{strongest possible} attack, i.e., it uses BPDA with 1000 PGD attack steps to ensure the results are convergent. Note that previous defenses (including the state-of-the-art) only consider up to 40 steps.

% ==================== CIFAR White Compare (Adv. trained) ===================== %
\begin{table}[!t]
\centering
\scalebox{0.81}{
\begin{tabular}{@{}ccccc@{}}
\toprule
\textbf{Network}            & \textbf{Method} & \textbf{Type}      & \textbf{Steps} & \textbf{Accuracy} \\ \midrule\midrule
\multirow{2.5}{*}{ResNet-18}  & Madry           & Adv. train         & 1000            & 45.0\%            \\ \cmidrule(l){2-5}
                            & \textbf{ME-Net}  & Prep. + Adv. train   & 1000   & \textbf{52.8\%}   \\ \midrule\midrule
\multirow{3.9}{*}{WideResNet} & Madry           & Adv. train         & 1000            & 46.8\%          \\ \cmidrule(l){2-5}  
                            & Thermometer     & Prep. + Adv. train   & 1000            & 12.3\%             \\ \cmidrule(l){2-5} 
                            & \textbf{ME-Net}  & Prep. + Adv. train & 1000   & \textbf{55.1\%}   \\ \bottomrule
\end{tabular}
}
\vspace{-1ex}
\caption{\textbf{White-box attack results for adversarial training}. We use 1000 steps PGD or BPDA attacks in white-box setting to ensure the results are convergent. ME-Net achieves state-of-the-art white-box robustness when combined with adversarial training.}
\label{table:cifar white sota}
\vskip -0.15in
\end{table}
% =========================================================================== %

Table~\ref{table:cifar white sota} summarizes the results.  As shown in the table, \name\ combined with adversarial training outperforms the state-of-the-art results under white-box attacks, achieving a 52.8\% accuracy with ResNet and a 55.1\% accuracy with WideResNet. In contrast, the Thermometer method that also uses preprocessing plus adversarial training cannot survive the strong white-box adversary.

%In Fig.~\ref{fig:cifar white}, we further show a more detailed comparison between \name\ and~\citep{madry2017towards} under different attack steps. With fewer attack steps~(i.e., 7 or 20 steps which current methods use), \name\ can achieve an even higher robustness. However,  a stronger adversary should be considered for truly understanding robustness~\citep{obfuscated-gradients}. As shown, with 1000 attack steps \name\ is verified to achieve the highest robustness under white-box attacks.

\begin{figure*}[!t]
\centering
\subfigure[CIFAR-10]{
    \label{fig:cifar white}
    \includegraphics[height=0.182\textwidth]{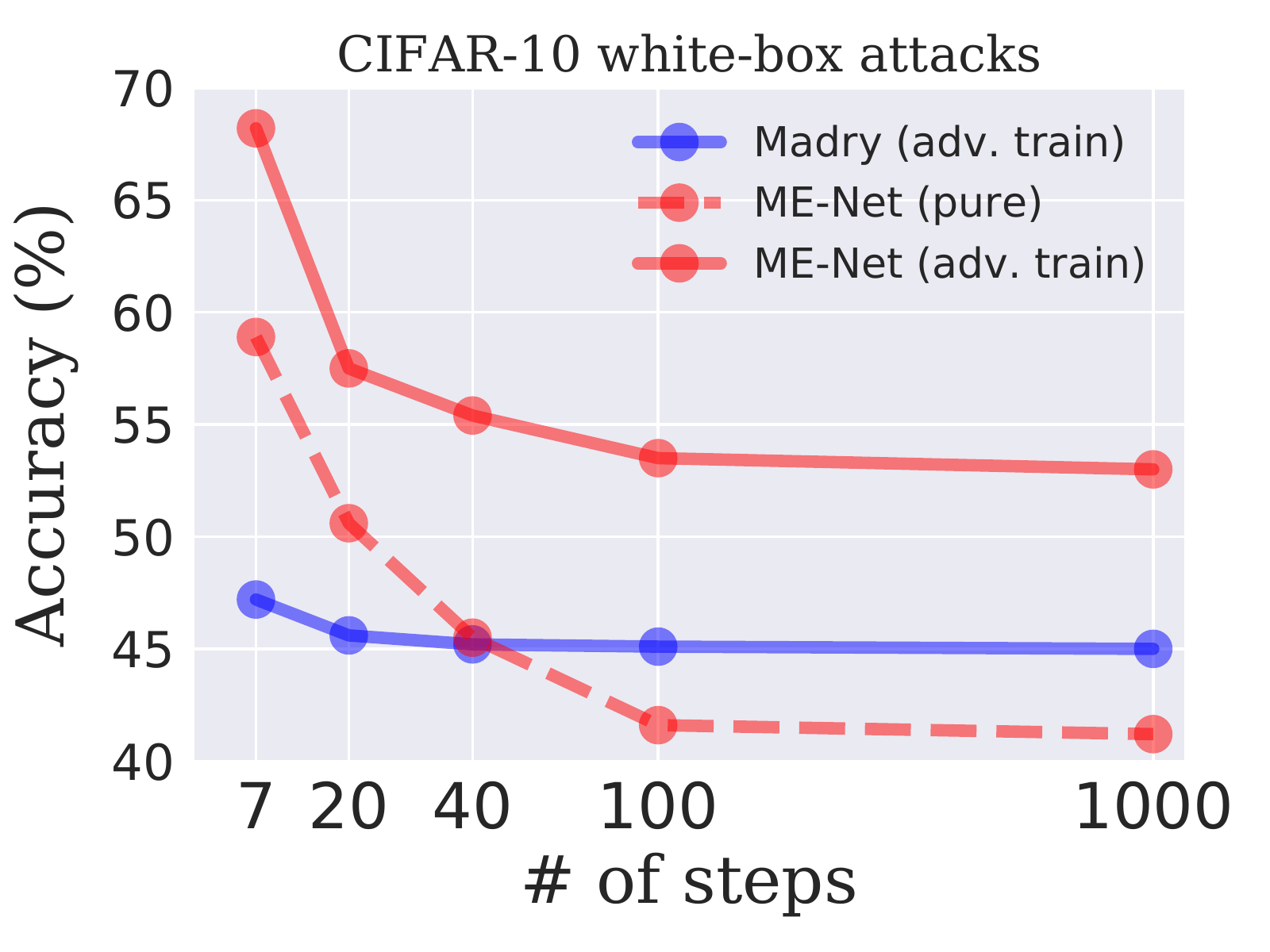}
}
\hspace{-1ex}
\subfigure[MNIST]{
    \label{fig:mnist white}
    \includegraphics[height=0.182\textwidth]{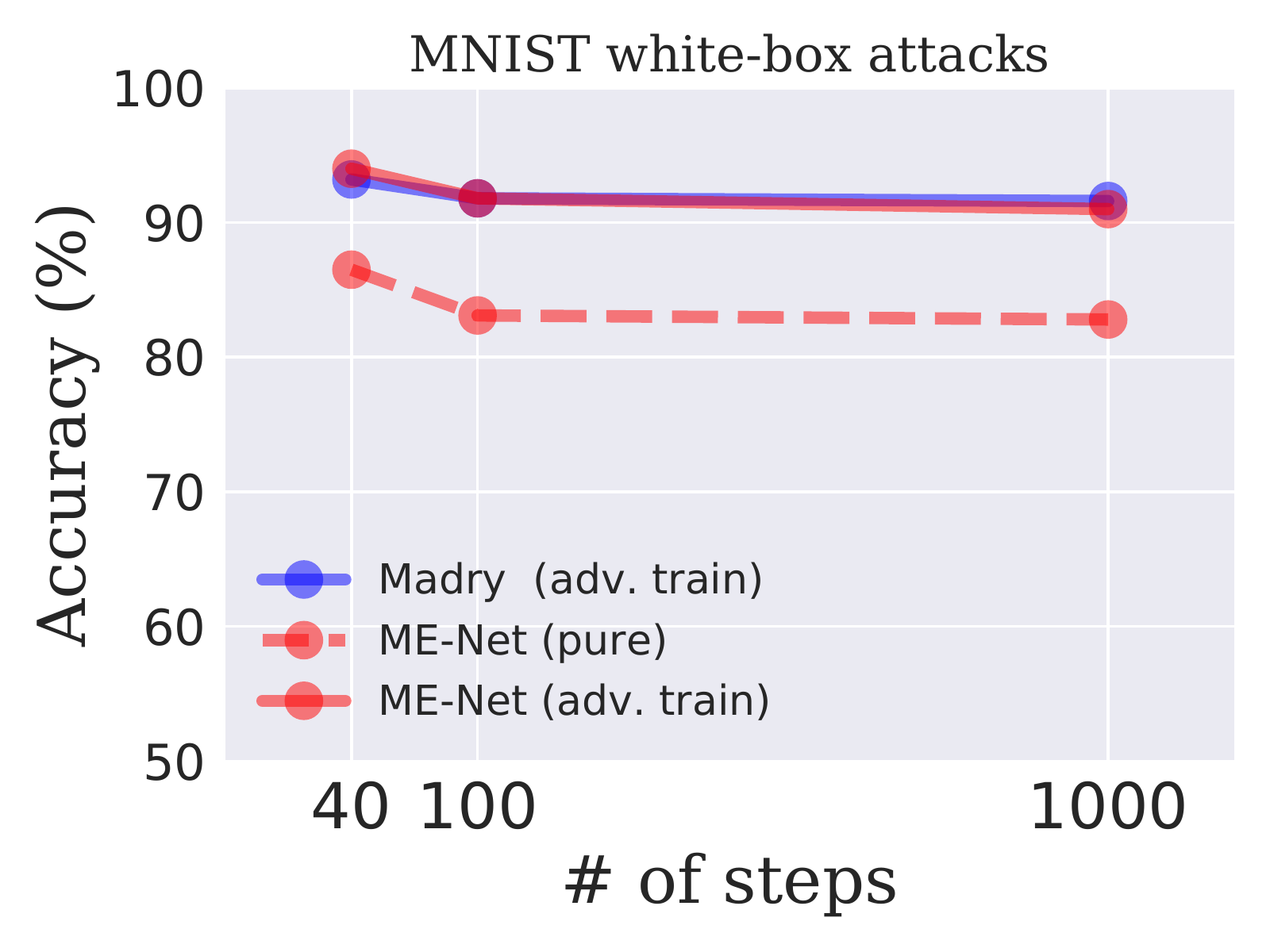}
}
% \hspace{-1ex}
\subfigure[SVHN]{
    \label{fig:svhn white}
    \includegraphics[height=0.182\textwidth]{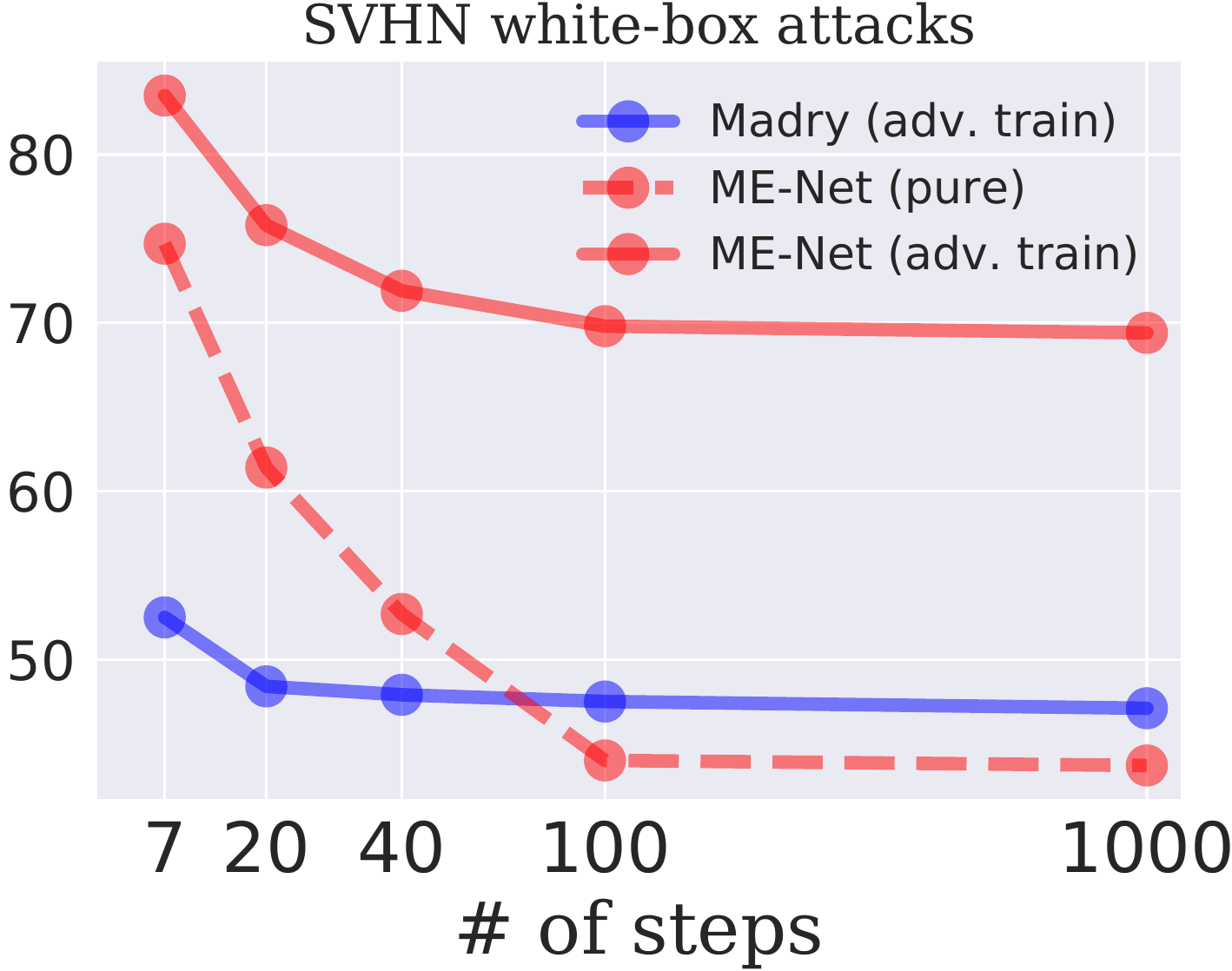}
}
% \hspace{-1ex}
\subfigure[Tiny-ImageNet]{
    \label{fig:timgnet white}
    \includegraphics[height=0.182\textwidth]{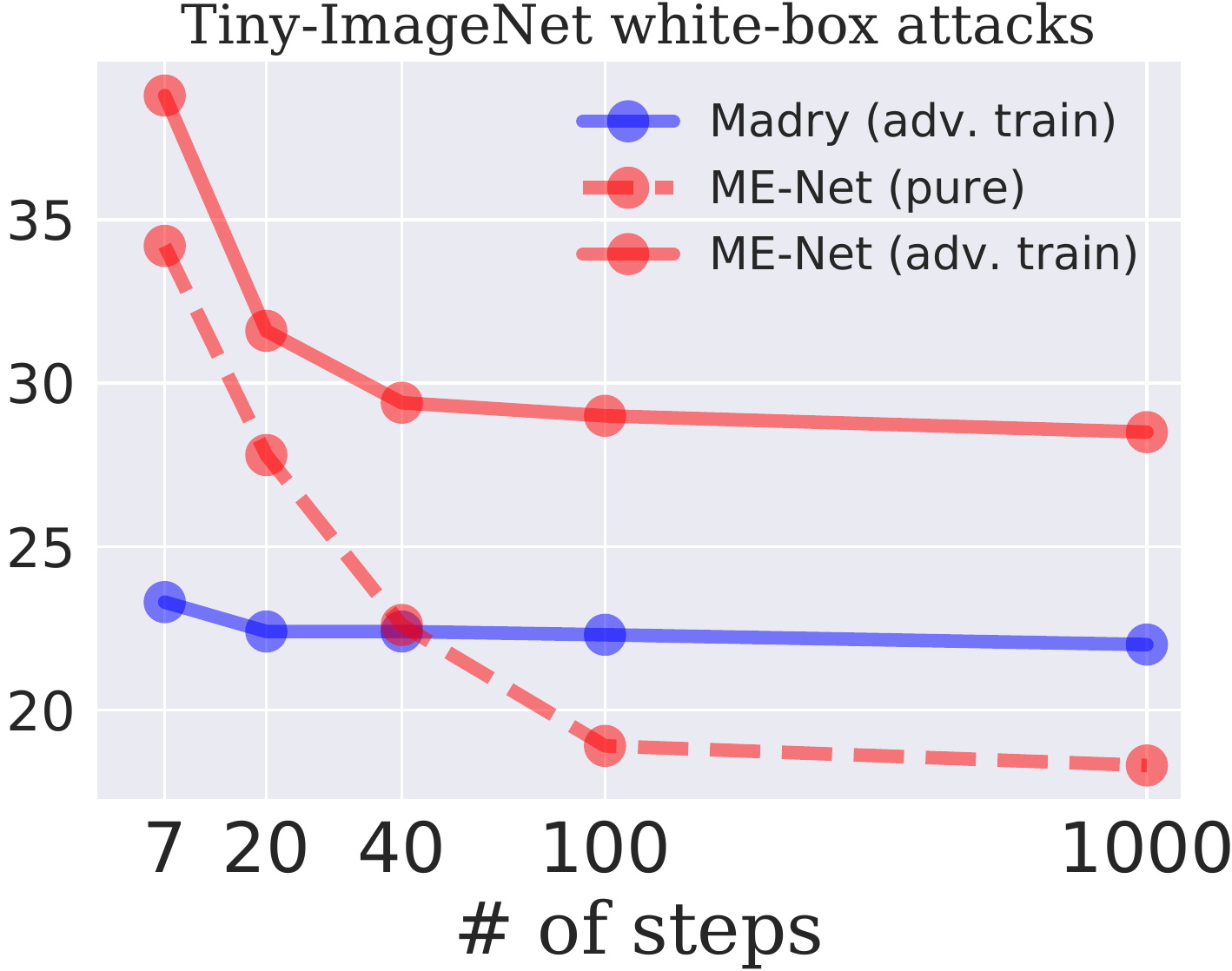}
}
\vspace{-0.3cm}
\caption{\textbf{White-box attack results on different datasets}. We compare ME-Net with \cite{madry2017towards} under PGD or BPDA attack with different attack steps up to 1000. We show both the pure ME-Net without adversarial training, and ME-Net with adversarial training. For Tiny-ImageNet, we report the Top-1 adversarial robustness.}
\label{fig:white box}
\vspace{-0.3cm}
\end{figure*}

\subsection{Evaluation with Different Datasets}
\label{sec:evaluation different datasets}
We evaluate \name\  on MNIST, SVHN, CIFAR-10, and Tiny-ImageNet and compare its performance across these datasets. For space limitations, we present only the results for the white-box attacks.  We provide results for black-box attacks and additional attacks in Appendix~\ref{appendix:cifar}, \ref{appendix:mnist}, \ref{appendix:svhn}, and~\ref{appendix:timgnet}.

For each dataset, we use the network architecture and parameters commonly used in past work on adversarial robustness to help in comparing our results to past work.  For MNIST, we use the LeNet model with two convolutional layers as in~\cite{madry2017towards}. We also use the same attack parameters as total perturbation scale of $76.5/255$~(0.3), and step size $2.55/255$~(0.01). Besides using 40 and 100 total attack steps, we also increase to 1000 steps to further strengthen the adversary. For \name\ with adversarial training, we follow their settings to use 40 steps PGD during training.
We use standard ResNet-18 for SVHN and CIFAR-10, and DenseNet-121 for Tiny-ImageNet, and set attack parameters as follows: total perturbation of $8/255$ (0.031), step size of $2/255$ (0.01), and with up to 1000 total attack steps. Since in~\cite{madry2017towards} the authors did not examine on SVHN and Tiny-ImageNet, we follow their methods to retrain their model on these datasets.  We use 7 steps PGD for adversarial training. We keep all the training hyper-parameters the same for \name\ and~\cite{madry2017towards}.

Fig.~\ref{fig:white box} shows the performance of \name\ on the four datasets and compares it with~\citep{madry2017towards}, a state-of-the-art defense against white-box attacks. %Note that the technique in~\citep{madry2017towards} uses adversarial training we compare it against \name\ with adversarial training.
We plot both the result of a pure version of \name, and \name\ with adversarial training. The figure reveals the following results. First, it shows that  \name\ with adversarial training outperforms the state-of-the-art defense against white-box attacks. 
Interestingly however, the gains differ from one dataset to another. Specifically, \name\ is comparable to~\citep{madry2017towards} on MNIST, provides about 8\% gain on CIFAR-10 and Tiny-ImageNet, and yields 23\% gain on SVHN.

We attribute the differences in accuracy gains across datasets to differences in their properties. 
MNIST is too simple (single channel with small 28$\times$28 pixels), and hence \name\ and~\cite{madry2017towards} both achieve over 90\% accuracy. The other datasets are all more complex and have 3 RGB channels and bigger images. More importantly, Fig.~\ref{fig:dataset cdf} shows that the vast majority of images in SVHN have a very low rank, and hence very strong global structure, which is a property that \name\ leverages to yield an accuracy gain of 23\%. CIFAR-10 and Tiny-ImageNet both have relatively low rank images but not as low as SVHN. The CDF shows that 90\% of the images in CIFAR have a rank lower than 5, whereas 90\% of the images in Tiny-ImageNet have a rank below 10. When taking into account that the dimension of Tiny-ImageNet is twice as CIFAR~(64$\times$64 vs. 32$\times$32), one would expect \name's gain on these datasets to be comparable, which is compatible with the empirical results.

% ========================== Adaptive Attacks ========================== %

\subsection{Evaluation against Adaptive Attacks}
Since \name\ provides a new preprocessing method, we examine customized attacks where the adversary takes advantage of knowing the details of  \name's pipeline. We propose two kinds of white-box attacks:
1) \emph{Approximate input attack}: since \name\ would preprocess the image, this adversary attacks not the original image, but uses the exact preprocess method to approximate/reconstruct an input, and attacks the newly constructed image using the BPDA procedure~\cite{obfuscated-gradients}.
2) \emph{Projected BPDA attack}: since \name\ focuses on the global structure of an image, this adversary aims to attack directly the main structural space of the image. Specifically, it uses BPDA to approximate the gradient, and then projects the gradient to the low-rank space of the image iteratively, i.e., it projects on the space constructed by the top few singular vectors of the original image, to construct the adversarial noise.
Note that these two attacks are based on the BPDA white-box attack which has shown most effective against preprocessing. 
Table~\ref{table:adaptive attacks} shows the results of these attacks, which demonstrates that \name\ is robust to these adaptive white-box attacks.

\begin{table}[!t]
\centering
\scalebox{0.8}{
\begin{tabular}{@{}ccccc@{}}
\toprule
\textbf{Method}                 & \textbf{Training} & \textbf{Steps} & \textbf{Approx. Input} & \textbf{Projected BPDA} \\ \midrule\midrule
\multirow{2.5}{*}{\textbf{\name}} & Pure              & 1000           & \textbf{41.5\%}        & \textbf{64.9\%}         \\ \cmidrule(l){2-5}
                                & Adversarial       & 1000           & \textbf{62.5\%}        & \textbf{74.7\%}         \\ \bottomrule
\end{tabular}
}
\vspace{-1ex}
\caption{{\bf Results of \name\ against adaptive white-box attacks on CIFAR-10.} We use 1000 steps PGD-based BPDA for the two newly proposed attacks, and report the accuracy of ME-Net.}
\label{table:adaptive attacks}
\vskip -0.15in
\end{table}

\subsection{Comparison of Different ME Methods}
Matrix estimation (ME) is a well studied topic with several established ME techniques.  The results in the other sections are with the Nuclear Norm minimization algorithm~\citep{candes2009exact}. Here we compare the performance of three ME methods: the Nuclear Norm minimization algorithm, the Soft-Impute algorithm~\cite{mazumder2010spectral}, and the universal singular value thresholding~(USVT) approach~\cite{chatterjee2015matrix}.  

We train \name\ models using different ME methods on CIFAR-10 with ResNet-18.  We apply transfer-based PGD black-box attacks with 40 attack steps, as well as white-box BPDA attack with 1000 attack steps.
We compare the complexity, generalization and adversarial robustness of these methods. More details can be found in Appendix~\ref{appendix:diff me methods}.

Table~\ref{table:me methods compare} shows the results of our comparison.  The table shows that all the three ME methods are able to improve the original standard generalization, and achieve almost the same test accuracy.  The nuclear norm minimization algorithm takes much longer time and more computation power. The Soft-Impute algorithm simplifies the process but still requires certain computation resources, while the USVT approach is much simpler and faster. The performance of different ME methods is slightly different, as more complex algorithms may gain better performances.

% ========================== ME methods comparison ========================== %
\begin{table}[!t]
\centering
\scalebox{0.78}{
\begin{tabular}{@{}ccccc@{}}
\toprule
\textbf{Method}   & \textbf{Complexity}  & \textbf{Clean}      & \textbf{Black-box} & \textbf{White-box} \\ \midrule\midrule
Vanilla           & $-$                  & 93.4\%              & 0.0\%              & 0.0\%              \\ \midrule\midrule
ME-Net - USVT      & Low                  & \textbf{94.8\%}     & \textbf{89.4\%}    & \textbf{51.9\%}    \\ \midrule
ME-Net - Soft-Imp. & Medium               & \textbf{94.9\%}     & \textbf{91.3\%}    & \textbf{52.3\%}    \\ \midrule
ME-Net - Nuc. Norm & High                 & \textbf{94.8\%}     & \textbf{91.0\%}    & \textbf{52.8\%}    \\ \bottomrule
\end{tabular}
}
\vspace{-1.5ex}
\caption{\textbf{Comparisons between different ME methods.} We report the generalization and adversarial robustness of three ME-Net models using different ME methods on CIFAR-10. We apply transfer-based 40 steps PGD attack as black-box adversary, and 1000 steps PGD-based BPDA as white-box adversary.}
\label{table:me methods compare}
% \vskip -0.15in
\end{table}
% =========================================================================== % 

% =========================================================================== %

\subsection{Improving Generalization}
\label{sec:improving generalization}
As a preprocessing method, \name\ also serves as a data augmentation technique during training. We show that besides adversarial robustness, \name\ can also improve generalization~(i.e., the test accuracy) on clean data. 
We distinguish between two training procedures: 1) non-adversarial training, where the model is trained only with clean data, and 2) adversarial training where the model is trained with adversarial examples. For each case we compare \name\ with the best performing model for that training type. We show results for different datasets, where each dataset is trained with the typical model in past work as stated in Section~\ref{sec:evaluation different datasets}.
Table~\ref{table:generalization clean} shows the results, which demonstrate the benefit of \name\ as a method for improving generalization under both adversarial and non-adversarial training.

% ========================= Standard Generalization ========================= %
\begin{table}[!t]
\centering
\scalebox{0.75}{
\begin{tabular}{@{}cccccc@{}}
\toprule
\textbf{Method} & \textbf{Training}  & \textbf{MNIST} & \textbf{CIFAR-10} & \textbf{SVHN} & \textbf{Tiny-ImageNet} \\ \midrule\midrule
Vanilla                      &  Pure              & 98.8\%                          & 93.4\%                             & 95.0\%                     & 66.4\%           \\ \midrule
\textbf{ME-Net}         & Pure         & \textbf{99.2\%}                 & \textbf{94.9\%}                    & \textbf{96.0\%}            &   \textbf{67.7\%}         \\ \midrule\midrule
Madry                       & Adversarial   & 98.5\%                          & 79.4\%                          & 87.4\%                & 45.6\%                \\ \midrule
\textbf{ME-Net}           & Adversarial           & \textbf{98.8\%}                 & \textbf{85.5\%}                 & \textbf{93.5\%}               & \textbf{57.0\%}              \\ \bottomrule
\end{tabular}
}
\vspace{-1ex}
\caption{\textbf{Generalization performance on clean data.} For each dataset, we use the same network for all the schemes. \name\ improves generalization for both adversarial and non-adversarial training. For Tiny-ImageNet, we report the Top-1 accuracy.} 
\label{table:generalization clean}
\vskip -0.15in
\end{table}
% =========================================================================== %

%%%%%%%%%%%%%%%%%%%%%%%%%%%%%%%%%%%%%%%%%%%%%%%%%%%%%%%%%%%%
%%%%%%%%%%%%%%%%%%%%%%  Discussion  %%%%%%%%%%%%%%%%%%%%%%%%
%%%%%%%%%%%%%%%%%%%%%%%%%%%%%%%%%%%%%%%%%%%%%%%%%%%%%%%%%%%%
%\section{Discussion}
%\label{discuss}
%\input{discuss.tex}

%%%%%%%%%%%%%%%%%%%%%%%%%%%%%%%%%%%%%%%%%%%%%%%%%%%%%%%%%%%%
%%%%%%%%%%%%%%%%%%%%%  Related Works  %%%%%%%%%%%%%%%%%%%%%%
%%%%%%%%%%%%%%%%%%%%%%%%%%%%%%%%%%%%%%%%%%%%%%%%%%%%%%%%%%%%
\section{Related Work}
\label{related work}
\iffalse
\yuzhe{Related Works / Preliminaries / Background? Seems ``Preliminaries'' is better}

\guo{Some review or comments talked about new blackbox method: 
1. https://openreview.net/forum?id=B1xOYoA5tQ
2. https://openreview.net/forum?id=HyM7AiA5YX
3. https://openreview.net/forum?id=S1lIMn05F7
4. https://openreview.net/forum?id=Sk4jFoA9K7
5. https://openreview.net/forum?id=HygUOoC5KX\yuzhe{This link is important, see the comments and rebuttals, good guides for our writings.}
6. https://openreview.net/forum?id=Hygp1nR9FQ}
\fi

% \subsection{Defenses}
Due to the large body of work on adversarial robustness, we focus on methods that are most directly related to our work, and refer readers to the survey~\cite{akhtar2018threat} for a more comprehensive and broad literature review.

\textbf{Adversarial Training.}~~
Currently, the most effective way to defend against adversarial attacks is adversarial training, which trains the model on adversarial examples generated by different kinds of attacks~\cite{madry2017towards,szegedy2013intriguing,goodfellow2015explaining}. Authors of~\cite{madry2017towards} showed that training on adversarial examples generated by PGD with a random start can achieve state-of-the-art performance on MNIST and CIFAR-10 under $\ell_\infty$ constraint.
One major difficulty of adversarial training is that it tends to overfit to the adversarial examples. 
%Moreover, it suffers from the decrease of standard generalization. 
Authors in \cite{schmidt2018adversarially} thus demonstrated and proved that much more data is needed to achieve  
good generalization under adversarial training. \name\ can leverage adversarial training for increased robustness. Further its data augmentation capability helps improving generalization. 

\textbf{Preprocessing.}~~
%\yuzhe{Data Augmentation?}.}~~
Many defenses preprocess the images with a transformation prior to classification.  Typical preprocessing  includes image re-scaling~\cite{xie2017mitigating}, discretization~\cite{chen2018improving}, thermometer encoding~\cite{buckman2018thermometer}, feature squeezing~\cite{xu2017feature}, image quilting~\cite{guo2017countering}, and neural-based transformations~\cite{song2018pixeldefend,defensegan}. These defenses can cause \emph{gradient masking} when using gradient-based attacks. However, as shown in~\cite{obfuscated-gradients}, by applying the Backward Pass Differentiable Approximation (BPDA) attacks designed for obfuscated gradients, the accuracy of all of these methods can be brought to near zero. \name\ is the first preprocessing method that remains effective under the strongest BPDA attack, which could be attributed to its ability to leverage adversarial training. 

% \textbf{Our method.}~~
% We position our method to also be a kind of data preprocessing. Because of different view on the problem, it improves robustness even under the strongest BPDA attack. It works in harmony with adversarial training, improves the robustness of adversarial trained network even further, ont only its weak $\ell_0$, $\ell_2$ performance \cite{}, but also its STOA $\ell_\infty$ performance.

%\textbf{Network Structure.}~~\yuzhe{related papers?}
%There are also some work trying to do robust network by changing the structure of NN \cite{}. Their main method is to introduce transformation or randomness that make the gradient of network to be random or missing so that it is hard for the attacker to generate the right adversarial examples. However, these method also does not survive under BPDA white box attack.

\textbf{Matrix Estimation.}~~
Matrix estimation recovers a data matrix from noisy and incomplete samples of its entries. A classical application is  recommendation systems, such as the Netflix problem \cite{netflix_win}, but it also has richer connections to other learning challenges such as graphon estimation~\cite{airoldi2013stochastic,borgs2017thy}, community detection~\cite{abbe2015recovering,abbe2015community} and time series analysis~\cite{agarwal2018model}.  Many efficient algorithms exist such as the universal singular value thresholding approach~\cite{chatterjee2015matrix}, the convex nuclear norm minimization formulation~\cite{candes2009exact} and even non-convex methods~\cite{jain2013low,chen2015fast,ge2016matrix}. The key promise is that as long as there are some structures underlying the data matrix, such as being low-rank, then exact or approximate recovery can be guaranteed.
As such, ME is an ideal reconstruction scheme for recovering global structures.

%%%%%%%%%%%%%%%%%%%%%%%%%%%%%%%%%%%%%%%%%%%%%%%%%%%%%%%%%%%%
%%%%%%%%%%%%%%%%%%%%%%  Conclusion  %%%%%%%%%%%%%%%%%%%%%%%%
%%%%%%%%%%%%%%%%%%%%%%%%%%%%%%%%%%%%%%%%%%%%%%%%%%%%%%%%%%%%
\section{Conclusion}
\label{conclusion}
We introduced \name, which leverages matrix estimation to improve the robustness to adversarial attacks. Extensive experiments under strong black-box and white-box attacks demonstrated the significance of \name, where it consistently improves the state-of-the-art robustness in different benchmark datasets.
Furthermore, \name\ can easily be embedded into existing networks, and can also bring additional benefits such as improving standard generalization.

%%%%%%%%%%%%%%%%%%%%%%%%%%%%%%%%%%%%%%%%%%%%%%%%%%%%%%%%%%%%
%%%%%%%%%%%%%%%%%%%%  Acknowledgement  %%%%%%%%%%%%%%%%%%%%%
%%%%%%%%%%%%%%%%%%%%%%%%%%%%%%%%%%%%%%%%%%%%%%%%%%%%%%%%%%%%
\section*{Acknowledgements}
The authors thank the anonymous reviewers for their helpful comments in revising the paper. We are grateful to the members of NETMIT and CSAIL for their insightful discussions and supports. Zhi Xu is supported by the Siemens FutureMakers Fellowship.

%%%%%%%%%%%%%%%%%%%%%%%%%%%%%%%%%%%%%%%%%%%%%%%%%%%%%%%%%%%%
%%%%%%%%%%%%%%%%%%%%%%%  Reference  %%%%%%%%%%%%%%%%%%%%%%%%
\nocite{lecuyer2018certified}

\bibliography{ourbib}

\begin{thebibliography}{36}
\providecommand{\natexlab}[1]{#1}
\providecommand{\url}[1]{\texttt{#1}}
\expandafter\ifx\csname urlstyle\endcsname\relax
  \providecommand{\doi}[1]{doi: #1}\else
  \providecommand{\doi}{doi: \begingroup \urlstyle{rm}\Url}\fi

\bibitem[Abbe \& Sandon(2015{\natexlab{a}})Abbe and Sandon]{abbe2015community}
Abbe, E. and Sandon, C.
\newblock Community detection in general stochastic block models: Fundamental
  limits and efficient algorithms for recovery.
\newblock In \emph{Foundations of Computer Science (FOCS), 2015 IEEE 56th
  Annual Symposium on}, pp.\  670--688. IEEE, 2015{\natexlab{a}}.

\bibitem[Abbe \& Sandon(2015{\natexlab{b}})Abbe and Sandon]{abbe2015recovering}
Abbe, E. and Sandon, C.
\newblock Recovering communities in the general stochastic block model without
  knowing the parameters.
\newblock In \emph{Advances in neural information processing systems}, pp.\
  676--684, 2015{\natexlab{b}}.

\bibitem[Agarwal et~al.(2018)Agarwal, Amjad, Shah, and Shen]{agarwal2018model}
Agarwal, A., Amjad, M.~J., Shah, D., and Shen, D.
\newblock Model agnostic time series analysis via matrix estimation.
\newblock \emph{ACM SIGMETRICS performance evaluation review}, 2\penalty0 (3),
  2018.

\bibitem[Airoldi et~al.(2013)Airoldi, Costa, and Chan]{airoldi2013stochastic}
Airoldi, E.~M., Costa, T.~B., and Chan, S.~H.
\newblock Stochastic blockmodel approximation of a graphon: Theory and
  consistent estimation.
\newblock In \emph{Advances in Neural Information Processing Systems}, pp.\
  692--700, 2013.

\bibitem[Akhtar \& Mian(2018)Akhtar and Mian]{akhtar2018threat}
Akhtar, N. and Mian, A.
\newblock Threat of adversarial attacks on deep learning in computer vision: A
  survey.
\newblock \emph{arXiv preprint arXiv:1801.00553}, 2018.

\bibitem[Athalye et~al.(2018)Athalye, Carlini, and
  Wagner]{obfuscated-gradients}
Athalye, A., Carlini, N., and Wagner, D.
\newblock Obfuscated gradients give a false sense of security: Circumventing
  defenses to adversarial examples.
\newblock In \emph{Proceedings of the 35th International Conference on Machine
  Learning, {ICML} 2018}, July 2018.
\newblock URL \url{https://arxiv.org/abs/1802.00420}.

\bibitem[Bell \& Koren(2007)Bell and Koren]{netflix_win}
Bell, R.~M. and Koren, Y.
\newblock Lessons from the netflix prize challenge.
\newblock \emph{SIGKDD Explor. Newsl.}, 9\penalty0 (2):\penalty0 75--79,
  December 2007.
\newblock ISSN 1931-0145.
\newblock \doi{10.1145/1345448.1345465}.
\newblock URL \url{http://doi.acm.org/10.1145/1345448.1345465}.

\bibitem[Borgs et~al.(2017)Borgs, Chayes, Lee, and Shah]{borgs2017thy}
Borgs, C., Chayes, J., Lee, C.~E., and Shah, D.
\newblock Thy friend is my friend: Iterative collaborative filtering for sparse
  matrix estimation.
\newblock In \emph{Advances in Neural Information Processing Systems}, pp.\
  4715--4726, 2017.

\bibitem[Brendel et~al.(2017)Brendel, Rauber, and Bethge]{brendel2017decision}
Brendel, W., Rauber, J., and Bethge, M.
\newblock Decision-based adversarial attacks: Reliable attacks against
  black-box machine learning models.
\newblock \emph{arXiv preprint arXiv:1712.04248}, 2017.

\bibitem[Buckman et~al.(2018)Buckman, Roy, Raffel, and
  Goodfellow]{buckman2018thermometer}
Buckman, J., Roy, A., Raffel, C., and Goodfellow, I.
\newblock Thermometer encoding: One hot way to resist adversarial examples.
\newblock 2018.
\newblock URL \url{https://openreview.net/pdf?id=S18Su--CW}.

\bibitem[Cand{\`e}s \& Recht(2009)Cand{\`e}s and Recht]{candes2009exact}
Cand{\`e}s, E.~J. and Recht, B.
\newblock Exact matrix completion via convex optimization.
\newblock \emph{Foundations of Computational mathematics}, 9\penalty0
  (6):\penalty0 717, 2009.

\bibitem[Carlini \& Wagner(2017)Carlini and Wagner]{carlini2017towards}
Carlini, N. and Wagner, D.
\newblock Towards evaluating the robustness of neural networks.
\newblock In \emph{2017 IEEE Symposium on Security and Privacy (SP)}, pp.\
  39--57. IEEE, 2017.

\bibitem[Chatterjee et~al.(2015)]{chatterjee2015matrix}
Chatterjee, S. et~al.
\newblock Matrix estimation by universal singular value thresholding.
\newblock \emph{The Annals of Statistics}, 43\penalty0 (1):\penalty0 177--214,
  2015.

\bibitem[Chen et~al.(2018)Chen, Wu, Liang, and Jha]{chen2018improving}
Chen, J., Wu, X., Liang, Y., and Jha, S.
\newblock Improving adversarial robustness by data-specific discretization.
\newblock \emph{CoRR}, abs/1805.07816, 2018.

\bibitem[Chen \& Chi(2018)Chen and Chi]{chen2018harnessing}
Chen, Y. and Chi, Y.
\newblock Harnessing structures in big data via guaranteed low-rank matrix
  estimation.
\newblock \emph{arXiv preprint arXiv:1802.08397}, 2018.

\bibitem[Chen \& Wainwright(2015)Chen and Wainwright]{chen2015fast}
Chen, Y. and Wainwright, M.~J.
\newblock Fast low-rank estimation by projected gradient descent: General
  statistical and algorithmic guarantees.
\newblock \emph{arXiv preprint arXiv:1509.03025}, 2015.

\bibitem[Davenport \& Romberg(2016)Davenport and
  Romberg]{davenport2016overview}
Davenport, M.~A. and Romberg, J.
\newblock An overview of low-rank matrix recovery from incomplete observations.
\newblock \emph{arXiv preprint arXiv:1601.06422}, 2016.

\bibitem[Ge et~al.(2016)Ge, Lee, and Ma]{ge2016matrix}
Ge, R., Lee, J.~D., and Ma, T.
\newblock Matrix completion has no spurious local minimum.
\newblock In \emph{Advances in Neural Information Processing Systems}, pp.\
  2973--2981, 2016.

\bibitem[Goodfellow et~al.(2015)Goodfellow, Shlens, and
  Szegedy]{goodfellow2015explaining}
Goodfellow, I., Shlens, J., and Szegedy, C.
\newblock Explaining and harnessing adversarial examples.
\newblock In \emph{International Conference on Learning Representations}, 2015.
\newblock URL \url{http://arxiv.org/abs/1412.6572}.

\bibitem[Guo et~al.(2017)Guo, Rana, Cisse, and van~der
  Maaten]{guo2017countering}
Guo, C., Rana, M., Cisse, M., and van~der Maaten, L.
\newblock Countering adversarial images using input transformations.
\newblock \emph{arXiv preprint arXiv:1711.00117}, 2017.

\bibitem[He et~al.(2016)He, Zhang, Ren, and Sun]{He_2016_CVPR}
He, K., Zhang, X., Ren, S., and Sun, J.
\newblock Deep residual learning for image recognition.
\newblock In \emph{The IEEE Conference on Computer Vision and Pattern
  Recognition (CVPR)}, June 2016.

\bibitem[Huang et~al.(2017)Huang, Liu, van~der Maaten, and
  Weinberger]{huang2017densely}
Huang, G., Liu, Z., van~der Maaten, L., and Weinberger, K.~Q.
\newblock Densely connected convolutional networks.
\newblock In \emph{Proceedings of the IEEE Conference on Computer Vision and
  Pattern Recognition}, 2017.

\bibitem[Jain et~al.(2013)Jain, Netrapalli, and Sanghavi]{jain2013low}
Jain, P., Netrapalli, P., and Sanghavi, S.
\newblock Low-rank matrix completion using alternating minimization.
\newblock In \emph{Proceedings of the forty-fifth annual ACM symposium on
  Theory of computing}, pp.\  665--674. ACM, 2013.

\bibitem[Keshavan et~al.(2010)Keshavan, Montanari, and Oh]{keshavan2010matrix}
Keshavan, R.~H., Montanari, A., and Oh, S.
\newblock Matrix completion from noisy entries.
\newblock \emph{Journal of Machine Learning Research}, 11\penalty0
  (Jul):\penalty0 2057--2078, 2010.

\bibitem[Lecuyer et~al.(2018)Lecuyer, Atlidakis, Geambasu, Hsu, and
  Jana]{lecuyer2018certified}
Lecuyer, M., Atlidakis, V., Geambasu, R., Hsu, D., and Jana, S.
\newblock Certified robustness to adversarial examples with differential
  privacy.
\newblock \emph{arXiv preprint arXiv:1802.03471}, 2018.

\bibitem[Maaten \& Hinton(2008)Maaten and Hinton]{maaten2008visualizing}
Maaten, L. v.~d. and Hinton, G.
\newblock Visualizing data using t-sne.
\newblock \emph{Journal of machine learning research}, 9\penalty0
  (Nov):\penalty0 2579--2605, 2008.

\bibitem[Madry et~al.(2017)Madry, Makelov, Schmidt, Tsipras, and
  Vladu]{madry2017towards}
Madry, A., Makelov, A., Schmidt, L., Tsipras, D., and Vladu, A.
\newblock Towards deep learning models resistant to adversarial attacks.
\newblock \emph{arXiv preprint arXiv:1706.06083}, 2017.

\bibitem[Mazumder et~al.(2010)Mazumder, Hastie, and
  Tibshirani]{mazumder2010spectral}
Mazumder, R., Hastie, T., and Tibshirani, R.
\newblock Spectral regularization algorithms for learning large incomplete
  matrices.
\newblock \emph{Journal of machine learning research}, 11\penalty0
  (Aug):\penalty0 2287--2322, 2010.

\bibitem[Mosbach et~al.(2018)Mosbach, Andriushchenko, Trost, Hein, and
  Klakow]{mosbach2018logit}
Mosbach, M., Andriushchenko, M., Trost, T., Hein, M., and Klakow, D.
\newblock Logit pairing methods can fool gradient-based attacks.
\newblock 2018.

\bibitem[Samangouei et~al.(2018)Samangouei, Kabkab, and Chellappa]{defensegan}
Samangouei, P., Kabkab, M., and Chellappa, R.
\newblock Defense-gan: Protecting classifiers against adversarial attacks using
  generative models.
\newblock In \emph{International Conference on Learning Representations}, 2018.

\bibitem[Schmidt et~al.(2018)Schmidt, Santurkar, Tsipras, Talwar, and
  Madry]{schmidt2018adversarially}
Schmidt, L., Santurkar, S., Tsipras, D., Talwar, K., and Madry, A.
\newblock Adversarially robust generalization requires more data.
\newblock \emph{NIPS}, 2018.
\newblock URL \url{http://arxiv.org/abs/1804.11285}.

\bibitem[Song et~al.(2018)Song, Kim, Nowozin, Ermon, and
  Kushman]{song2018pixeldefend}
Song, Y., Kim, T., Nowozin, S., Ermon, S., and Kushman, N.
\newblock Pixeldefend: Leveraging generative models to understand and defend
  against adversarial examples.
\newblock In \emph{International Conference on Learning Representations}, 2018.

\bibitem[Szegedy et~al.(2013)Szegedy, Zaremba, Sutskever, Bruna, Erhan,
  Goodfellow, and Fergus]{szegedy2013intriguing}
Szegedy, C., Zaremba, W., Sutskever, I., Bruna, J., Erhan, D., Goodfellow, I.,
  and Fergus, R.
\newblock Intriguing properties of neural networks.
\newblock \emph{arXiv preprint arXiv:1312.6199}, 2013.

\bibitem[Uesato et~al.(2018)Uesato, O'Donoghue, Oord, and
  Kohli]{uesato2018adversarial}
Uesato, J., O'Donoghue, B., Oord, A. v.~d., and Kohli, P.
\newblock Adversarial risk and the dangers of evaluating against weak attacks.
\newblock \emph{arXiv preprint arXiv:1802.05666}, 2018.

\bibitem[Xie et~al.(2018)Xie, Wang, Zhang, Ren, and Yuille]{xie2017mitigating}
Xie, C., Wang, J., Zhang, Z., Ren, Z., and Yuille, A.
\newblock Mitigating adversarial effects through randomization.
\newblock In \emph{International Conference on Learning Representations}, 2018.

\bibitem[Xu et~al.(2017)Xu, Evans, and Qi]{xu2017feature}
Xu, W., Evans, D., and Qi, Y.
\newblock Feature squeezing: Detecting adversarial examples in deep neural
  networks.
\newblock \emph{arXiv preprint arXiv:1704.01155}, 2017.

\end{thebibliography}
\bibliographystyle{icml2019}

%%%%%%%%%%%%%%%%%%%%%%%%%%%%%%%%%%%%%%%%%%%%%%%%%%%%%%%%%%%%
%%%%%%%%%%%%%%%%%%%%%%%  Appendix  %%%%%%%%%%%%%%%%%%%%%%%%%
%%%%%%%%%%%%%%%%%%%%%%%%%%%%%%%%%%%%%%%%%%%%%%%%%%%%%%%%%%%%
\onecolumn
\newpage
\appendix
\begin{center}
\LARGE{\textbf{Supplementary Material}}
\end{center}

\section{Pseudo Code for ME-Net}
\label{appendix:pseudo code}

\begin{algorithm}[!th]
   \caption{ME-Net training \& inference}
   \label{alg:example}
\begin{algorithmic}
   \STATE \texttt{/*}~~\texttt{ME-Net}~~\texttt{Training}~~\texttt{*/}
   \STATE {\bfseries Input:} training set $S=\{ (X_i, y_i) \}_{i=1}^M$, prescribed masking probability $\bm{p}=\{p_1,p_2,\dots,p_n\}$, network $N$
   \FORALL{$X_i\in S$}
   \STATE Randomly sample $n$ masks with probability $\{p_1,p_2,\dots,p_n\}$
   \STATE Generate $n$ masked images $\{X^{(1)}_i, X^{(2)}_i,\dots, X^{(n)}_i\}$
   \STATE Apply ME to obtain reconstructed images $\{\hat{X}^{(1)}_i, \hat{X}^{(2)}_i,\dots, \hat{X}^{(n)}_i\}$
   \STATE Add $\{\hat{X}^{(1)}_i, \hat{X}^{(2)}_i,\dots, \hat{X}^{(n)}_i\}$ into new training set $S'$
   \ENDFOR
   \STATE Randomly initialize network $N$
   \FOR{number of training iterations}
   \STATE Sample a mini-batch $B = \{ (\hat{X}_i, y_i) \}_{i=1}^m$ from $S'$
   \STATE Do one training step of network $N$ using mini-batch $B$
   \ENDFOR
   \STATE
   \STATE \texttt{/*}~~\texttt{ME-Net}~~\texttt{Inference}~~\texttt{*/}
   \STATE {\bfseries Input:} test image $X$, masking probability $\bm{p}=\{p_1,p_2,\dots,p_n\}$ used during training
   \STATE {\bfseries Output:} predicted label $y$
   \STATE Randomly sample one mask with probability $p=\frac{1}{n}\sum_{i=1}^n p_i$
   \STATE Generate masked image and apply ME to reconstruct $\hat{X}$
   \STATE Input $\hat{X}$ to the trained network $N$ to get the predicted label $y$
\end{algorithmic}
\end{algorithm}

\section{Training Details}  % Experimental Settings
\label{appendix:training details}
\textbf{Training settings.}~~
We summarize our training hyper-parameters in Table~\ref{table:appendix training}. We follow the standard data augmentation scheme as in~\cite{He_2016_CVPR} to do zero-padding with 4 pixels on each side, and then random crop back to the original image size. We then randomly flip the images horizontally and normalize them into $[0,1]$.
Note that \name's preprocessing is performed before the training process as in Algorithm \ref{alg:example}.

% =========================================================================== %
\begin{table}[!th]
\centering
%\scalebox{0.95}{
\begin{tabular}{@{}cccccccc@{}}
\toprule
\textbf{Dataset}       & \textbf{Model}                                                  & \textbf{Data Aug.} & \textbf{Optimizer} & \textbf{Momentum} & \textbf{Epochs} & \textbf{LR} & \textbf{LR decay}   \\ \midrule\midrule
\textbf{CIFAR-10}      & \begin{tabular}[c]{@{}c@{}}ResNet-18\\ Wide-ResNet\end{tabular} & $\surd$       & SGD & 0.9 & 200             & 0.1         & step (100, 150)     \\ \midrule
\textbf{MNIST}         & LeNet                                                           & $\times$             & SGD & 0.9 & 200             & 0.01        & step (100, 150) \\ \midrule
\textbf{SVHN}          & ResNet-18                                                       &  $\surd$                & SGD & 0.9 & 200             & 0.01        & step (100, 150) \\ \midrule
\textbf{Tiny-ImageNet} & DenseNet-121                                                    & $\surd$                  & SGD & 0.9 & 90              & 0.1         & step (30, 60)       \\ \bottomrule
\end{tabular}
%}
\vspace{-1ex}
\caption{\textbf{Training details of \name\ on different datasets.} Learning rate is decreased at selected epochs with a step factor of 0.1.}
\label{table:appendix training}
\end{table}
% =========================================================================== %

\textbf{ME-Net details.}~~
As was mentioned in Section \ref{sec:model}, one could either operate on the three RGB channels separately as independent matrices or jointly by concatenating them into one wide matrix. For the former approach, given an image, we can apply the same mask to each channel and then separately run ME to recover the matrix. For the latter approach, the RGB channels are first concatenated along the column dimension to produce a wide matrix, i.e., if each channel is of size $32\times 32$, then the concatenated matrix, $[\textrm{RGB}]$, is of size $32\times 96$. A mask is applied to the wide matrix and the whole matrix is then recovered. This approach is a common, simple method for estimating tensor data. Since this work focuses on structures of the image and channels within an image are closely related, we adopt the latter approach in this paper.

In our experiments, we use the following method to generate masks with different observing probability: for each image, we select $n$ masks in total with observing probability $p$ ranging from $a\to b$.
%with equal interval~($a,b\in [0,1]$). 
We use $n=10$ for most experiments. To provide an example, ``$p:0.6\to 0.8$'' indicates that we select 10 masks in total with observing probability from 0.6 to 0.8 with an equal interval of 0.02, i.e., $0.6,0.62,0.64,\dots$.
% We report a hyper-parameter study for the number of masks $n$ and different probability values in Appendix~\ref{appendix:hyperparam}.
Note that we only use this simple selection scheme for mask generation. We believe further improvement can be achieved with better designed selection schemes, potentially tailored to each image.

\section{Additional Results on CIFAR-10}
\label{appendix:cifar}
\subsection{Black-box Attacks}
We provide additional results of ME-Net against different black-box attacks on CIFAR-10. We first show the complete results using different kinds of black-box attacks, i.e., transfer-based~(FGSM, PGD, CW), decision-based~(Boundary) and score-based~(SPSA) attacks.
For CW attack, we follow the settings in~\cite{madry2017towards} to use different confidence values $\kappa$.
We report ME-Net results with different training settings on Table~\ref{table:appendix cifar black}. Here we use pure ME-Net as a preprocessing method without adversarial training.
% We select the vanilla model, \cite{buckman2018thermometer} and \cite{madry2017towards} as competitors.
As shown, previous defenses only consider limited kinds of black-box attacks. We by contrast show extensive and also advanced experimental results.

% =========================================================================== %
\begin{table}[!th]
\centering
\scalebox{0.95}{
\begin{tabular}{@{}ccccccccccc@{}}
\toprule
\multicolumn{2}{c}{\multirow{2.5}{*}{\textbf{Method}}} & \multirow{2.5}{*}{\textbf{Clean}} & \multirow{2.5}{*}{\textbf{FGSM}} & \multicolumn{3}{c}{\textbf{PGD}}                    & \multicolumn{2}{c}{\textbf{CW}}   & \multirow{2.5}{*}{\textbf{Boundary}} & \multirow{2.5}{*}{\textbf{SPSA}} \\ \cmidrule(lr){5-7}\cmidrule(lr){8-9}
\multicolumn{2}{c}{}                                 &                                 &                                & 7 steps         & 20 steps        & 40 steps        & $\kappa$ = 20      & $\kappa$ = 50      &                                    &                                \\ \midrule\midrule
\multicolumn{2}{c}{Vanilla}                          & 93.4\%                          & 24.8\%                         & 7.6\%           & 1.8\%           & 0.0\%           & 9.3\%           & 8.9\%           & 3.5\%                              & 1.4\%                          \\ \midrule
\multicolumn{2}{c}{Madry}                            & 79.4\%                          & 67.0\%                         & 64.2\%          & $-$             & $-$             & 78.7\%          & $-$             & $-$                                & $-$                            \\ \midrule
\multicolumn{2}{c}{Thermometer}                            & 87.5\%                          & $-$                          & 77.7\%          & $-$             & $-$             & $-$           & $-$             & $-$                                & $-$                            \\ \midrule\midrule
\multirow{3.9}{*}{\textbf{ME-Net}}    & $p: 0.8\to 1$      & \textbf{94.9\%}                 & \textbf{92.2\%}                & \textbf{91.8\%} & \textbf{91.8\%} & \textbf{91.3\%} & \textbf{93.6\%} & \textbf{93.6\%} & \textbf{87.4\%}                    & \textbf{93.0\%}                \\ \cmidrule(l){2-11} 
                & $p: 0.6\to 0.8$    & 92.1\%                          & 85.1\%                         & 84.5\%          & 83.4\%          & 81.8\%          & 89.2\%          & 89.0\%          & 81.8\%                             & 90.9\%                         \\ \cmidrule(l){2-11} 
                & $p: 0.4\to 0.6$    & 89.2\%                          & 75.7\%                         & 74.9\%          & 73.0\%          & 70.9\%          & 82.0\%          & 82.0\%          & 77.5\%                             & 87.1\%                         \\ \bottomrule
\end{tabular}
}
\vspace{-1ex}
\caption{\textbf{CIFAR-10 extensive black-box attack results.} Different kinds of strong black-box attacks are used, including transfer-, decision-, and score-based attacks.}
\label{table:appendix cifar black}
\end{table}
% =========================================================================== %

Further, we define and apply another stronger black-box attack, where we provide the architecture and weights of our trained model to the black-box adversary to make it stronger. This kind of attack is also referred as ``semi-black-box'' or ``gray-box'' attack in some instances, while we still view it as a black-box one. This time the adversary is not aware of the preprocessing layer but has full access to the trained network, and directly performs white-box attacks to the network. We show the results in Table~\ref{table:appendix cifar gray}.

% =========================================================================== %
\begin{table}[!th]
\centering
% \scalebox{0.95}{
\begin{tabular}{@{}cccccccc@{}}
\toprule
\multicolumn{2}{c}{\multirow{2.5}{*}{\textbf{Method}}} & \multirow{2.5}{*}{\textbf{FGSM}} & \multicolumn{3}{c}{\textbf{PGD}}  & \multicolumn{2}{c}{\textbf{CW}} \\ \cmidrule(lr){4-6}\cmidrule(l){7-8}
\multicolumn{2}{c}{}      &    & 7 steps         & 20 steps        & 40 steps        & $\kappa$ = 20      & $\kappa$ = 50    \\ \midrule\midrule
\multirow{3.9}{*}{\textbf{ME-Net}}    & $p: 0.8\to 1$    & \textbf{85.1\%}   & \textbf{84.9\%} & \textbf{84.0\%} & \textbf{82.9\%} & 75.8\% & 75.2\%  \\ \cmidrule(l){2-8} 
                & $p: 0.6\to 0.8$     & 83.2\%                         & 82.8\%          & 81.7\%          & 79.6\%          & 81.5\%          & 76.8\%   \\ \cmidrule(l){2-8} 
                & $p: 0.4\to 0.6$     & 80.5\%                         & 80.2\%          & 79.2\%    & 76.4\%  & \textbf{84.0\%}  & \textbf{77.1\%}   \\ \bottomrule
\end{tabular}
% }
\vspace{-1ex}
\caption{\textbf{CIFAR-10 additional black-box attack results where adversary has limited access to the trained network.} We provide the architecture and weights of our trained model to the black-box adversary to make it stronger.}
\label{table:appendix cifar gray}
\end{table}
% =========================================================================== %

\subsection{White-box Attacks}

\subsubsection{Pure ME-Net}
We first show the extensive white-box attack results with pure ME-Net in Table~\ref{table:appendix cifar white pure}. We use strongest white-box BPDA attack~\cite{obfuscated-gradients} with different attack steps. We select three preprocessing methods~\cite{song2018pixeldefend,buckman2018thermometer,guo2017countering} as competitors. We re-implement the total variation minimization approach~\cite{guo2017countering} and apply the same training settings as ME-Net on CIFAR-10. The experiments are performed under total perturbation $\varepsilon$ of $8/255$ (0.031). By comparison, ME-Net is demonstrated to be the first preprocessing method that is effective under strongest white-box attacks.

% =========================================================================== %
\begin{table}[!ht]
\centering
\setlength{\tabcolsep}{4.5mm}{
\begin{tabular}{@{}ccccccc@{}}
\toprule
\multicolumn{2}{c}{\multirow{2.5}{*}{\textbf{Method}}} & \multirow{2.5}{*}{\textbf{Type}} & \multicolumn{4}{c}{\textbf{Attack Steps}}                             \\ \cmidrule(l){4-7} 
\multicolumn{2}{c}{}                                 &                                & \textbf{7}    & \textbf{20}   & \textbf{40}     & \textbf{100}     \\ \midrule\midrule
\multicolumn{2}{c}{Vanilla}                       & $-$                            & 0.0\%           & 0.0\%           & 0.0\%           & 0.0\%           \\ \midrule
\multicolumn{2}{c}{Thermometer}                      & Prep.                          & $-$             & $-$             & 0.0\%*          & 0.0\%*          \\ \midrule
\multicolumn{2}{c}{PixelDefend}                      & Prep.                          & $-$             & $-$             & $-$             & 9.0\%*          \\ \midrule
\multicolumn{2}{c}{TV Minimization}                      & Prep.                          & 14.7\%         & 5.1\%          & 2.7\%             & 0.4\%          \\ \midrule\midrule
\multirow{6.8}{*}{\textbf{ME-Net}}  & $p: 0.8\to 1$      & Prep.                          & 46.2\%          & 33.2\%          & 26.8\%          & 23.5\%          \\ \cmidrule(l){2-7} 
                                   & $p: 0.7\to 0.9$    & Prep.                          & 50.3\%          & 40.4\%          & 33.7\%          & 29.5\%          \\ \cmidrule(l){2-7} 
                                   & $p: 0.6\to 0.8$    & Prep.                          & 53.0\%          & 45.6\%          & 37.8\%          & 35.1\%          \\ \cmidrule(l){2-7} 
                                   & $p: 0.5\to 0.7$    & Prep.                          & 55.7\%          & 47.3\%          & 38.6\%          & 35.9\%          \\ \cmidrule(l){2-7} 
                                   & $p: 0.4\to 0.6$    & Prep.                          & \textbf{59.8\%} & \textbf{52.6\%} & \textbf{45.5\%} & \textbf{41.6\%} \\ \bottomrule
\end{tabular}
}
\vspace{-1ex}
\caption{\textbf{CIFAR-10 extensive white-box attack results with pure ME-Net.} We use the strongest PGD or BPDA attacks in white-box setting with different attack steps. We compare ME-Net with other pure preprocessing methods~\cite{buckman2018thermometer,song2018pixeldefend,guo2017countering}. We show that ME-Net is the first preprocessing method to be effective under white-box attacks. *Data from~\cite{obfuscated-gradients}.}
\label{table:appendix cifar white pure}
\end{table}
% =========================================================================== %

\begin{figure*}[!ht]
\centering
\subfigure{
    \label{fig:cifar white pure eps2}
    \includegraphics[height=0.243\textwidth]{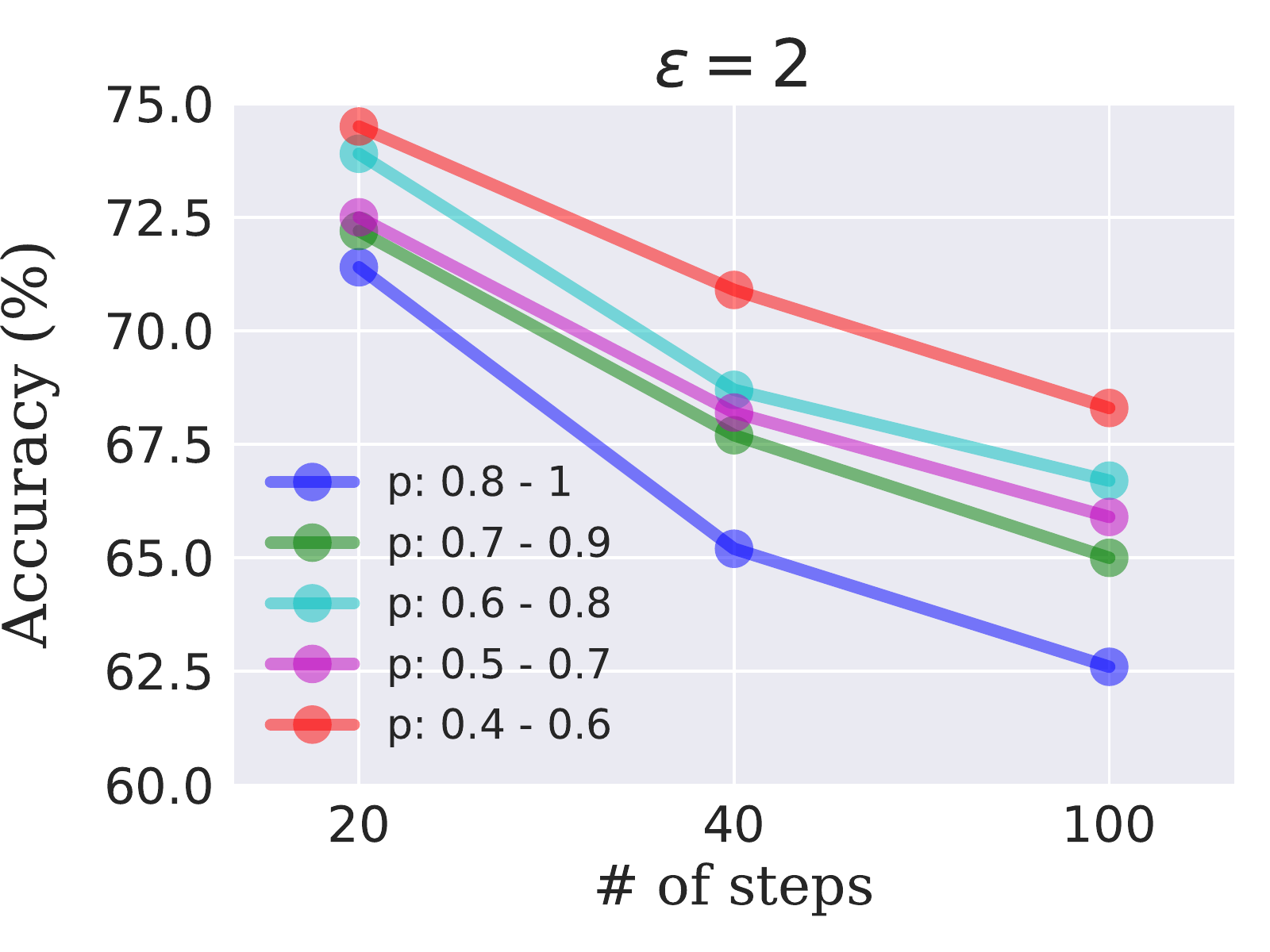}
}
\hspace{-1ex}
\subfigure{
    \label{fig:cifar white pure eps4}
    \includegraphics[height=0.243\textwidth]{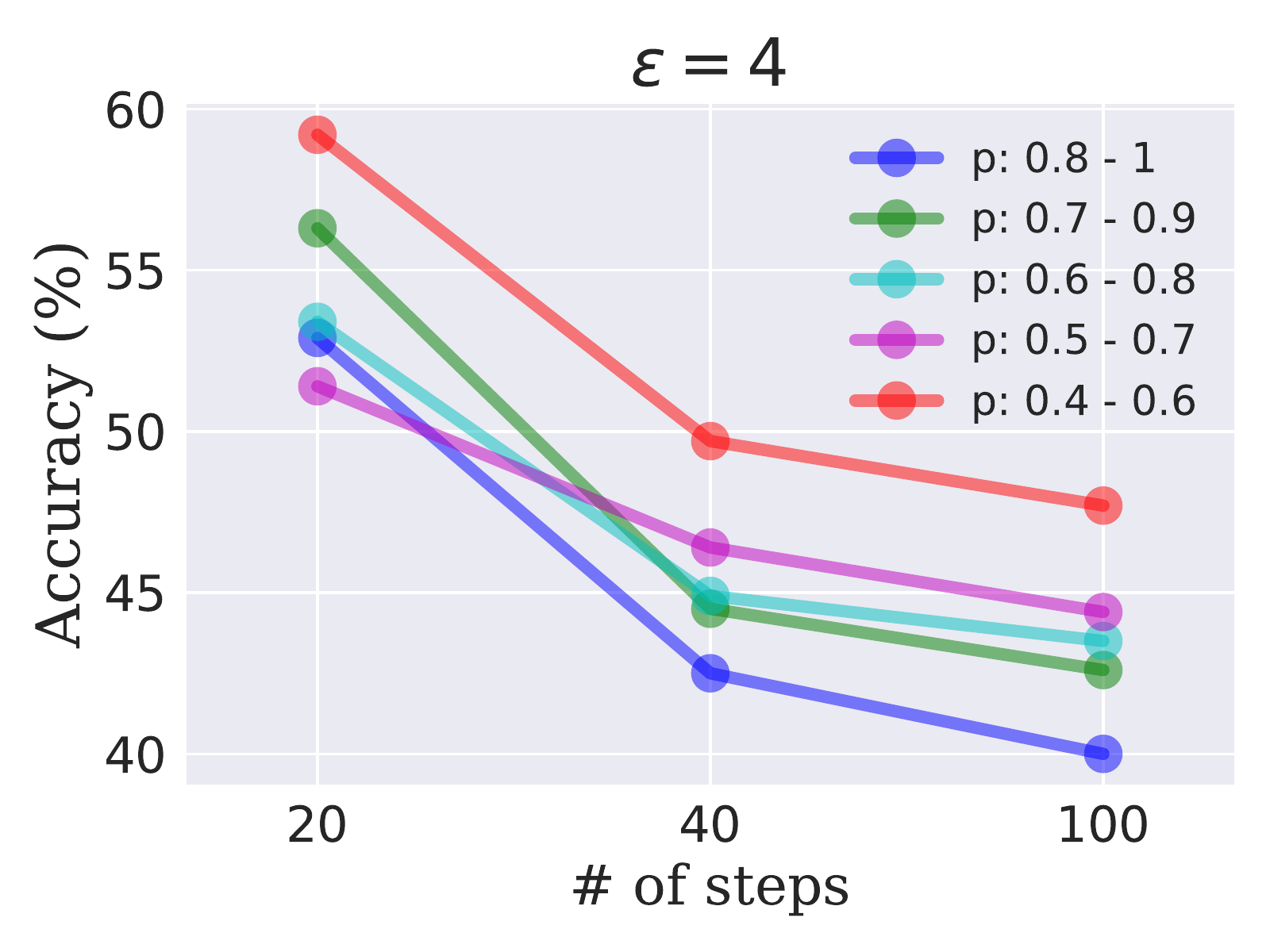}
}
\hspace{-1ex}
\subfigure{
    \label{fig:cifar white pure eps8}
    \includegraphics[height=0.243\textwidth]{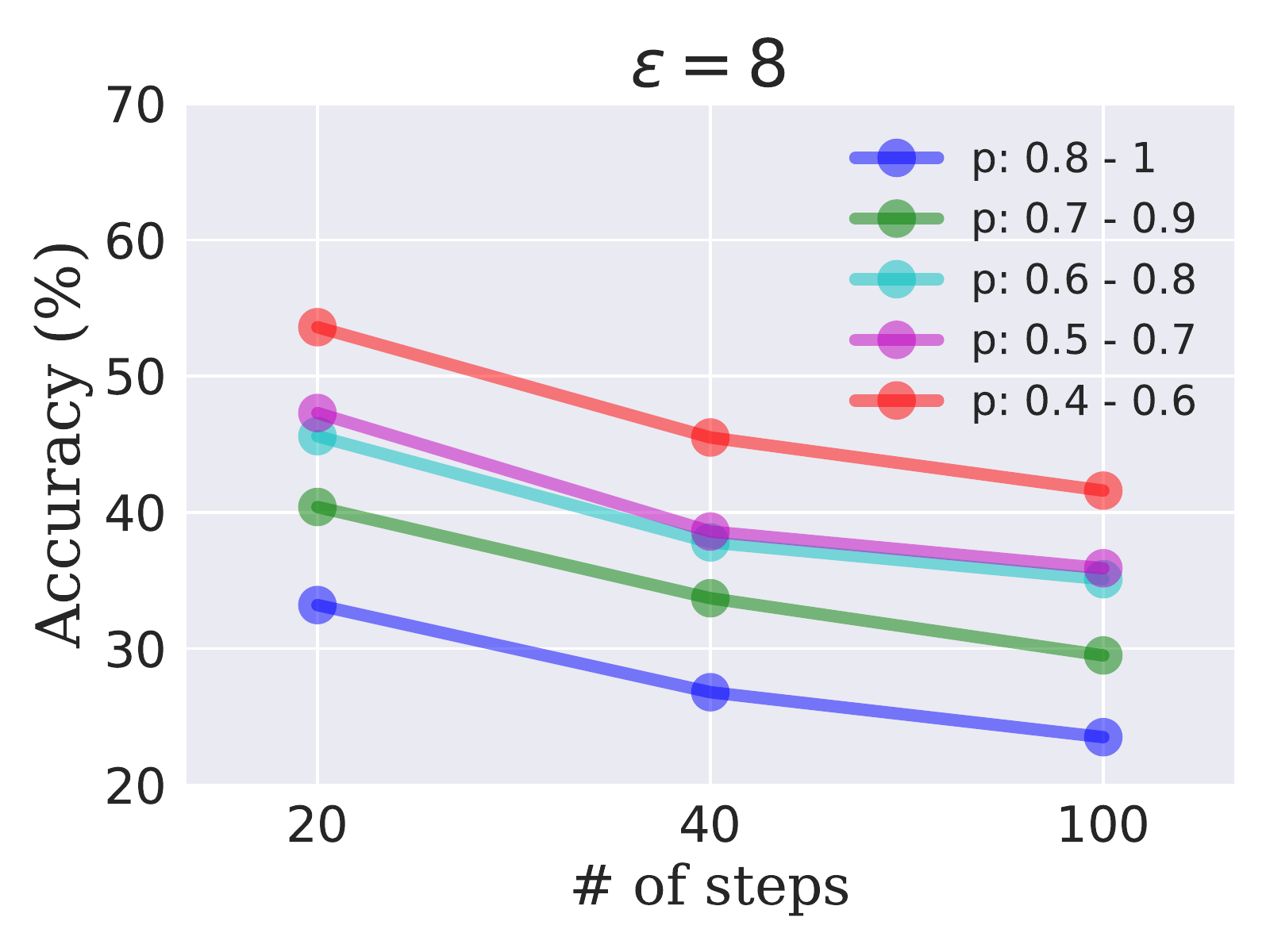}
}
\vspace{-0.6cm}
\caption{\textbf{CIFAR-10 white-box attack results of pure ME-Net with different perturbation $\varepsilon$}. We report ME-Net results with different training settings under various attack steps.}
\label{fig:appendix cifar white pure}
\vspace{-0.3cm}
\end{figure*}

Further, we study the performance of ME-Net under different $\varepsilon$ in Fig.~\ref{fig:appendix cifar white pure}. Besides using $\varepsilon=8$ which is commonly used in CIFAR-10 attack settings~\cite{madry2017towards}, we additionally provide more results including $\varepsilon=2$ and $4$ to study the performance of pure ME-Net under strongest BPDA white-box attacks.

Besides the strongest BPDA attack, we also design and apply another white-box attack to further study the effect of the preprocessing layer. We assume the adversary is aware of the preprocessing layer, but not use the backward gradient approximation. Instead, it performs iterative attacks only for the network part after the preprocessing layer. This attack helps study how the preprocessing affects the network robustness against white-box adversary.
The results in Table~\ref{table:appendix cifar white pure add} shows that pure ME-Net provides sufficient robustness if the white-box adversary does not attack the preprocessing layer.

% =========================================================================== %
\begin{table}[!ht]
\centering
% \scalebox{0.95}{
\begin{tabular}{@{}cccccccc@{}}
\toprule
\multicolumn{2}{c}{\multirow{2.5}{*}{\textbf{Method}}} & \multirow{2.5}{*}{\textbf{FGSM}} & \multicolumn{3}{c}{\textbf{PGD}}  & \multicolumn{2}{c}{\textbf{CW}} \\ \cmidrule(lr){4-6}\cmidrule(l){7-8}
\multicolumn{2}{c}{}      &    & 7 steps         & 20 steps        & 40 steps        & $\kappa$ = 20      & $\kappa$ = 50    \\ \midrule\midrule
\multirow{3.9}{*}{\textbf{ME-Net}}    & $p: 0.8\to 1$    & \textbf{84.3\%}   & \textbf{83.7\%} & \textbf{83.1\%} & \textbf{82.5\%} & 77.0\% & 75.9\%  \\ \cmidrule(l){2-8} 
                & $p: 0.6\to 0.8$     & 82.6\%                         & 82.1\%          & 81.5\%          & 80.3\%          & 76.9\%          & 76.4\%   \\ \cmidrule(l){2-8} 
                & $p: 0.4\to 0.6$     & 79.1\%                         & 79.0\%          & 78.3\%    & 77.4\%  & \textbf{77.5\%}  & \textbf{77.2\%}   \\ \bottomrule
\end{tabular}
% }
\vspace{-1ex}
\caption{\textbf{CIFAR-10 additional white-box attack results where the white-box adversary does not attack the preprocessing layer.} We remain the same attack setups as in the white-box BPDA attack, while only attacking the network part after the preprocessing layer of ME-Net.}
\label{table:appendix cifar white pure add}
\end{table}
% =========================================================================== %

\subsubsection{Combining with Adversarial Training}
We provide more advanced and extensive results of ME-Net when combining with adversarial training in Table~\ref{table:appendix cifar white adv}. As shown, preprocessing methods are not necessarily compatible with adversarial training, as they can perform worse than adversarial training alone~\cite{buckman2018thermometer}. Compared to current state-of-the-art~\cite{madry2017towards}, ME-Net achieves consistently better results under strongest white-box attacks.

% =========================================================================== %
\begin{table}[!ht]
\centering
\scalebox{1.01}{
\begin{tabular}{@{}ccccccccc@{}}
\toprule
\multirow{2.5}{*}{\textbf{Network}} & \multirow{2.5}{*}{\textbf{Method}} & \multirow{2.5}{*}{\textbf{Type}} & \multirow{2.5}{*}{\textbf{Clean}} & \multicolumn{5}{c}{\textbf{Attack Steps}}                                               \\ \cmidrule(l){5-9} 
                                  &                                  &                                &                                 & \textbf{7}      & \textbf{20}     & \textbf{40}     & \textbf{100}    & \textbf{1000}   \\ \midrule\midrule
\multirow{5.4}{*}{ResNet-18}        & Madry                            & Adv. train                     & 79.4\%                          & 47.2\%          & 45.6\%          & 45.2\%          & 45.1\%          & 45.0\%          \\ \cmidrule(l){2-9} 
                                  & \textbf{ME-Net} $p: 0.8\to 1$      & Prep. + Adv. train             & \textbf{85.5\%}                 & 57.4\%          & 51.5\%          & 49.3\%          & 48.1\%          & 47.4\%          \\ \cmidrule(l){2-9} 
                                  & \textbf{ME-Net} $p: 0.6\to 0.8$      & Prep. + Adv. train             & 84.8\%                          & 62.1\%          & 53.0\%          & 51.2\%          & 50.0\%          & 49.6\%          \\ \cmidrule(l){2-9} 
                                  & \textbf{ME-Net} $p: 0.4\to 0.6$      & Prep. + Adv. train             & 84.0\%                          & \textbf{68.2\%} & \textbf{57.5\%} & \textbf{55.4\%} & \textbf{53.5\%} & \textbf{52.8\%} \\ \midrule\midrule
\multirow{5.3}{*}{Wide-ResNet}      & Madry                            & Adv. train                     & 87.3\%                          & 50.0\%          & 47.1\%          & 47.0\%          & 46.9\%          & 46.8\%          \\ \cmidrule(l){2-9} 
                                  & Thermometer                      & Prep. + Adv. train             &  89.9\%                 & 59.4\%          & 34.9\%          & 26.0\%          & 18.4\%          & 12.3\%          \\ \cmidrule(l){2-9} 
                                  & \textbf{ME-Net} $p: 0.6\to 0.8$      & Prep. + Adv. train             & \textbf{91.0\%}                          & 69.7\%          & 58.0\%          & 54.9\%          & 53.4\%          & 52.9\%          \\ \cmidrule(l){2-9} 
                                  & \textbf{ME-Net} $p: 0.4\to 0.6$      & Prep. + Adv. train             &  88.7\%                 & \textbf{74.1\%} & \textbf{61.6\%} & \textbf{57.4\%} & \textbf{55.9\%} & \textbf{55.1\%} \\ \bottomrule
\end{tabular}
}
\vspace{-1ex}
\caption{\textbf{CIFAR-10 extensive white-box attack results.} We apply up to 1000 steps PGD or BPDA attacks in white-box setting to ensure the results are convergent. We use the released models in~\cite{madry2017towards,obfuscated-gradients} but change the attack steps up to 1000 for comparison. ME-Net shows significant advanced results by consistently outperforming the current state-of-the-art defense method~\cite{madry2017towards}.}
\label{table:appendix cifar white adv}
\end{table}
% =========================================================================== %

\section{Additional Results on MNIST}
\label{appendix:mnist}
\subsection{Black-box Attacks}
In Table~\ref{table:appendix mnist black}, we report extensive results of ME-Net under different strong black-box attacks on MNIST. We follow~\cite{madry2017towards} to use the same LeNet model and the same attack parameters with a total perturbation scale of $76.5/255$~(0.3). We use a step size of $2.55/255$ (0.01) for PGD attacks. We use the same settings as in CIFAR-10 for Boundary and SPSA attacks~(i.e., 1000 steps for Boundary attack, and a batch size of 2048 for SPSA attack) to make them stronger.
Note that we only use the \emph{strongest} transfer-based attacks, i.e., we use \emph{white-box} attacks on the independently trained copy to generate black-box examples. As shown, ME-Net shows significantly more effective results against different strongest black-box attacks.

% =========================================================================== %
\begin{table}[!ht]
\centering
% \scalebox{0.95}{
\begin{tabular}{@{}cccccccccc@{}}
\toprule
\multicolumn{2}{c}{\multirow{2.5}{*}{\textbf{Method}}} & \multirow{2.5}{*}{\textbf{Clean}} & \multirow{2.5}{*}{\textbf{FGSM}} & \multicolumn{2}{c}{\textbf{PGD}}  & \multicolumn{2}{c}{\textbf{CW}}   & \multirow{2.5}{*}{\textbf{Boundary}} & \multirow{2.5}{*}{\textbf{SPSA}} \\ \cmidrule(lr){5-6}\cmidrule(lr){7-8}
\multicolumn{2}{c}{}                                 &                                 &                                & 40 steps        & 100 steps       & $\kappa$ = 20      & $\kappa$ = 50      &                                    &                                \\ \midrule\midrule
\multicolumn{2}{c}{Vanilla}                          & 98.8\%                          & 28.2\%                         & 0.1\%           & 0.0\%           & 14.1\%          & 12.6\%          & 3.7\%                              & 6.2\%                          \\ \midrule
\multicolumn{2}{c}{Madry}                            & 98.5\%                          & \textbf{96.8\%}                & \textbf{96.0\%} & \textbf{95.7\%} & 96.4\%          & 97.0\%          & $-$                                & $-$                            \\ \midrule
\multicolumn{2}{c}{Thermometer}                      & 99.0\%                          & $-$                            & 41.1\%          & $-$             & $-$             & $-$             & $-$                                & $-$                            \\ \midrule\midrule
\multirow{5.3}{*}{\textbf{ME-Net}}    & $p: 0.8\to 1$      & \textbf{99.2\%}                 & 77.4\%                         & 73.9\%          & 73.6\%          & \textbf{98.8\%} & \textbf{98.7\%} & \textbf{89.3\%}                    & \textbf{98.1\%}                \\ \cmidrule(l){2-10} 
                                   & $p: 0.6\to 0.8$    & 99.0\%                          & 87.1\%                         & 85.1\%          & 84.9\%          & 98.6\%          & 98.4\%          & 88.6\%                             & 97.5\%                         \\ \cmidrule(l){2-10} 
                                   & $p: 0.4\to 0.6$   & 98.4\%                          & 91.1\%                         & 90.7\%          & 88.9\%          & 98.4\%          & 98.3\%          & 88.0\%                             & 97.0\%                         \\ \cmidrule(l){2-10} 
                                   & $p: 0.2\to 0.4$   & 96.8\%                          & \textbf{93.2\%}                & \textbf{92.8\%} & \textbf{92.2\%} & 96.6\%          & 96.5\%          & 88.1\%                             & 96.1\%                         \\ \bottomrule
\end{tabular}
% }
\vspace{-1ex}
\caption{\textbf{MNIST extensive black-box attack results.} Different kinds of strong black-box attacks are used, including transfer-, decision-, and score-based attacks.}
\label{table:appendix mnist black}
\end{table}
% =========================================================================== %

We further provide the architecture and weights of our trained model to the black-box adversary to make it stronger, and provide the results in Table~\ref{table:appendix mnist gray}. As shown, ME-Net can still maintain high adversarial robustness against stronger black-box adversary under this setting.

% =========================================================================== %
\begin{table}[!ht]
\centering
\setlength{\tabcolsep}{4mm}{
\begin{tabular}{@{}ccccccc@{}}
\toprule
\multicolumn{2}{c}{\multirow{2.5}{*}{\textbf{Method}}} & \multirow{2.5}{*}{\textbf{FGSM}} & \multicolumn{2}{c}{\textbf{PGD}}  & \multicolumn{2}{c}{\textbf{CW}} \\ \cmidrule(lr){4-5}\cmidrule(l){6-7}
\multicolumn{2}{c}{}      &    & 40 steps        & 100 steps        & $\kappa$ = 20      & $\kappa$ = 50    \\ \midrule\midrule
\multirow{5.3}{*}{\textbf{ME-Net}}    & $p: 0.8\to 1$  & 93.0\% &  91.9\% &  85.5\% & \textbf{98.8\%} & \textbf{98.7\%}  \\ \cmidrule(l){2-7} 
                & $p: 0.6\to 0.8$    & 95.0\%  & 94.2\%  & 93.7\%  & 98.3\%  & 98.2\%   \\ \cmidrule(l){2-7}
                & $p: 0.4\to 0.6$    & \textbf{96.2\%}  & \textbf{95.9\%}  & \textbf{95.3\%}  & 98.3\%  & 98.0\%   \\ \cmidrule(l){2-7} 
                & $p: 0.2\to 0.4$    & 94.5\%  & 94.2\%  & 93.4\%  & 96.5\%  & 96.5\%   \\ \bottomrule
\end{tabular}
}
\vspace{-1ex}
\caption{\textbf{MNIST additional black-box attack results where adversary has limited access to the trained network.} We provide the architecture and weights of our trained model to the black-box adversary to make it stronger.}
\label{table:appendix mnist gray}
\end{table}
% =========================================================================== %

\subsection{White-box Attacks}
Table~\ref{table:appendix mnist white} shows the extensive white-box attack results on MNIST. As discussed, we follow~\cite{madry2017towards} to use 40 steps PGD during training when combining ME-Net with adversarial training. We apply up to 1000 steps strong BPDA-based PGD attack to ensure the results are convergent. For the competitor, we use the released model in~\cite{madry2017towards}, but change the attack steps to 1000 for comparison.

% =========================================================================== %
\begin{table}[!ht]
\centering
\setlength{\tabcolsep}{4.3mm}{
\begin{tabular}{@{}ccccccc@{}}
\toprule
\multicolumn{2}{c}{\multirow{2.5}{*}{\textbf{Method}}} & \multirow{2.5}{*}{\textbf{Type}} & \multirow{2.5}{*}{\textbf{Clean}} & \multicolumn{3}{c}{\textbf{Attack Steps}}           \\ \cmidrule(l){5-7} 
\multicolumn{2}{c}{}                                 &                                &                                 & \textbf{40}     & \textbf{100}    & \textbf{1000}   \\ \midrule\midrule
\multicolumn{2}{c}{Madry}                            & Adv. train                     & 98.5\%                          & 93.2\%          & 91.8\%          & \textbf{91.6\%} \\ \midrule\midrule
\multirow{5.3}{*}{\textbf{ME-Net}}    & $p: 0.8\to 1$      & Prep.                          & \textbf{99.2\%}                 & 22.9\%          & 21.8\%          & 18.9\%          \\ \cmidrule(l){2-7} 
                                   & $p: 0.6\to 0.8$    & Prep.                          & 99.0\%                          & 47.6\%          & 42.4\%          & 40.8\%          \\ \cmidrule(l){2-7} 
                                   & $p: 0.4\to 0.6$   & Prep.                          & 98.4\%                          & 65.2\%          & 62.1\%          & 60.6\%          \\ \cmidrule(l){2-7} 
                                   & $p: 0.2\to 0.4$  & Prep.                          & 96.8\%                          & \textbf{86.5\%} & \textbf{83.1\%} & \textbf{82.6\%} \\ \midrule\midrule
\multirow{5.3}{*}{\textbf{ME-Net}}    & $p: 0.8\to 1$      & Prep. + Adv. train             & 97.6\%                          & 87.8\%          & 81.7\%          & 78.0\%          \\ \cmidrule(l){2-7} 
                                   & $p: 0.6\to 0.8$   & Prep. + Adv. train             & 97.7\%                          & 90.5\%          & 88.1\%          & 86.5\%          \\ \cmidrule(l){2-7} 
                                   & $p: 0.4\to 0.6$   & Prep. + Adv. train             & \textbf{98.8\%}                 & 92.1\%          & 89.4\%          & 88.2\%          \\ \cmidrule(l){2-7} 
                                   & $p: 0.2\to 0.4$  & Prep. + Adv. train             & 97.4\%                          & \textbf{94.0\%} & \textbf{91.8\%} & \textbf{91.0\%} \\ \bottomrule
\end{tabular}
}
\vspace{-1ex}
\caption{\textbf{MNIST extensive white-box attack results.} We apply up to 1000 steps PGD or BPDA attacks in white-box setting to ensure the results are convergent. We use the released models in~\cite{madry2017towards} but change the attack steps up to 1000 for comparison. We show both pure ME-Net results and the results when combining with adversarial training.}
\label{table:appendix mnist white}
\end{table}
% =========================================================================== %

\section{Additional Results on SVHN}
\label{appendix:svhn}
\subsection{Black-box Attacks}
Table~\ref{table:appendix svhn black} shows extensive black-box attack results of ME-Net on SVHN. We use standard ResNet-18 as the network, and use a total perturbation of $\varepsilon=8/255$~(0.031). We use the same strong black-box attacks as previously used~(i.e., transfer-, decision-, and score-based attacks), and follow the same attack settings and parameters.
As there are few results on SVHN dataset, we compare only with the vanilla model which uses the same network and training process as ME-Net. 
As shown, ME-Net provides significant adversarial robustness against various black-box attacks.

% =========================================================================== %
\begin{table}[!ht]
\centering
% \scalebox{0.95}{
\begin{tabular}{@{}ccccccccccc@{}}
\toprule
\multicolumn{2}{c}{\multirow{2.5}{*}{\textbf{Method}}} & \multirow{2.5}{*}{\textbf{Clean}} & \multirow{2.5}{*}{\textbf{FGSM}} & \multicolumn{3}{c}{\textbf{PGD}}                    & \multicolumn{2}{c}{\textbf{CW}}   & \multirow{2.5}{*}{\textbf{Boundary}} & \multirow{2.5}{*}{\textbf{SPSA}} \\ \cmidrule(lr){5-7}\cmidrule(lr){8-9}
\multicolumn{2}{c}{}                                 &                                 &                                & 7 steps         & 20 steps        & 40 steps        & $\kappa$ = 20      & $\kappa$ = 50      &                                    &                                \\ \midrule\midrule
\multicolumn{2}{c}{Vanilla}                          & 95.0\%                          & 31.2\%                         & 8.5\%           & 1.8\%           & 0.0\%           & 20.4\%          & 7.6\%           & 4.5\%                              & 3.7\%                          \\ \midrule\midrule
\multirow{5.3}{*}{\textbf{ME-Net}}    & $p: 0.8\to 1$     & \textbf{96.0\%}                 & \textbf{91.8\%}                & \textbf{91.1\%} & \textbf{90.9\%} & \textbf{89.8\%} & \textbf{95.5\%} & \textbf{95.2\%} & 79.2\%                             & \textbf{95.5\%}                \\ \cmidrule(l){2-11} 
                                   & $p: 0.6\to 0.8$  & 95.5\%                          & 88.9\%                         & 88.7\%          & 86.4\%          & 86.2\%          & 95.1\%          & 94.9\%          & 80.6\%                             & 94.6\%                         \\ \cmidrule(l){2-11} 
                                   & $p: 0.4\to 0.6$  & 94.0\%                          & 87.0\%                         & 86.4\%          & 85.8\%          & 84.4\%          & 93.6\%          & 93.4\%          & \textbf{85.3\%}                    & 93.8\%                         \\ \cmidrule(l){2-11} 
                                   & $p: 0.2\to 0.4$   & 88.3\%                          & 80.7\%                         & 76.4\%          & 75.3\%          & 74.2\%          & 87.4\%          & 87.4\%          & 83.3\%                             & 87.6\%                         \\ \bottomrule
\end{tabular}
% }
\vspace{-1ex}
\caption{\textbf{SVHN extensive black-box attack results.} Different kinds of strong black-box attacks are used, including transfer-, decision-, and score-based attacks.}
\label{table:appendix svhn black}
\end{table}
% =========================================================================== %

Again, we strengthen the black-box adversary by providing the network architecture and weights of our trained model. We then apply various attacks and report the results in Table~\ref{table:appendix svhn gray}. ME-Net can still maintain high adversarial robustness under this setting.

% =========================================================================== %
\begin{table}[!ht]
\centering
\setlength{\tabcolsep}{4mm}{
\begin{tabular}{@{}cccccccc@{}}
\toprule
\multicolumn{2}{c}{\multirow{2.5}{*}{\textbf{Method}}} & \multirow{2.5}{*}{\textbf{FGSM}} & \multicolumn{3}{c}{\textbf{PGD}}  & \multicolumn{2}{c}{\textbf{CW}} \\ \cmidrule(lr){4-6}\cmidrule(l){7-8}
\multicolumn{2}{c}{}      &    & 7 steps         & 20 steps        & 40 steps        & $\kappa$ = 20      & $\kappa$ = 50    \\ \midrule\midrule
\multirow{5.3}{*}{\textbf{ME-Net}}    & $p: 0.8\to 1$    & 83.8\%   & 83.3\% & 81.3\% & 78.6\% & \textbf{95.2\%} & \textbf{95.0\%} \\ \cmidrule(l){2-8} 
                & $p: 0.6\to 0.8$     & 85.8\%                         & 85.7\%          & 84.0\%          & 82.1\%          & 94.9\%          & 94.8\%   \\ \cmidrule(l){2-8}
                & $p: 0.4\to 0.6$     & \textbf{88.8\%} & \textbf{88.6\%}          & \textbf{87.4\%}          & \textbf{86.8\%}          & 93.5\%          & 93.3\%   \\ \cmidrule(l){2-8} 
                & $p: 0.2\to 0.4$     & 86.6\%                         & 86.3\%          & 85.7\%    & 85.5\%  & 88.2\%  & 88.2\%   \\ \bottomrule
\end{tabular}
}
\vspace{-1ex}
\caption{\textbf{SVHN additional black-box attack results where adversary has limited access to the trained network.} We provide the architecture and weights of our trained model to the black-box adversary to make it stronger.}
\label{table:appendix svhn gray}
\end{table}
% =========================================================================== %

\subsection{White-box Attacks}
For white-box attacks, we set attack parameters the same as in CIFAR-10, and use strongest white-box BPDA attack with different attack steps~(up to 1000 for convergence). We show results of both pure ME-Net and adversarially trained one. We use 7 steps for adversarial training. Since in~\cite{madry2017towards} the authors did not provide results on SVHN, we follow their methods to retrain a model. The training process and hyper-parameters are kept identical to ME-Net.

Table~\ref{table:appendix svhn white} shows the extensive results under white-box attacks. ME-Net achieves significant adversarial robustness against the strongest  white-box adversary, as it can consistently outperform~\cite{madry2017towards} by a certain margin.

% =========================================================================== %
\begin{table}[!ht]
\centering
\setlength{\tabcolsep}{3mm}{
\begin{tabular}{@{}ccccccccc@{}}
\toprule
\multicolumn{2}{c}{\multirow{2.5}{*}{\textbf{Method}}} & \multirow{2.5}{*}{\textbf{Type}} & \multirow{2.5}{*}{\textbf{Clean}} & \multicolumn{5}{c}{\textbf{Attack Steps}}                                               \\ \cmidrule(l){5-9} 
\multicolumn{2}{c}{}                                 &                                &                                 & \textbf{7}      & \textbf{20}     & \textbf{40}     & \textbf{100}    & \textbf{1000}   \\ \midrule\midrule
\multicolumn{2}{c}{Madry}                            & Adv. train                     & 87.4\%                          & 52.5\%          & 48.4\%          & 47.9\%          & 47.5\%          & 47.1\%          \\ \midrule\midrule
\multirow{5.3}{*}{\textbf{ME-Net}}    & $p: 0.8\to 1$     & Prep.                          & \textbf{96.0\%}                 & 42.1\%          & 27.2\%          & 14.2\%          & 8.0\%           & 7.2\%           \\ \cmidrule(l){2-9} 
                                   & $p: 0.6\to 0.8$   & Prep.                          & 95.5\%                          & 52.4\%          & 39.6\%          & 28.2\%          & 17.1\%          & 15.9\%          \\ \cmidrule(l){2-9} 
                                   & $p: 0.4\to 0.6$   & Prep.                          & 94.0\%                          & 60.3\%          & 48.7\%          & 40.1\%          & 27.4\%          & 25.8\%          \\ \cmidrule(l){2-9} 
                                   & $p: 0.2\to 0.4$   & Prep.                          & 88.3\%                          & \textbf{74.7\%} & \textbf{61.4\%} & \textbf{52.7\%} & \textbf{44.0\%} & \textbf{43.4\%} \\ \midrule\midrule
\multirow{5.3}{*}{\textbf{ME-Net}}    & $p: 0.8\to 1$    & Prep. + Adv. train             & \textbf{93.5\%}                 & 62.2\%          & 41.4\%          & 37.5\%          & 35.5\%          & 34.3\%          \\ \cmidrule(l){2-9} 
                                   & $p: 0.6\to 0.8$   & Prep. + Adv. train             & 92.6\%                          & 72.1\%          & 57.1\%          & 49.6\%          & 47.8\%          & 46.5\%          \\ \cmidrule(l){2-9} 
                                   & $p: 0.4\to 0.6$   & Prep. + Adv. train             & 91.2\%                          & 79.9\% & 69.1\%          & 64.2\%          & 62.3\%          & 61.7\%          \\ \cmidrule(l){2-9} 
                                   & $p: 0.2\to 0.4$   & Prep. + Adv. train             & 87.6\%                          & \textbf{83.5\%}          & \textbf{75.8\%} & \textbf{71.9\%} & \textbf{69.8\%} & \textbf{69.4\%} \\ \bottomrule
\end{tabular}
}
\vspace{-1ex}
\caption{\textbf{SVHN extensive white-box attack results.} We apply up to 1000 steps PGD or BPDA attacks in white-box setting to ensure the results are convergent. We show results of both pure ME-Net and adversarially trained ones. ME-Net shows significantly better results as it consistently outperforms~\cite{madry2017towards} by a certain margin.}
\label{table:appendix svhn white}
\end{table}
% =========================================================================== %

\section{Additional Results on Tiny-ImageNet}
\label{appendix:timgnet}
In this section, we extend our experiments to evaluate ME-Net on a larger and more complex dataset. We use Tiny-ImageNet, which is a subset of ImageNet and contains 200 classes. Each class has 500 images for training and 50 for testing. All images are 64$\times$64 colored ones.
Since ME-Net requires to train the model from scratch, due to the limited computing resources, we do not provide results on even larger dataset such as ImageNet.
However, we envision ME-Net to perform better on such larger datasets as it can leverage the global structures of those larger images.

\subsection{Black-box Attacks}
For black-box attacks on Tiny-ImageNet, we only report the Top-1 adversarial accuracy. We use standard DenseNet-121~\cite{huang2017densely} as our network, and set the attack parameters as having a total perturbation $\varepsilon=8/255$~(0.031). We use the same black-box attacks as before and follow the same attack settings.
The extensive results are shown in Table~\ref{table:appendix timgnet black}.

% =========================================================================== %
\begin{table}[!ht]
\centering
% \scalebox{0.95}{
\begin{tabular}{@{}ccccccccccc@{}}
\toprule
\multicolumn{2}{c}{\multirow{2.5}{*}{\textbf{Method}}} & \multirow{2.5}{*}{\textbf{Clean}} & \multirow{2.5}{*}{\textbf{FGSM}} & \multicolumn{3}{c}{\textbf{PGD}}                    & \multicolumn{2}{c}{\textbf{CW}}   & \multirow{2.5}{*}{\textbf{Boundary}} & \multirow{2.5}{*}{\textbf{SPSA}} \\ \cmidrule(lr){5-7}\cmidrule(lr){8-9}
\multicolumn{2}{c}{}                                 &                                 &                                & 7 steps         & 20 steps        & 40 steps        & $\kappa$ = 20      & $\kappa$ = 50      &                                    &                                \\ \midrule\midrule
\multicolumn{2}{c}{Vanilla}                          & 66.4\%                          & 15.2\%                         & 1.3\%           & 0.0\%           & 0.0\%           & 8.0\%           & 7.7\%           & 2.6\%                              & 1.2\%                          \\ \midrule\midrule
\multirow{4}{*}{\textbf{ME-Net}}    & $p: 0.8\to 1$   & \textbf{67.7\%}                 & \textbf{67.1\%}                & \textbf{66.3\%} & \textbf{66.0\%} & \textbf{65.8\%} & \textbf{67.6\%} & \textbf{67.4\%} & \textbf{62.4\%}                    & \textbf{67.4\%}                \\ \cmidrule(l){2-11} 
                                   & $p: 0.6\to 0.8$  & 64.1\%                          & 63.6\%                         & 63.1\%          & 63.1\%          & 62.4\%          & 63.8\%          & 63.6\%          & 61.9\%                             & 63.8\%                         \\ \cmidrule(l){2-11} 
                                   & $p: 0.4\to 0.6$ & 58.9\%                          & 54.8\%                         & 51.7\%          & 51.6\%          & 50.4\%          & 58.2\%          & 58.2\%          & 58.9\%                             & 58.1\%                         \\ \bottomrule
\end{tabular}
% }
\vspace{-1ex}
\caption{\textbf{Tiny-ImageNet extensive black-box attack results.} Different kinds of strong black-box attacks are used, including transfer-, decision-, and score-based attacks.}
\label{table:appendix timgnet black}
\end{table}
% =========================================================================== %

Further, additional black-box attack results are provided in Table~\ref{table:appendix timgnet gray}, where the black-box adversary has limited access to ME-Net. The results again demonstrate the effectiveness of the preprocessing layer.

% =========================================================================== %
\begin{table}[!ht]
\centering
\setlength{\tabcolsep}{4mm}{
\begin{tabular}{@{}cccccccc@{}}
\toprule
\multicolumn{2}{c}{\multirow{2.5}{*}{\textbf{Method}}} & \multirow{2.5}{*}{\textbf{FGSM}} & \multicolumn{3}{c}{\textbf{PGD}}  & \multicolumn{2}{c}{\textbf{CW}} \\ \cmidrule(lr){4-6}\cmidrule(l){7-8}
\multicolumn{2}{c}{}      &    & 7 steps         & 20 steps        & 40 steps        & $\kappa$ = 20      & $\kappa$ = 50    \\ \midrule\midrule
\multirow{4}{*}{\textbf{ME-Net}}    & $p: 0.8\to 1$    & \textbf{66.5\%}   & \textbf{64.0\%} & \textbf{62.6\%} & 59.1\% & 55.8\% & 56.0\% \\ \cmidrule(l){2-8} 
                & $p: 0.6\to 0.8$     & 61.1\%                         & 60.9\%          & 60.7\%          & \textbf{59.2\%}  & 57.6\%          & 57.6\%   \\ \cmidrule(l){2-8}
                & $p: 0.4\to 0.6$     & 58.8\%                         & 58.2\%          & 57.5\%    & 56.9\%  & \textbf{58.3\%}  & \textbf{58.2\%}   \\ \bottomrule
\end{tabular}
}
\vspace{-1ex}
\caption{\textbf{Tiny-ImageNet additional black-box attack results where adversary has limited access to the trained network.} We provide the architecture and weights of our trained model to the black-box adversary to make it stronger.}
\label{table:appendix timgnet gray}
\end{table}
% =========================================================================== %

\subsection{White-box Attacks}
In white-box settings, we set the attack hyper-parameters as follows: a total perturbation of $8/255$ (0.031), a step size of $2/255$~(0.01), and 7 steps PGD for adversarial training. We still use strongest BPDA attack with different attack steps up to 1000. We re-implement~\cite{madry2017towards} to be the baseline, and keep all training process the same for ME-Net and~\cite{madry2017towards}. Finally, we report both Top-1 and Top-5 adversarial accuracy in Table~\ref{table:appendix timgnet white}, which demonstrates the significant adversarial robustness of ME-Net.

% =========================================================================== %
\begin{table}[!ht]
\centering
\setlength{\tabcolsep}{2.8mm}{
\begin{tabular}{@{}ccccccccc@{}}
\toprule
\multirow{2.5}{*}{\textbf{Metrics}}      & \multirow{2.5}{*}{\textbf{Method}} & \multirow{2.5}{*}{\textbf{Type}} & \multirow{2.5}{*}{\textbf{Clean}} & \multicolumn{5}{c}{\textbf{Attack Steps}}                                               \\ \cmidrule(l){5-9} 
                                &                                  &                                &                                 & \textbf{7}      & \textbf{20}     & \textbf{40}     & \textbf{100}    & \textbf{1000}   \\ \midrule\midrule
\multirow{5.3}{*}{\textbf{Top-1}} & Madry                            & Adv. train                     & 45.6\%                          & 23.3\%          & 22.4\%          & 22.4\%          & 22.3\%          & 22.1\%          \\ \cmidrule(l){2-9} 
                                & \textbf{ME-Net} $p: 0.8\to 1$        & Prep. + Adv. train             & 53.9\%                          & 28.1\%          & 25.7\%          & 25.3\%          & 25.0\%          & 24.5\%          \\ \cmidrule(l){2-9} 
                                & \textbf{ME-Net} $p: 0.6\to 0.8$  & Prep. + Adv. train             & \textbf{57.0\%}                 & 33.7\%          & 28.4\%          & 27.3\%          & 26.8\%          & 26.3\%          \\ \cmidrule(l){2-9} 
                                & \textbf{ME-Net} $p: 0.4\to 0.6$      & Prep. + Adv. train             & 55.6\%                          & \textbf{38.8\%} & \textbf{30.6\%} & \textbf{29.4\%} & \textbf{29.0\%} & \textbf{28.5\%} \\ \midrule\midrule
\multirow{5.3}{*}{\textbf{Top-5}} & Madry                            & Adv. train                     & 71.4\%                          & 47.5\%          & 46.0\%          & 45.9\%          & 45.8\%          & 45.0\%          \\ \cmidrule(l){2-9} 
                                & \textbf{ME-Net} $p: 0.8\to 1$    & Prep. + Adv. train             & 77.4\%                          & 54.8\%          & 52.2\%          & 51.9\%          & 51.2\%          & 50.6\%          \\ \cmidrule(l){2-9} 
                                & \textbf{ME-Net} $p: 0.6\to 0.8$   & Prep. + Adv. train             & \textbf{80.3\%}                 & 62.1\%          & 57.1\%          & 56.7\%          & 56.4\%          & 55.1\%          \\ \cmidrule(l){2-9} 
                                & \textbf{ME-Net} $p: 0.4\to 0.6$   & Prep. + Adv. train             & 78.8\%                          & \textbf{66.7\%} & \textbf{59.5\%} & \textbf{58.5\%} & \textbf{58.0\%} & \textbf{56.9\%} \\ \bottomrule
\end{tabular}
}
\vspace{-1ex}
\caption{\textbf{Tiny-ImageNet extensive white-box attack results.} We apply up to 1000 steps PGD or BPDA attacks in white-box setting to ensure the results are convergent. We select~\cite{madry2017towards} as the baseline and keep the training process the same for both~\cite{madry2017towards} and ME-Net. We show both Top-1 and Top-5 adversarial accuracy under different attack steps. ME-Net shows advanced results by outperforming~\cite{madry2017towards} consistently in both Top-1 and Top-5 adversarial accuracy.}
\label{table:appendix timgnet white}
\end{table}
% =========================================================================== %

\section{Trade-off between Adversarial Robustness and Standard Generalization}
\label{appendix:tradeoff}
In this section, we briefly discuss the trade-off between standard generalization and adversarial robustness, which can be affected by training ME-Net with different hyper-parameters. 
When the masks are generated with higher observing probability $p$, the recovered images will contain more details and are more similar to the original ones. In this case, the generalization ability will be similar to the vanilla network~(or even be enhanced). However, the network will be sensible to the adversarial noises, as the adversarial structure in the noise is only destroyed a bit, and thus induces low robustness.
On the other hand, when given lower observing probability $p$, much of the adversarial structure in the noise will be eliminated, which can greatly increase the adversarial robustness. Nevertheless, the generalization on clean data can decrease as it becomes harder to reconstruct the images and the input images may not be similar to the original ones.
In summary, there exists an inherent trade-off between standard generalization and adversarial robustness.
The trade-off should be further studied to acquire a better understanding and performance of ME-Net.

\begin{figure*}[!ht]
\centering
\subfigure[MNIST]{
    \label{fig:appendix tradeoff mnist}
    \includegraphics[height=0.185\textwidth]{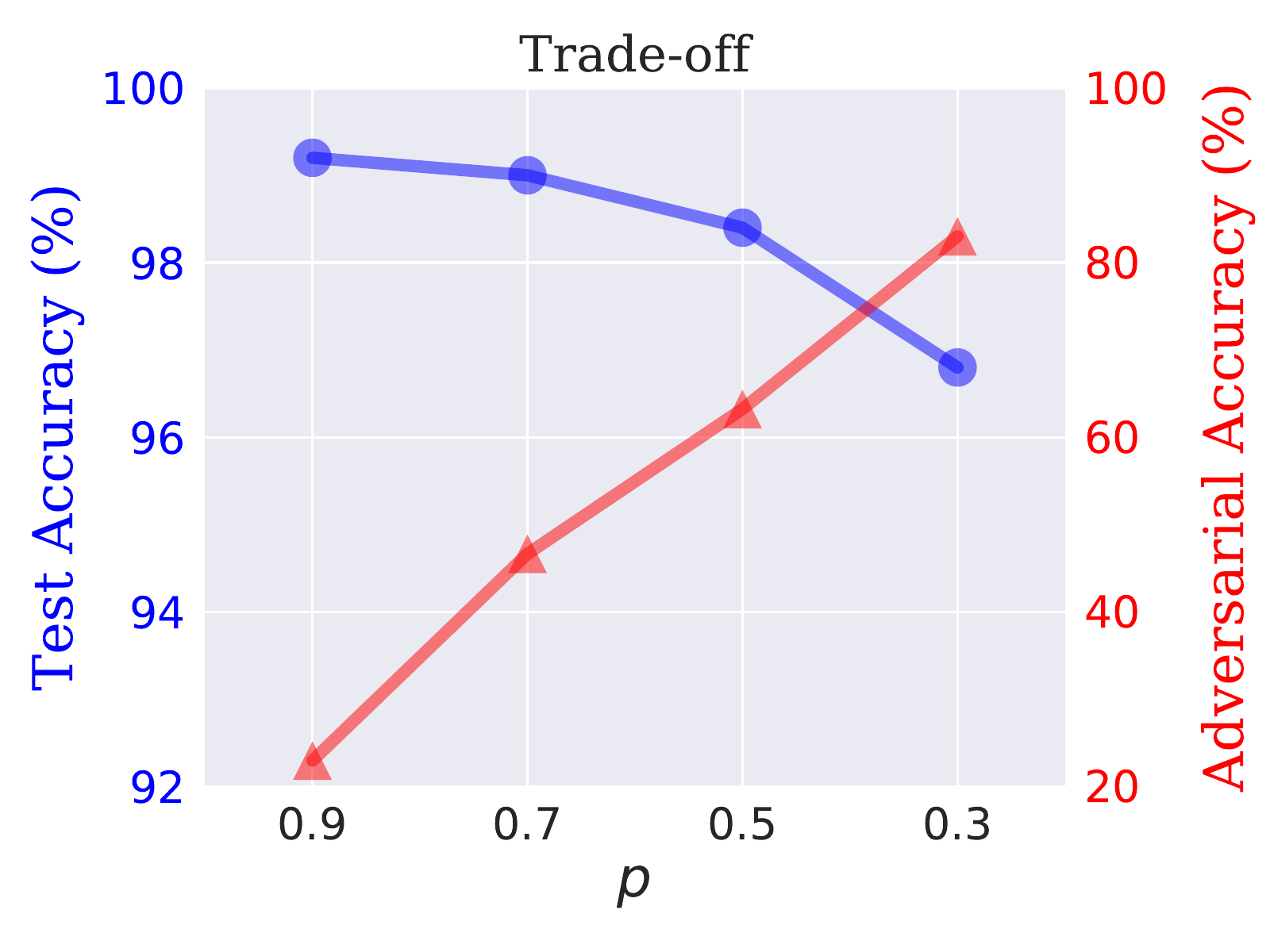}
}
\hspace{-1ex}
\subfigure[CIFAR-10]{
    \label{fig:appendix tradeoff cifar}
    \includegraphics[height=0.185\textwidth]{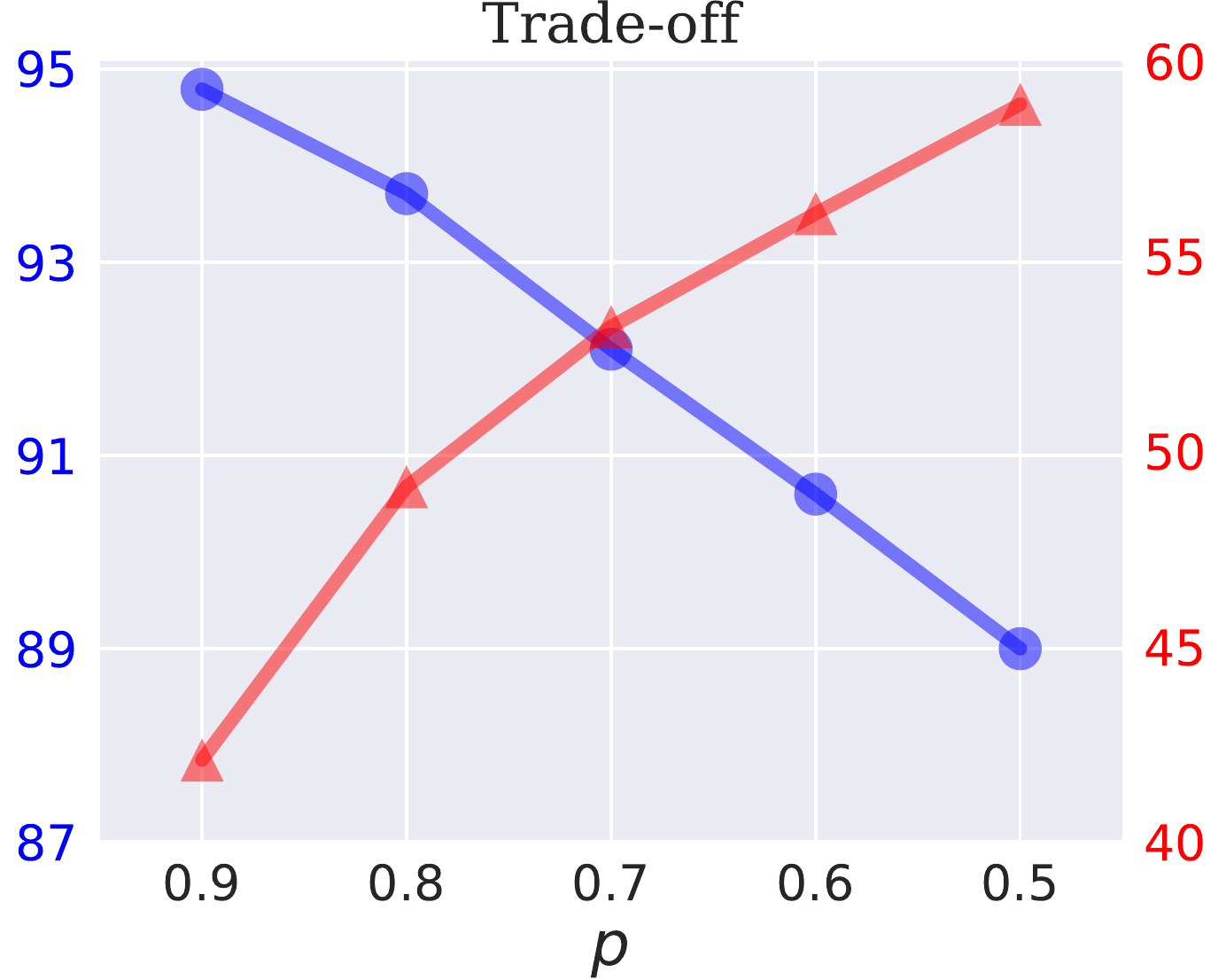}
}
\hspace{-1ex}
\subfigure[SVHN]{
    \label{fig:appendix tradeoff svhn}
    \includegraphics[height=0.185\textwidth]{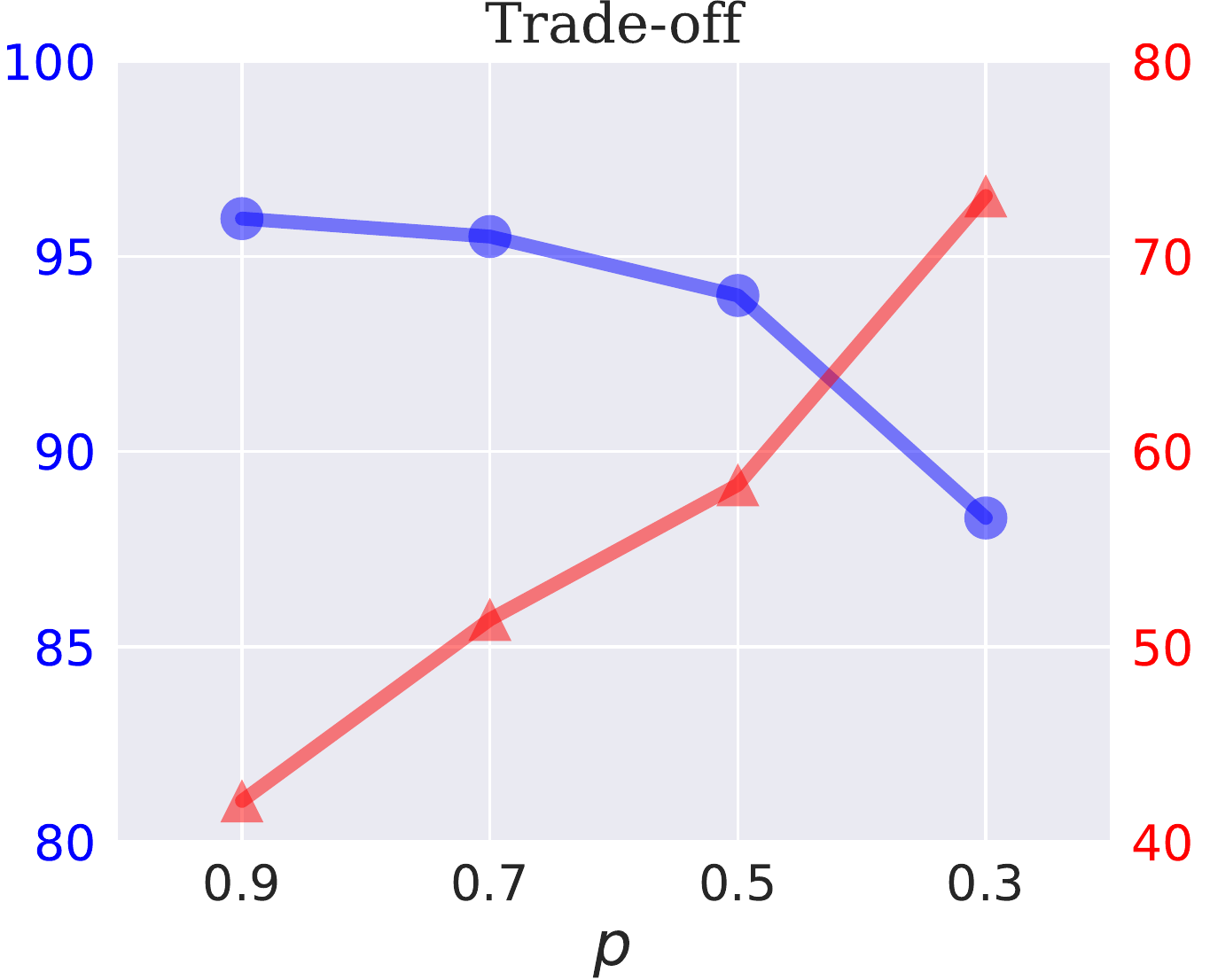}
}
\hspace{-1ex}
\subfigure[Tiny-ImageNet]{
    \label{fig:appendix tradeoff timgnet}
    \includegraphics[height=0.185\textwidth]{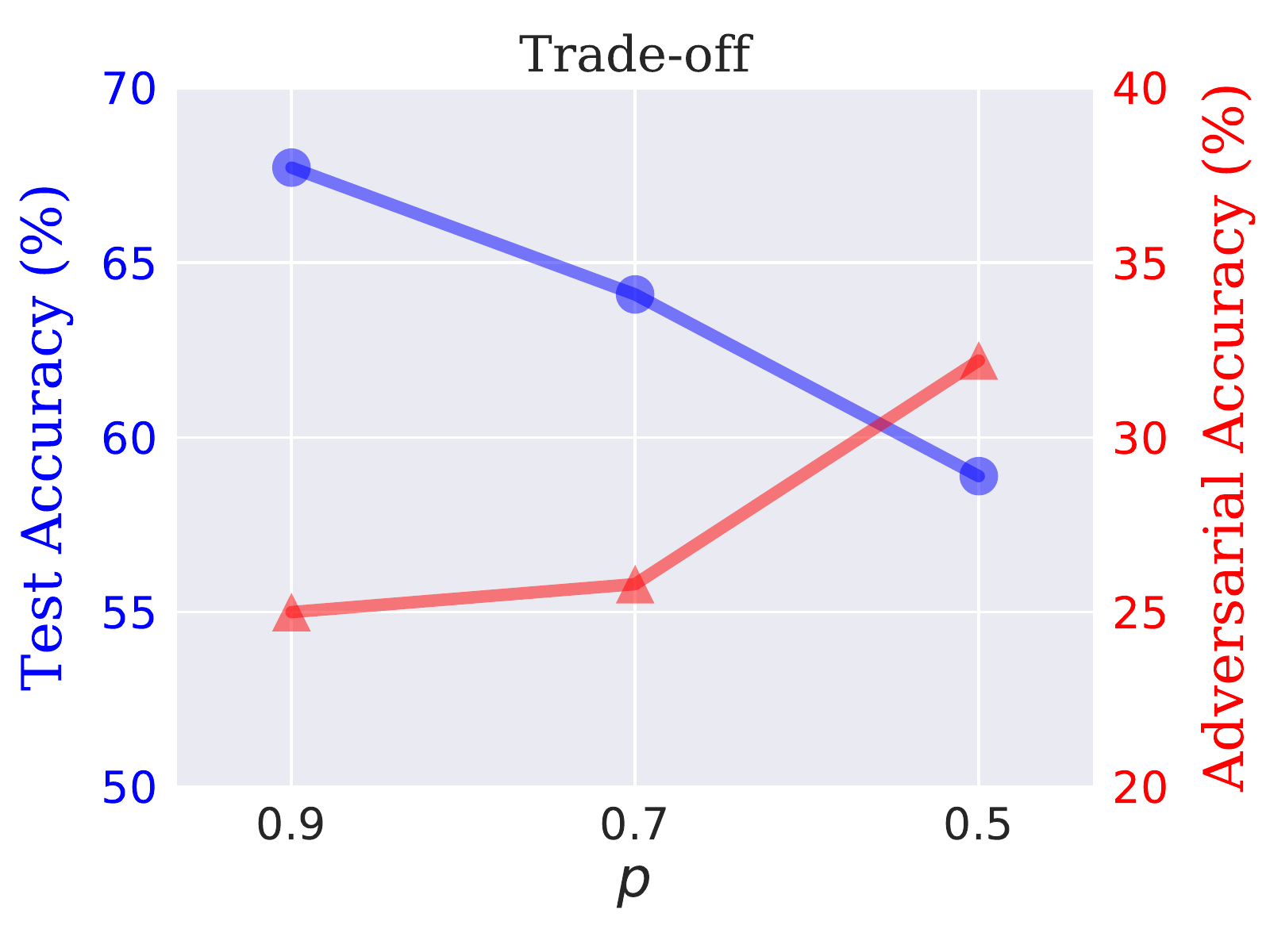}
}
\vspace{-0.3cm}
\caption{\textbf{The trade-off between adversarial robustness and standard generalization on different datasets}. We use pure ME-Net during training, and apply 7 steps white-box BPDA attack for the adversarial accuracy. For Tiny-ImageNet we only report the Top-1 accuracy. The results verify the consistent trade-off across different datasets.}
\label{fig:appendix tradeoff}
\vspace{-0.3cm}
\end{figure*}

We provide results of the inherent trade-off between adversarial robustness and standard generalization on different datasets. As shown in Fig.~\ref{fig:appendix tradeoff}, we change the observing probability $p$ of the masks to train different ME-Net models, and apply 7 steps white-box BPDA attack to each of them. As $p$ decreases, the generalization ability becomes lower, while the adversarial robustness grows rapidly. We show the consistent trade-off phenomena on different datasets.

\section{Additional Results of Different ME Methods}
\label{appendix:diff me methods}
\subsection{Black-box Attacks}
We first provide additional experimental results using different ME methods against black-box attacks. We train different ME-Net models on CIFAR-10 using three ME methods, including the USVT approach, the Soft-Impute algorithm and the Nuclear Norm minimization algorithm. The training processes are identical for all models. For the black-box adversary, we use different transfer-based attacks and report the results in Table~\ref{table:appendix me methods black}.

% =========================================================================== %
\begin{table}[!ht]
\centering
\scalebox{1.035}{
\begin{tabular}{@{}cccccccccc@{}}
\toprule
  &  &  &  &  & \multicolumn{3}{c}{\textbf{PGD}} & \multicolumn{2}{c}{\textbf{CW}}   \\ \cmidrule(lr){6-8}\cmidrule(l){9-10} 
\multirow{-2.5}{*}{\textbf{Method}} & \multirow{-2.5}{*}{\textbf{Complexity}} & \multirow{-2.5}{*}{\textbf{Type}} & \multirow{-2.5}{*}{\textbf{Clean}} & \multirow{-2.5}{*}{\textbf{FGSM}} & 7 steps           & 20 steps         & 40 steps         & $\kappa$ = 20        & $\kappa$ = 50        \\ \midrule\midrule
Vanilla                           & $-$                                   & $-$                             & 93.4\%                           & 24.8\%                          & 7.6\%             & 1.8\%            & 0.0\%            & 9.3\%           & 8.9\%           \\ \midrule\midrule
ME-Net - USVT                      & Low                                   & Prep.                           & 94.8\%                           & 90.5\%                          & 90.3\%            & 89.4\%           & 88.9\%           & \textbf{93.6\%} & \textbf{93.6\%} \\ \midrule
ME-Net - Soft-Imp.                 & Medium                                & Prep.                           & \textbf{94.9\%}                  & \textbf{92.2\%}                 & \textbf{91.8\%}   & \textbf{91.8\%}  & \textbf{91.3\%}  & \textbf{93.6\%} & 93.5\%          \\ \midrule
ME-Net - Nuc. Norm                 & High                                  & Prep.                           & 94.8\%                           & 92.0\%                          & 91.7\%            & 91.4\%           & 91.0\%           & 93.3\%          & 93.4\%          \\ \bottomrule
\end{tabular}
}
\vspace{-1ex}
\caption{\textbf{Comparison between different ME methods against black-box attacks.} We report the generalization and adversarial robustness of three ME-Net models using different ME methods on CIFAR-10. We apply transfer-based black-box attacks as the adversary.}
\label{table:appendix me methods black}
\end{table}
% =========================================================================== %

\subsection{White-box Attacks}
We further report the white-box attack results of different ME-Net models in Table~\ref{table:appendix me methods white}. We use 7 steps PGD to adversarially train all ME-Net models with different ME methods on CIFAR-10. We apply up to 1000 steps strongest white-box BPDA attacks as the adversary. Compared to the previous state-of-the-art~\cite{madry2017towards} on CIFAR-10, all the three ME-Net models can outperform them by a certain margin, while also achieving higher generalizations. The performance of different ME-Net models may vary slightly, where we can observe that more complex methods can lead to slightly better performance.
%We therefore envision even higher adversarial robustness of ME-Net with more powerful ME algorithms.

% =========================================================================== %
\begin{table}[!ht]
\centering
\scalebox{1.035}{
\begin{tabular}{@{}ccccccccc@{}}
\toprule
\multirow{2.5}{*}{\textbf{Method}} & \multirow{2.5}{*}{\textbf{Complexity}} & \multirow{2.5}{*}{\textbf{Type}} & \multirow{2.5}{*}{\textbf{Clean}} & \multicolumn{5}{c}{\textbf{Attack Steps}}                                               \\ \cmidrule(l){5-9} 
                                 &                                      &                                &                                 & \textbf{7}      & \textbf{20}     & \textbf{40}     & \textbf{100}    & \textbf{1000}   \\ \midrule\midrule
Madry                            & $-$                                  & Adv. train                     & 79.4\%                          & 47.2\%          & 45.6\%          & 45.2\%          & 45.1\%          & 45.0\%          \\ \midrule\midrule
ME-Net - USVT                     & Low                                  & Prep. + Adv. train             & \textbf{85.5\%}                 & 67.3\%          & 55.8\%          & 53.7\%          & 52.6\%          & 51.9\%          \\ \midrule
ME-Net - Soft-Imp.                & Medium                               & Prep. + Adv. train             & \textbf{85.5\%}                 & 67.5\%          & 56.5\%          & 54.8\%          & 53.0\%          & 52.3\%          \\ \midrule
ME-Net - Nuc. Norm                & High                                 & Prep. + Adv. train             & 85.0\%                          & \textbf{68.2\%} & \textbf{57.5\%} & \textbf{55.4\%} & \textbf{53.5\%} & \textbf{52.8\%} \\ \bottomrule
\end{tabular}
}
\vspace{-1ex}
\caption{\textbf{Comparison between different ME methods against white-box attacks.} We adversarially trained three ME-Net models using different ME methods on CIFAR-10, and compare the results with~\cite{madry2017towards}. We apply up to 1000 steps PGD or BPDA white-box attacks as adversary.}
\label{table:appendix me methods white}
\end{table}
% =========================================================================== %

\section{Additional Studies of Attack Parameters}
\label{appendix:random restart}
We present additional studies of attack parameters, including different random restarts and step sizes for further evaluations of ME-Net. Authors in~\cite{mosbach2018logit} show that using multiple random restarts and different step sizes can drastically affect the performance of PGD adversaries. We consider the same white-box BPDA-based PGD adversary as in Table~\ref{table:cifar white sota}, and report the results on CIFAR-10. Note that with $n$ random restarts, given an image, we consider a classifier successful only if it was not fooled by any of these $n$ attacks. In addition, this also significantly increases the computational overhead. We hence fix the number of attack steps as 100~(results are almost flattened; see for example Fig.~\ref{fig:white box}), and select three step sizes and restart values. We again compare ME-Net with~\cite{madry2017towards}.

% =========================================================================== %
\begin{table}[!ht]
\centering
\setlength{\tabcolsep}{3mm}{
\begin{tabular}{@{}ccccccccc@{}}
\toprule
\multirow{2.5}{*}{\textbf{Method}}                     & \multicolumn{1}{l}{\multirow{2.5}{*}{\textbf{Step sizes}}} & \multicolumn{3}{c}{\textbf{Random restarts}}        \\ \cmidrule(l){3-5} 
                                                     & \multicolumn{1}{l}{}                                     & 10              & 20              & 50              \\ \midrule\midrule
\multirow{3.9}{*}{Madry}                               & $2/255$                                                    & 43.4\%          & 42.7\%          & 41.7\%          \\ \cmidrule(l){2-5} 
                                                     & $4/255$                                                    & 43.8\%          & 43.3\%          & 41.9\%          \\ \cmidrule(l){2-5} 
                                                     & $8/255$                                                    & 44.0\%          & 43.3\%          & 41.9\%          \\ \midrule\midrule
\multirow{3.9}{*}{\textbf{ME-Net}} & $2/255$                                                    & \textbf{48.7\%} & \textbf{47.2\%} & \textbf{44.8\%} \\ \cmidrule(l){2-5} 
                                 & $4/255$                                                   & \textbf{49.7\%} & \textbf{48.4\%} & \textbf{45.2\%} \\ \cmidrule(l){2-5} 
                                 & $8/255$                                                    & \textbf{50.8\%} & \textbf{49.8\%} & \textbf{46.0\%} \\ \bottomrule
\end{tabular}
}
\vspace{-1ex}
\caption{\textbf{Results of white-box attacks with different random restarts and step sizes on CIFAR-10.} We compare ME-Net with~\cite{madry2017towards} using three different step sizes and random restart values. We apply 100 steps PGD or BPDA white-box attacks as adversary.}
\label{table:appendix random start}
\end{table}
% =========================================================================== %

As shown in Table~\ref{table:appendix random start}, with different step sizes, the performance of ME-Net varies slightly. Specifically, the smaller the step size~(e.g., $2/255$) is, the stronger the adversary becomes for both ME-Net and \cite{madry2017towards}. This is as expected, since a smaller step size enables a finer search for the adversarial perturbation.

\name\ leverages randomness through masking, and it would be helpful to understand how random restarts, with a hard success criterion, affect the overall pipeline. As observed in Table~\ref{table:appendix random start}, more restarts can reduce the robust accuracy by a few percent. However, we note that ME-Net can still outperform~\cite{madry2017towards} by a certain margin across different attack parameters. We remark that arguably, one could potentially always handle such drawbacks by introducing restarts during training as well, so as to maximally match the training and testing conditions. This introduces in unnecessary overhead that might be less meaningful. We hence focus on other parameters such as the number of attack steps in the main paper.%In this paper, we follow most of the literature by focusing on standard, common settings.% to enable a fair and meaningful comparison.
%Note that since ME-Net leverages randomness, the restart numbers can cause larger drop in accuracy for ME-Net. However, one can always handle such drawback by introducing more restarts during training. Therefore, we do not focus on such xxx.

\section{Additional Benefits by Majority Voting}
\label{appendix:majority vote}
It is common to apply an ensemble or vote scheme during the prediction stage to further improve accuracy. %However, in the adversarial setting, the attacker can apply different attacks for each model, which may lead to even worse result. Moreover, the computation overhead is greatly increased for training multiple models.
%\cite{engstrom2017rotation} proposed to augment the training with random input transformation~(e.g., rotations), and vote for a most predicted label over a number of random transformations as the output for each image in the inference time. By doing so, they hope to further increase classification accuracy in the adversarial setting. However, the input transformations are mainly handcrafted. The random transforms in the inference stage provides limited underlying rationale, and may even perform worse than the standard one.
ME-Net naturally provides a majority voting scheme. As we apply masks with different observation probability $p$ during training, an intuitive method is to also use multiple masks with the same $p$~(rather than only one $p$) for each image during inference, and output a majority vote over predicted labels. One can even use more masks with different $p$ within the training range. By such, the training procedure and model can remain unchanged while the inference overhead only gets increased by a small factor.

% ================================== Voting ================================= %
\begin{table}[!ht]
\centering
\setlength{\tabcolsep}{3mm}{
\begin{tabular}{@{}ccccccc@{}}
\toprule
\multirow{2.5}{*}{\textbf{\begin{tabular}[c]{@{}c@{}}Attack\\ Steps\end{tabular}}} & \multirow{2.5}{*}{\textbf{Method}} & \multirow{2.5}{*}{\textbf{MNIST}} & \multirow{2.5}{*}{\textbf{CIFAR-10}} & \multirow{2.5}{*}{\textbf{SVHN}} & \multicolumn{2}{c}{\textbf{Tiny-ImageNet}} \\ \cmidrule(l){6-7} 
                                                                                 &                                  &                                 &                                 &                                & Top-1                & Top-5               \\ \midrule\midrule
\multirow{2.5}{*}{\textbf{40}}                                                     & Standard                         & 94.0\%                          & 55.4\%                          & 71.9\%                         & 29.4\%               & 58.5\%              \\ \cmidrule(l){2-7} 
                                                                                 & \textbf{Vote}                    & \textbf{95.9\%}                 & \textbf{59.3\%}                 & \textbf{76.0\%}                & \textbf{33.8\%}      & \textbf{68.9\%}     \\ \midrule
\multirow{2.5}{*}{\textbf{100}}                                                    & Standard                         & 91.8\%                          & 53.5\%                          & 69.8\%                         & 29.0\%               & 58.0\%              \\ \cmidrule(l){2-7} 
                                                                                 & \textbf{Vote}                    & \textbf{94.2\%}                 & \textbf{56.2\%}                 & \textbf{73.1\%}                & \textbf{31.2\%}      & \textbf{65.4\%}     \\ \midrule
\multirow{2.5}{*}{\textbf{1000}}                                                   & Standard                         & 91.0\%                          & 52.8\%                          & 69.4\%                         & 28.5\%               & 56.9\%              \\ \cmidrule(l){2-7} 
                                                                                 & \textbf{Vote}                    & \textbf{92.6\%}                 & \textbf{54.2\%}                 & \textbf{71.4\%}                & \textbf{29.8\%}      & \textbf{59.5\%}     \\ \bottomrule
\end{tabular}
}
\vspace{-1ex}
\caption{\textbf{Comparison between majority vote and standard inference.} For each image, we apply 10 masks with same $p$ used during training, and the model outputs a majority vote over predicted labels. The standard inference only uses one mask with the mean probability of those during training. We use 40, 100 and 1000 steps white-box BPDA attack and report the results on each dataset.}
\label{table:voting}
\end{table}
% =========================================================================== %

In Table~\ref{table:voting}, we report the majority voting result of ME-Net on different datasets, where voting can consistently improve the adversarial robustness of the standard one by a certain margin. This is especially helpful in real-world settings where the defender can get more robust output without highly increasing the computational overhead.
Note that by using majority vote, we can further improve the state-of-the-art white-box robustness.

\section{Hyper-Parameters Study}
\label{appendix:hyperparam}
\subsection{Observation Probability $p$}
\begin{figure}[!htbp]
    \centering
    \includegraphics[width=1\textwidth]{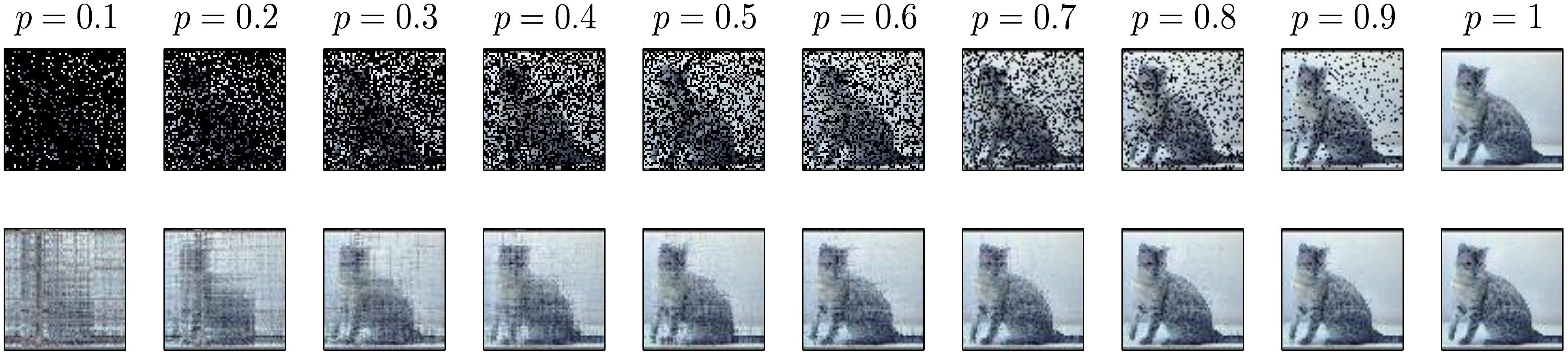}
\vspace{-0.3cm}
    \caption{\textbf{Visualization of ME result with different observation probability $p$.} \textbf{First row:} Images after applying masks with different observation probabilities. \textbf{Second row:} The recovered images by applying ME. We can observe that the global structure of the image is maintained even when $p$ is small.}
    \label{fig:appendix visual mc}
\vspace{-0.3cm}
\end{figure}

As studied previously, by applying different masks with different observation probability $p$, the performance of ME-Net can change differently. We have already reported extensive quantitative results of different ME-Net models trained with different $p$. Here we present the qualitative results by visualizing the effect of different $p$ on the original images.
As illustrated in Fig.~\ref{fig:appendix visual mc}, the first row shows the masked image with different $p$, and the second row shows the recovered image by ME. It can be observed that the global structure of the image is maintained even when $p$ is small.

\subsection{Number of Selected Masks}
Another hyper-parameter of ME-Net is the number of selected masked images for each input image. In the main paper, all experiments are carried out using 10 masks. We here provide the hyper-parameter study on how the number of masks affects the performance of ME-Net. We train ME-Net models on CIFAR-10 using different number of masks and keep other settings the same.
In Table~\ref{table:appendix mask number}, we show the results of both standard generalization and adversarial robustness. We use transfer-based 40 steps PGD as black-box adversary, and 1000 steps BPDA as white-box adversary. As expected, using more masks can lead to better performances. Due to the limited computation resources, we only try a maximum of 10 masks for each image. However, we expect ME-Net to perform even better with more sampled masks and better-tuned hyper-parameters.

% =========================================================================== %
\begin{table}[!ht]
\centering
\setlength{\tabcolsep}{3mm}{
\begin{tabular}{@{}cccccc@{}}
\toprule
\textbf{\# of Masks}          & \multicolumn{2}{c}{\textbf{Method}}    & \textbf{Clean}  & \textbf{Black-box} & \textbf{White-box} \\ \midrule\midrule
$-$                           & \multicolumn{2}{c}{Vanilla}            & 93.4\%          & 0.0\%              & 0.0\%                                     \\  \midrule\midrule
                              &                         & $p: 0.9$  & 92.7\%          & 82.3\%             & 44.1\%                                     \\ \cmidrule(l){3-6} 
\multirow{-2.5}{*}{\textbf{1}}  & \multirow{-2.5}{*}{ME-Net} & $p: 0.5$   & 79.8\%          & 59.7\%             & 47.4\%                                     \\ \midrule
                              &                         & $p: 0.8\to 1$ & 94.1\%          & 87.8\%             & 46.5\%                                    \\ \cmidrule(l){3-6} 
\multirow{-2.5}{*}{\textbf{5}}  & \multirow{-2.5}{*}{ME-Net} & $p: 0.4\to 0.6$ & 86.3\%          & 68.5\%             & 49.3\%                                    \\ \midrule
                              &                         & $p: 0.8\to 1$ & \textbf{94.9\%} & \textbf{91.3\%}    & 47.4\%                                    \\ \cmidrule(l){3-6} 
\multirow{-2.5}{*}{\textbf{10}} & \multirow{-2.5}{*}{ME-Net} & $p: 0.4\to 0.6$ & 89.2\%          & 70.9\%             & \textbf{52.8\%}                           \\ \bottomrule
\end{tabular}
}
\vspace{-1ex}
\caption{\textbf{Comparisons between different number of masked images used for each input image.} We report the generalization and adversarial robustness of ME-Net models trained with different number of masks on CIFAR-10. We apply transfer-based 40 steps PGD attack as black-box adversary, and 1000 steps PGD-based BPDA as white-box adversary.}
\label{table:appendix mask number}
\end{table}
% =========================================================================== %

\section{Additional Visualization Results}
\label{appendix:visualization}
\begin{figure*}[!htbp]
    \centering
    \includegraphics[width=1\textwidth]{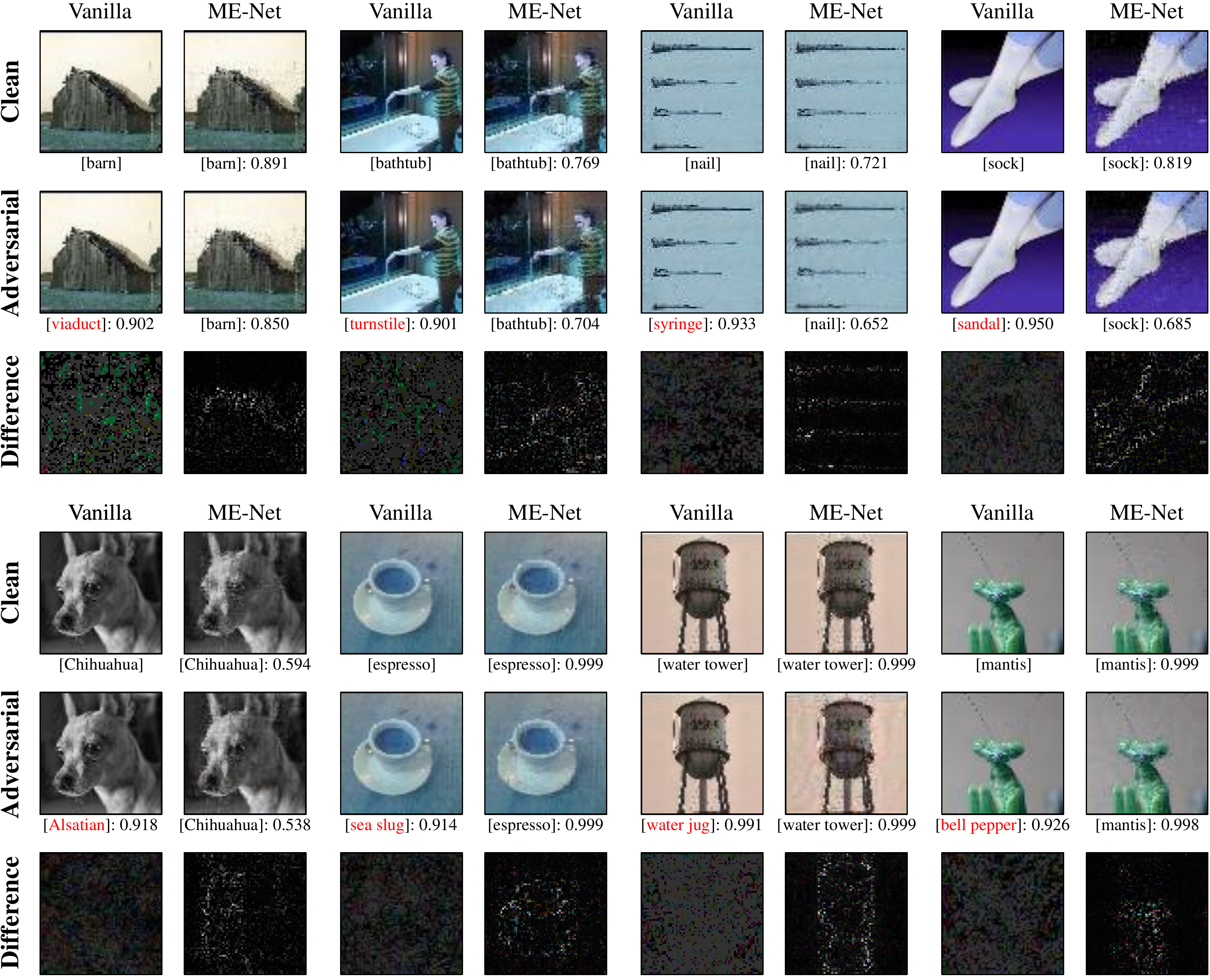}
\vspace{-0.3cm}
    \caption{\textbf{Visualization of ME-Net applied to clean images, adversarial images, and their differences on Tiny-ImageNet.} \textbf{First column} from top to bottom: the clean image, the adversarial example generated by PGD attacks, the difference between them~(i.e., the adversarial noises). \textbf{Second column} from top to bottom: the reconstructed clean image by ME-Net, the reconstructed adversarial example by ME-Net after performing PGD attacks, the difference between them~(i.e., the redistributed noises). Underlying each image is the predicted class and its probability. We multiply the difference images by a constant scaling factor to increase the visibility. The differences between the reconstructed clean image by ME-Net and the reconstructed adversarial example by ME-Net after performing PGD attacks, i.e., the new adversarial noises, are redistributed to the global structure.}
    \label{fig:appendix timgnet}
% \vspace{-0.3cm}
\end{figure*}

We finally provide more visualization results of ME-Net applied to clean images, adversarial images, and their differences. We choose Tiny-ImageNet since it has a higher resolution. As shown in Fig.~\ref{fig:appendix timgnet}, for vanilla model, the highly structured adversarial noises are distributed over the entire image, containing human imperceptible adversarial structure that is very likely to fool the network. In contrast, the redistributed noises in the reconstructed images from ME-Net mainly focus on the global structure of the images, which is well aligned with human perception. As such, we would expect ME-Net to be more robust against adversarial attacks.

% \begin{figure*}[!htbp]
%     \centering
%     \includegraphics[width=1\textwidth]{figures/appendix_cifar.pdf}
% \vspace{-0.3cm}
%     \caption{\textbf{Visualization of ME-Net applied to clean images, adversarial images, and their differences on CIFAR-10.} We multiply the difference images by a constant scaling factor to increase the visibility. The differences between the reconstructed clean image by ME-Net and the reconstructed adversarial example by ME-Net after performing PGD attacks, i.e., the new adversarial noises, are redistributed to the global structure.}
%     \label{fig:appendix cifar}
% % \vspace{-0.3cm}
% \end{figure*}

% \begin{figure*}[!htbp]
%     \centering
%     \includegraphics[width=1\textwidth]{figures/appendix_svhn.pdf}
% \vspace{-0.3cm}
%     \caption{\textbf{Visualization of ME-Net applied to clean images, adversarial images, and their differences on SVHN.} We multiply the difference images by a constant scaling factor to increase the visibility. The differences between the reconstructed clean image by ME-Net and the reconstructed adversarial example by ME-Net after performing PGD attacks, i.e., the new adversarial noises, are redistributed to the global structure.}
%     \label{fig:appendix svhn}
% % \vspace{-0.3cm}
% \end{figure*}

\end{document}